\documentclass[lettersize,journal]{IEEEtran}
\usepackage{amsmath,amsfonts}
\usepackage{algorithmic}
\usepackage{algorithm}
\usepackage{array}
\usepackage[caption=false,font=normalsize,labelfont=sf,textfont=sf]{subfig}
\usepackage{textcomp}
\usepackage{stfloats}
\usepackage{url}
\usepackage{verbatim}
\usepackage{graphicx}
\usepackage{cite}
\usepackage{amssymb}
\begin{document}

\title{BAFNet: Bilateral Attention Fusion Network for Lightweight Semantic Segmentation of Urban Remote Sensing Images}

\author{Wentao Wang and Xili Wang
\thanks{Manuscript received XXX XX, 20XX; revised XXX XX, 20XX. Wentao Wang and Xili Wang are with the School of Computer Science, Shaanxi Normal University, Xi'an 710119, China (email: \{wwt, wangxili\}@snnu.edu.cn). (Corresponding author: Xili Wang.)}}

\markboth{Journal of \LaTeX\ Class Files,~Vol.~xx, No.~xx, September~2024}%
{Shell \MakeLowercase{\textit{et al.}}: A Sample Article Using IEEEtran.cls for IEEE Journals}


\maketitle

\begin{abstract}
Large-scale semantic segmentation networks often achieve high performance, while their application can be challenging when faced with limited sample sizes and computational resources. In scenarios with restricted network size and computational complexity, models encounter significant challenges in capturing long-range dependencies and recovering detailed information in images. We propose a lightweight bilateral semantic segmentation network called bilateral attention fusion network (BAFNet) to efficiently segment high-resolution urban remote sensing images. The model consists of two paths, namely dependency path and remote-local path. The dependency path utilizes large kernel attention to acquire long-range dependencies in the image. Besides, multi-scale local attention and efficient remote attention are designed to construct remote-local path. Finally, a feature aggregation module is designed to effectively utilize the different features of the two paths. Our proposed method was tested on public high-resolution urban remote sensing datasets Vaihingen and Potsdam, with mIoU reaching 83.20$\%$ and 86.53$\%$, respectively. As a lightweight semantic segmentation model, BAFNet not only outperforms advanced lightweight models in accuracy but also demonstrates comparable performance to non-lightweight state-of-the-art methods on two datasets, despite a tenfold variance in floating-point operations and a fifteenfold difference in network parameters.
\end{abstract}


\section{Introduction}
\IEEEPARstart{T}{he} purpose of image semantic segmentation is to classify all pixels in the image, dividing it into several specific semantic regions to make the image easier to understand and analyze. With the advancement of deep learning technology, neural networks have become crucial tools for image segmentation because of their strong feature extraction capabilities \cite{zeng2024multiscale,han2020comparing,paul2024c,geng2023dual,cheng2020research}. Due to semantic segmentation being an intensive prediction task, neural networks need to extract high-resolution feature maps with pixel-level details and semantic information to produce satisfactory results, which is computationally expensive. Therefore, lightweight segmentation networks have attracted attention. In the field of remote sensing (RS) images, lightweight RS image segmentation models play a significant role in flood detection \cite{safavi2022comparative}, burned area detection \cite{bo2022basnet}, and weed detection in farmland \cite{deng2020lightweight}. In this paper, we developed a lightweight RS image segmentation model and demonstrated its effectiveness in enhancing the segmentation performance of lightweight models. The results of comparative experiments are shown in Fig. 1.
\begin{figure}[h!]
	\centering
	\includegraphics[width=3.5in]{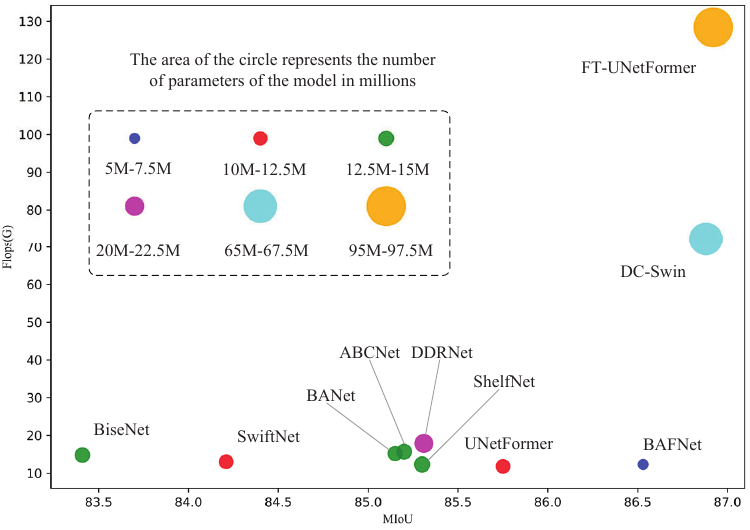}
	\caption{Comparison of floating-point operations (FLOPs) and mean intersection over union (mIoU, $\%$) of various models (results from the Potsdam dataset). Among them, DC-Swin and FT-UNetFormer are two non-lightweight state-of-the-art methods.} 
	\label{fig1}
\end{figure}

Semantic segmentation not only requires sufficient contextual information but also detailed information to achieve satisfactory results. Hence, an encoder-decoder structure is generally used. The encoder extracts contextual information from feature maps at various scales, while the decoder is responsible for restoring spatial resolution and generating segmentation maps of the original image size. In order to achieve high performance of segmentation models, optimization is mainly carried out from two aspects. One is to enhance the feature extraction process to capture more comprehensive context, thus acquiring information relevant to different categories. Another is to design a network structure that can fully restore detailed information, enabling the segmentation of object contours or small targets. For optimizing contextual information, the ability to capture long-range dependencies in images has been proven to be highly effective. The long-range dependencies in the image can guide the region based on the categories of surrounding objects, thereby improving segmentation accuracy. Convolutional neural networks (CNN) excel at extracting local fine-grained information due to the locality of convolution operations, but they struggle to effectively capture long-range dependencies in images. Some methods enhance CNN by expanding the receptive field to capture a wide range of dependencies in images, such as the dilated convolution \cite{chen2018encoder}, the large kernel convolution \cite{ding2022scaling}. 
For the precise preservation of detailed information, the commonly used approach is feature fusion, which fully utilizes multi-scale features to recover spatial details. U-Net \cite{ronneberger2015u} adopted a progressive upsampling strategy and proposes skip connections to gradually fuse feature maps from the encoding process to restore detailed information during decoding. MANet \cite{li2021multiattention} further enhanced the feature fusion process of skip connections by incorporating attention operations to merge features from the encoding process. DC-Swin \cite{wang2022novel} designed a new dense connection feature aggregation module (DCFAM) to fully utilize multi-scale information. FT-UNetFormer \cite{wang2022unetformer} adopted a U-shaped encoder-decoder structure and designed a global-local attention module to pay attention to both global and local information in the decoder.

However, performance-driven approaches often use larger encoders for feature extraction and design complex decoders to fuse features, resulting in a large number of parameters and a significant increase in computations of the segmentation network. In order to adapt to application scenarios that require high efficiency and lack computational power, it is necessary to provide lightweight and effective segmentation models. In addition to reducing network size and computational complexity, lightweight segmentation models also need to consider two aspects: acquiring contextual information and preserving detailed information to maintain segmentation performance as much as possible. At present, lightweight segmentation networks typically employ lighter networks for feature extraction, such as ShuffleNet \cite{zhang2018shufflenet}, MobileNet \cite{howard2017mobilenets}, ResNet18 \cite{he2016deep}, and others. Utilizing these networks for feature extraction can significantly reduce the size of the segmentation network. There are two primary architectures for lightweight segmentation networks designed to preserve detailed information. One is to continue using an encoder-decoder architecture that integrates features from the encoding process through skip connections \cite{zhuang2019shelfnet,orsic2019defense}. However, the detailed information lost during multiple downsampling operations in the feature extraction process cannot be fully recovered through a simple decoding structure. So, BiseNet \cite{yu2018bisenet} proposed a bilateral segmentation architecture, which separates the extraction of contextual information and detailed information. Specifically, the input image enters two parallel paths. One is the contextual path, continuously downsampling to low resolution to acquire rich semantic information. The other is the spatial path, which maintains high resolution. Simple convolution operations are used to extract local detailed information. This independent path design achieves the effect of capturing both contextual information and local detailed information, and the network is sufficiently lightweight. Fast-SCNN \cite{poudel2019fast} proposed learn-to-downsample strategy to improve the bilateral segmentation network. As the first few layers of the contextual path extract low-level features with details, the two paths can share the computations of these early layers instead of constructing the detailed or spatial path from scratch. DDRNet \cite{pan2022deep} adopted a bilateral structure. It incorporated a learn-to-downsample and introduced connections between the two paths to enhance information exchange between the contextual path and detailed path.

The aforementioned lightweight segmentation networks still face challenges due to computational limitations. Firstly, when it comes to extracting contextual information from RS images, it is essential to acknowledge that different objects often display spectra that are highly similar. Relying solely on local information may lead to inaccurate segmentation. Therefore, it is crucial to capture the long-range dependencies within the image to achieve precise segmentation of RS images. The ResNet18, widely used in lightweight segmentation models, is efficient. However, small-scale convolutions cannot effectively capture long-range dependencies or large-scale contextual information in the image. This limitation hinders the performance of the segmentation network during the encoding stage. Secondly, when it comes to restoring detailed information, using lightweight encoder-decoder structures, performing deconvolution on high-resolution feature maps during decoding will add extra computational burden.
Moreover, due to the disparity in information content between feature maps with rich semantic information and those with detailed information, direct concatenation cannot effectively restore most of the detailed information.
For most of bilateral networks, the detailed path is overly simplistic. By relying solely on a few basic convolutional layers to extract detailed information, a significant amount of unknown or irrelevant information is included. This introduces noise to the segmentation process, thereby limiting the effectiveness of enhancing the segmentation performance through the detailed path.
The insufficient extraction of contextual and detailed information ultimately leads to lightweight segmentation models that require further improvement. In RS image segmentation, there is a lack of lightweight and satisfactory segmentation models. This motivates us to delve deeper into this particular field.

Given the above limitations,
we propose a bilateral network comprising a dependency path and a remote-local path. The dependency path integrates visual attention networks (VAN) \cite{guo2023visual}, leveraging the large kernel attention (LKA) in VAN to efficiently capture long-range dependencies. The remote-local path originates from the shallow layers of the dependency path and consistently maintains high-resolution. For the remote-local path, we have devised an efficient remote attention module (ERAM) along with a multi-scale local attention module (MSLAM) to facilitate its construction. These modules empower the path to capture long-range dependencies and detailed information within the image. Furthermore, while extracting features by two paths, the features obtained from the dependency path undergo two exchanges with the features extracted from the remote-local path. These exchanges allow the remote-local path to acquire more profound semantic information, while the dependency path with lower-resolution feature maps can acquire some detailed information. Finally, a feature aggregation module (FAM) is developed to integrate the features generated by two paths with the goal of enhancing fusion outcomes.

The contributions of this article are given sa follows:

1. An efficient remote attention module has been proposed to capture large-scale contextual information or long-range dependencies, alongside a novel multi-scale local attention module designed to capture detailed local information.

2. A remote-local path is constructed by integrating the efficient remote attention module and the multi-scale local attention module as the central components. Additionally, visual attention network is incorporated as the dependency path to efficiently capture long-range dependencies. A feature aggregation module is developed to aggregate the features from two paths, thereby creating a bilateral lightweight segmentation network known as BAFNet.

3. The effectiveness of the proposed method has been validated through experimentation on public datasets Vaihingen and Potsdam, achieving mIoU scores of 83.20$\%$ and 86.53$\%$, respectively. This enhancement has led to improved segmentation performances compared to current lightweight segmentation models. Furthermore, when compared to state-of-the-art high-performance segmentation networks that utilize large-scale Transformers as encoders, BAFNet demonstrates comparable segmentation performance while reducing floating-point operations by a factor of ten.

\section{Related Work}
In this section, we primarily discuss representative non-lightweight high-performance segmentation models and lightweight segmentation models.
\subsection{High Performance Segmentation Model}
The early encoder-decoder architecture based on CNNs achieved success in semantic segmentation tasks \cite{long2015fully,ronneberger2015u,badrinarayanan2017segnet}.
The spatial constraints imposed by convolutional operations in CNN-based segmentation networks result in restricted receptive fields. Consequently, these networks may struggle to capture large-scale contextual information or long-range dependencies effectively, potentially resulting in misclassification of certain objects. Furthermore, the downsampling conducted in the process of feature extraction may lead to the elimination of small-scale features, consequently leading to inadequate retrieval of small targets during the decoding phase.
In addressing the challenges related to limited receptive fields and the lack of long-range dependencies, several studies \cite{ding2022scaling,yu2015multi,peng2017large} enhanced the convolution operation to increase its receptive field. Some studies chose to integrate attention mechanisms \cite{li2021multiattention,fu2019dual,huang2019ccnet} to enhance traditional CNNs. Several studies utilized the Transformer \cite{vaswani2017attention} for visual segmentation tasks \cite{zheng2021rethinking,cao2022swin,strudel2021segmenter,xie2021segformer,ranftl2021vision}, resulting in a notable enhancement in segmentation performance. To address the challenge of lossing small targets or detailed information, the main approach consists of reducing the downsampling rate or implementing feature fusion. For the method of reducing downsampling rate, the Deeplab series \cite{liang2015semantic,chen2017deeplab,chen2017rethinking} utilized dilated convolutions with specific dilation rates to expand the receptive field and capture sophisticated semantic features from high-resolution feature maps.
HRNet \cite{sun2019high} maintained the original image resolution during feature extraction by employing parallel branches with varying resolutions. This approach enables the extraction of advanced semantic information while preserving detailed information effectively.
For feature fusion strategies, certain studies leveraged the advantage of CNNs to capture local fine-grained information by integrating features at the encoder \cite{he2022swin}. Transfuse \cite{zhang2021transfuse} integrated Transformer and CNN in parallel, and proposed a novel BiFusion module to fuse the features extracted by both Transformer and CNN. Transunet \cite{chen2021transunet} integrated CNN and Transformer sequentially. It took features from the CNN, transformed them dimensionally, and fed them as input sequences to the Transformer blocks for capturing global context. In the decoding phase, the decoder upsampled the encoded features and merged them with the high-resolution feature maps produced by the CNN to ensure accurate localization.
Other studies focus on the meticulous development of decoders to integrate various scale features obtained from the encoder and recover intricate details.
FT-UNetFormer leveraged the UNet structure and incorporated Transformer blocks with global-local attention in its decoding process to effectively preserve both global and local information. Additionally, a feature refinement head was introduced to enhance skip connections and facilitate the fusion of feature maps containing spatial detailed information with those containing high-level semantic information. DC-Swin proposed a DCFAM in its decoding phase to extract semantic features enriched by multi-scale relationships for precise segmentation.
Both FT-UNetFormer and DC-Swin utilized the Swin Transformer \cite{liu2021swin} for feature extraction, enabling the encoder to acquire rich long-range dependencies. Furthermore, these two methods included carefully designed modules to ensure the restoration of detailed information, leading to their superior performance on RS datasets. However, despite their high accuracy, the large encoder and decoder architectures result in a large number of parameters and computational complexity, which may hinder their practical application.
\subsection{Lightweight Segmentation Model}
\subsubsection{Lightweight encoder-decoder structure}
Lightweight encoder-decoder segmentation models typically avoid computationally intensive operations such as dilated convolutions or large kernel convolutions. Instead, they primarily utilize lightweight networks constructed with small-scale or separable convolutions for feature extraction. For instance, ShelfNet \cite{zhuang2019shelfnet} incorporated multiple encoder-decoder branch pairs at each spatial level, leveraging skip connections to integrate information from encoders and decoders across varying feature map scales. Moreover, the network implemented weight sharing strategies within residual blocks to significantly reduce computational complexity. Similarly, SwiftNet \cite{orsic2019defense} employed a U-shaped encoder-decoder architecture featuring a lightweight ResNet18 encoder. It introduced a lightweight upsampling mechanism with horizontal connections for dense prediction to ensure efficiency and a pyramid pooling technique to combine representations across different spatial levels, thereby expanding receptive fields and enhancing overall performance. DFANet \cite{li2019dfanet} introduced a lightweight xception network based on depthwise separable convolution \cite{chollet2017xception} as the backbone of the segmentation network. This method aggregated features of various scales through cascaded sub-networks and sub-stages.
ShuffleSeg \cite{gamal2018shuffleseg} utilized ShuffleNet \cite{zhang2018shufflenet} as its backbone, incorporating channel rearrangement and group convolution techniques from ShuffleNet to reduce computational costs. CgNet \cite{wu2020cgnet} introduced CG blocks to effectively extract features from local, surrounding, and global contexts, thereby reducing parameters and memory usage. FANet \cite{hu2020real} achieved a balance between speed and accuracy through a fast attention module and downsampling strategy. SFNet \cite{li2020semantic} introduced a flow alignment module (FAM) to align adjacent feature maps for improved fusion. UNetFormer \cite{wang2022unetformer} employed ResNet18 as the encoder and a Transformer-like decoder to capture global and local context information during decoding.
\subsubsection{Dual-path structure}
The lightweight segmentation methods based on encoder-decoder structure conduct multiple downsampling stages in the encoding process to extract high-level semantic information associated with categories. Nonetheless, the spatial details lost due to downsampling cannot be entirely recovered by the decoder, thus affecting the segmentation outcomes.
BiseNet \cite{yu2018bisenet} proposed a dual path method, where the input image was fed into two separate paths: the context path, which rapidly downsampled the image to capture semantic information, and the spatial path, which maintained a high resolution to extract detailed information using basic convolution operations.
This architecture improved the performance of lightweight segmentation networks. BiseNetv2 \cite{yu2021bisenet} designed a guided aggregation layer aimed at strengthening the connectivity between the semantic path and detail path by integrating feature representations of different paths. In addition, an enhanced training strategy was designed to improve segmentation performance without increasing any inference cost. 
ContextNet \cite{poudel2018contextnet} had two inputs, with images of different resolutions being fed into two different branches. In order to achieve real-time performance, the deeper and lower-resolution branches were used to perceive context and provide category information, while the higher-resolution branches were shallower to refine segmentation boundaries. Fast-scnn \cite{poudel2019fast} proposed learn-to-downsample structure, which enables two separate paths to share the initial downsampling layers, thus reducing computational complexity. DDRNet \cite{pan2022deep} adopted the strategy of sharing downsampling layers and exchanged information twice between two paths during feature extraction. In addition, a deep aggregation pyramid pooling module (DAPPM) was designed to enhance the semantic information obtained from semantic path. In the field of RS image segmentation, ABCNet \cite{li2021abcnet} adopted the same bilateral architecture as BiseNet, introduced linear attention into lightweight segmentation model to enhance the features output by the context path, and designed a feature aggregation module with linear attention as the core to fully aggregate the features output by the context and spatial paths. Meanwhile, BANet \cite{wang2021transformer} introduced a lightweight Transformer Backbone \cite{zhang2021rest} in the semantic path to extract features, enabling lightweight segmentation models to capture long-range dependencies in images. However, these segmentation models with a dual-path structure pay little attention to extremely simple detailed path. Despite maintaining a smaller model size, there is still room for improvement in terms of accuracy.

\section{Networks}
The overall structure of BAFNet is shown in Fig. 2 (a), which is a bilateral structure composed of dependency path and remote-local path. The dependency path utilizes VAN to capture long-range dependencies during feature extraction. By downsampling four times, the feature maps in four VAN stages are respectively 1/4, 1/8, 1/16, and 1/32 of the original image size. The VAN stage is stacked from VAN blocks as shown in Fig. 2 (b).
The remote-local path is constructed using remote-local attention blocks (RLAB, shown in Fig. 2 (c)), with the number of blocks employed in the three stages being 2, 1, and 1, respectively. The feature maps are maintained at a relatively high resolution, which is 1/8 of the original image size, to extract long-range dependencies while capturing enough detailed information.
When extracting features from two separate paths, information exchange between the two paths is conducted twice. Finally, we aggregate the feature maps from the high-resolution path and the low-resolution path using FAM and input them into the segmentation head. The segmentation head consists of a 3x3 convolution followed by a 1x1 convolution. After the 3x3 convolution, the number of channels of the feature map is halved, and then mapped into the number of categories through the 1x1 convolution. The segmentation map of the original image is obtained through 8x bilinear interpolation upsampling.
\begin{figure*}[h!]
	\centering
	\subfloat[]
	{\includegraphics[width=2.0\columnwidth]{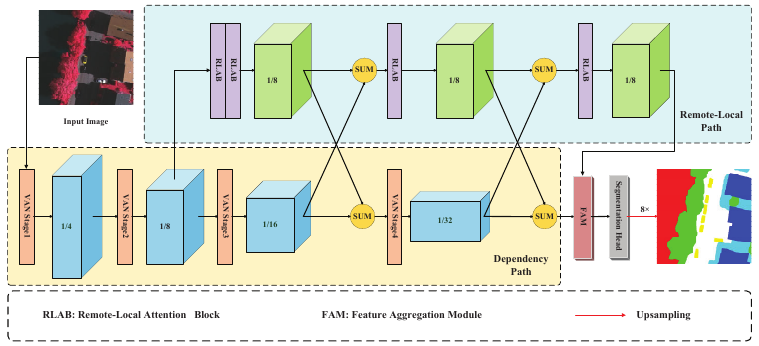}}\\
	\subfloat[]
	{\includegraphics[width=0.998\columnwidth]{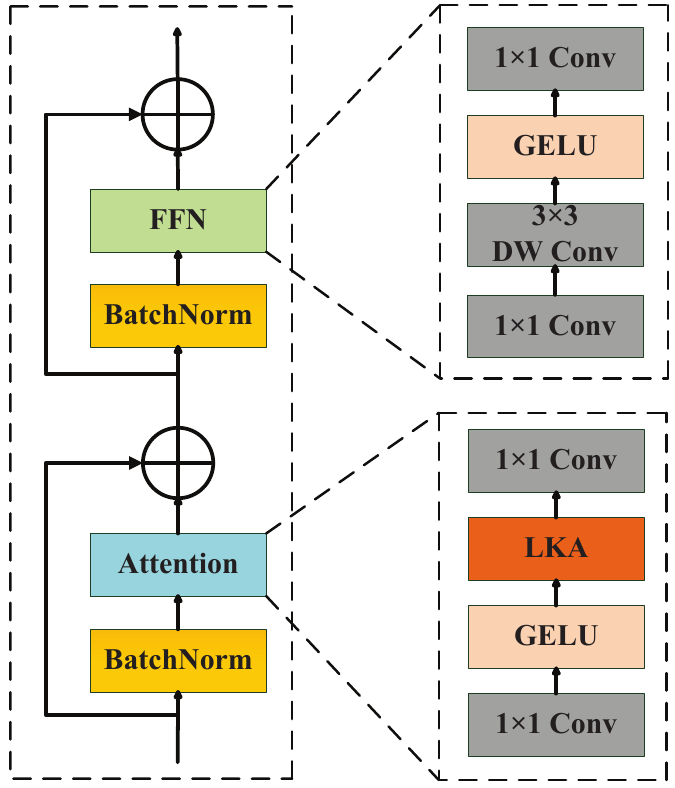}}
	\subfloat[]
	{\includegraphics[width=1.0\columnwidth]{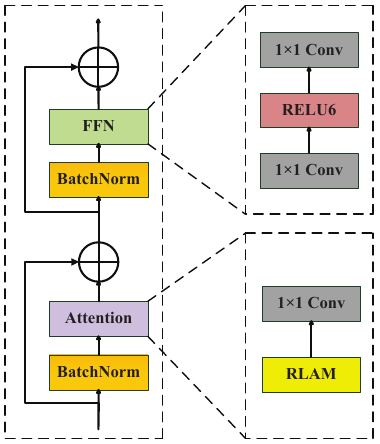}}
	\caption{(a) BAFNet, (b) VAN block, (c) RLAB.} \label{fig2}
\end{figure*}

\subsection{Dependency Path}

We introduce VAN as the feature extraction network for dependency path. Compared to the small-scale convolution used by ResNet, the large kernel convolution has a larger receptive field, which can effectively capture long-range dependencies. However, the larger convolution kernel significantly increases the computational complexity.
VAN has further improved the large kernel convolution by reducing the computational cost, enabling efficient capture of long-range dependencies within the image. Specifically, VAN decomposes the large kernel convolution using a depthwise convolution, a depthwise dilation convolution, and a 1x1 convolution, proving that the decomposed three components can effectively replace the large kernel convolution. Based on this, the LKA is proposed, as shown in Fig. 3. 
The LKA employs decomposed large kernel convolution to calculate the attention map for the input. Subsequently, it multiplies the attention map with the original input to generate the output. The calculation process is shown in Eq. 1 and Eq. 2.
\begin{equation}
	Attention = Con{v_{1 \times 1}}(DW\mbox{-}D\mbox{-}Conv(DW\mbox{-}Conv(F))),
\end{equation}
\begin{equation}
	Output = Attention \otimes F,
\end{equation}
among them, $DW\mbox{-}Conv$ represents depthwise convolution, $DW\mbox{-}D\mbox{-}Conv$ represents depthwise dilated convolution, and $Con{v_{1 \times 1}}$ represents 1x1 convolution. The input $F$ is sequentially processed through depthwise convolution, depthwise dilated convolution, and 1x1 convolution to derive the attention value, which is then multiplied element-wise with the input.
\begin{figure}[h!]
	\centering
	\includegraphics[scale=0.35]{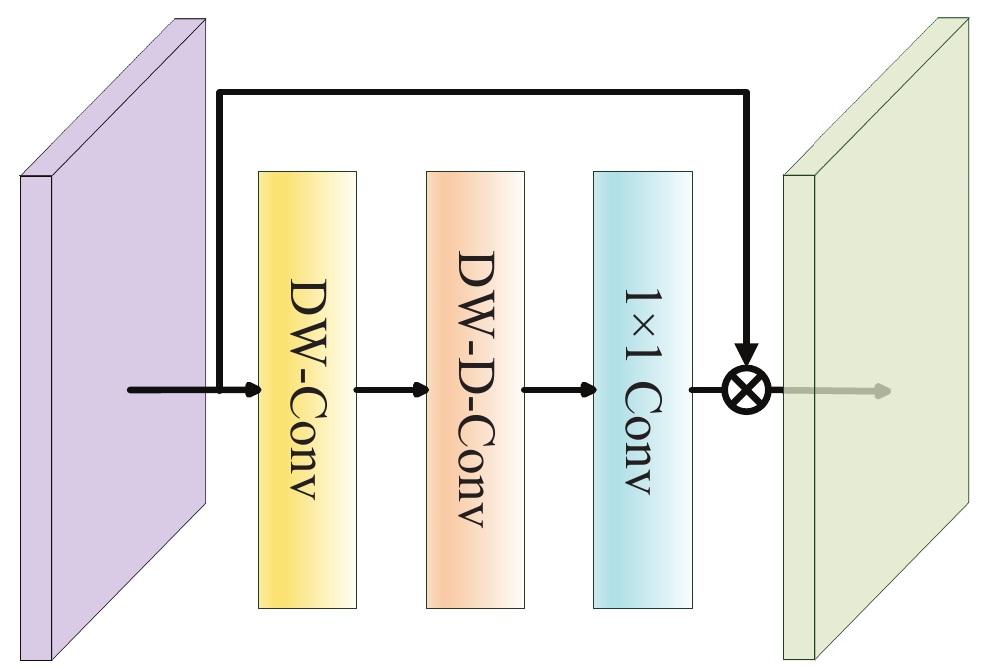}
	\caption{LKA.}
	\label{fig3}
\end{figure}

VAN focuses on LKA and constructs a VAN block, as shown in Fig. 2 (b). The model adpots a hierarchical Transformer-like architecture and offers seven versions, denoted from VAN-B0 to VAN-B6, which vary based on the network's size. To construct a lightweight segmentation model, we employ the smallest VAN-B0 as the feature extraction network for the dependency path, which has been pretrained on the ImageNet dataset. The number of feature map channels output by the four stages is 32, 64, 160, and 256, respectively.

\subsection{Remote-local Path}
The remote-local path is established by utilizing the RLAB as illustrated in Fig. 2 (c). The detailed structure of remote-local attention module (RLAM) is shown in Fig. 4. In Fig. 4, window partition refers to the window partitioning operation utilized in the Swin Transformer.
\begin{figure*}
	\centering
	\includegraphics[width=1.90\columnwidth]{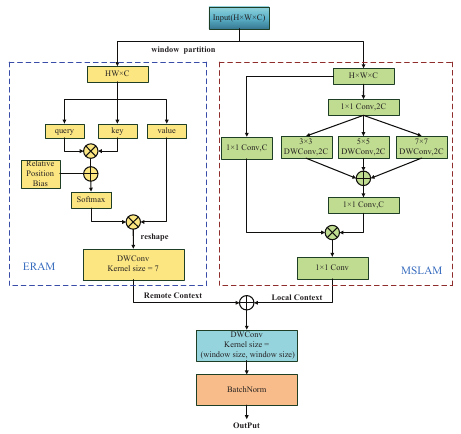}
	\caption{Remote-local attention module (RLAM).}
	\label{fig4}
\end{figure*}

\subsubsection{Multi-scale local attention module}
We design a multi-scale local attention module, as shown in Fig. 4, to fully capture detailed local information on high-resolution feature maps.
The inverted bottleneck, as shown in Fig. 5 (a), increases the channel numbers of the feature map using 1x1 convolution. It extracts features on feature maps with increased channels and subsequently reduces the channels to their original input using 1x1 convolution. This process ultimately enhances accuracy.
We use multiple depthwise separable convolutions with different scales to extract detailed information, then integrate them into the inverted bottleneck structure. To reduce computation, the expansion factor of the channels is set to 2, resulting in the multi-scale inversed bottleneck (MSIB) as shown in Fig. 5 (b).
In contrast to self-attention mechanism that involve generating three components of query (q), key (k), and value (v) to compute attention based on the similarity between q and k, our proposed MSLAM exclusively gets attention through the MSIB structure. As shown in Eq. 3 to Eq. 6, given an input $X \in {R^{h \times w \times c}}$, first, $\hat X$ is obtained by expanding the input channel $c$ using a 1x1 convolution. Subsequently, depthwise separable convolutions with kernel sizes of 3, 5, and 7 are utilized to extract features for $\hat X$. The outputs generated from the three separate scale branches are summed, and the summed features are subsequently processed through a 1×1 convolution to decrease the feature dimensionality to $c$. The obtained $Attention$ will be multiplied by the linearly transformed input. Finally, a 1×1 convolution will be applied to acquire the output of MSLAM.
\begin{equation}
	\hat X = conv_{1 \times 1}^{c \to 2c}(X),
\end{equation}
\begin{equation}
	\begin{split}
		Attention = conv_{1 \times 1}^{2c \to c}(DW\mbox{-}con{v_{3 \times 3}}(\hat X)\\
		+ DW\mbox{-}con{v_{5 \times 5}}(\hat X)\\
		+ DW\mbox{-}con{v_{7 \times 7}}(\hat X)),
	\end{split}
\end{equation}
\begin{equation}
	Value = conv_{1 \times 1}^{c \to c}(X),
\end{equation}
\begin{equation}
	Out = conv_{1 \times 1}^{c \to c}(Attention \otimes Value).
\end{equation}
\begin{figure}[!h]
	\centering
	\subfloat[]
	{\includegraphics[scale=0.65]{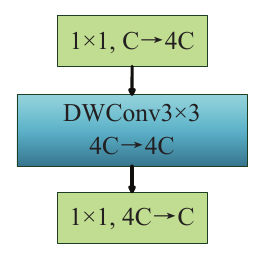}}\\
	\subfloat[]
	{\includegraphics[scale=0.58]{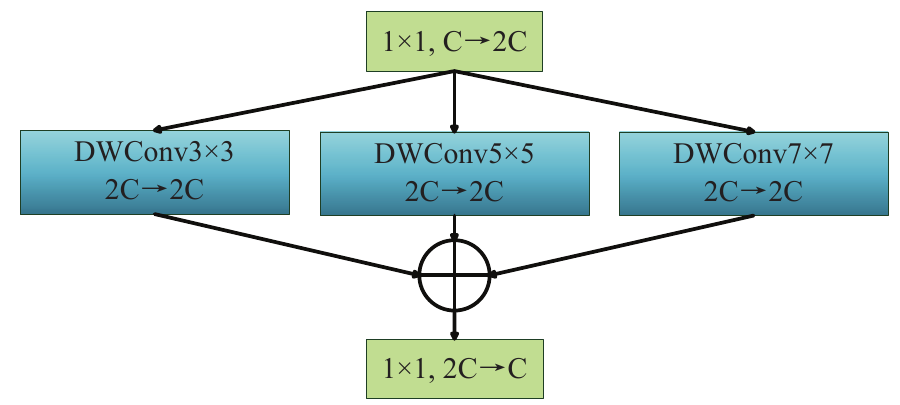}}
	\caption{Different bottleneck structures, (a) Inverted bottleneck, (b) Multi-scale inverted bottleneck}
	\label{fig5}
\end{figure}
\subsubsection{Efficient remote attention module}
In addition to capturing multi-scale detailed information comprehensively, remote-local path also possess specific capabilities for extracting long-range contextual information. To acquire semantic information containing long-range dependencies extracted from dependency path for guiding the extraction of detailed information, it is essential for remote-local path to possess corresponding structures that can capture these long-range dependencies. The self-attention mechanism employed in the Vision Transformer \cite{dosovitskiy2020image} demonstrates a capacity for global modeling, enabling the capture of global information within feature maps. Nevertheless, the computational complexity escalates significantly with the growing resolution of input images, as it compute attention values across the entire image range. The Swin Transformer divides the feature map into non-overlapping window regions, calculates attention values within each window, and employs shift operations to facilitate information exchange among distinct windows. 
However, the process of shift operations are time-consuming. Consequently, a Swin Transformer block contains two distinct attention calculation processes, namely multi-head self-attention (MHSA) and shift window multi-head self-attention (SW-MHSA). Stacking the attention calculation process twice within a block in a lightweight segmentation network may not be optimal. To fully utilize the window self-attention in the Swin Transformer and facilitate its efficient and flexible into lightweight segmentation networks, we have removed the shift operations. Instead, a 7×7 deepwise convolution is introduced to facilitate information exchange among different windows, following window attention. Hence, we have introduced ERAM, as shown in Fig. 4. ERAM has the ability to capture long-range contextual information or dependencies, and its calculation process is shown in Eq. 7 and Eq. 8.
\begin{equation}
	{\hat h^l} = W\mbox{-}MHSA(LN({h^{l - 1}})) + {h^{l - 1}},
\end{equation}
\begin{equation}
	{h^l} = Conv(MLP(LN({\hat h^l})) + {\hat h^l}),
\end{equation}
among them, ${h^{l - 1}}$ represents the input to $l\mbox{-}th$ ERAM, $LN$ represents layer normalization operation, $W\mbox{-}MHSA$ stands for the multi-head self-attention in the window, $MLP$ represents MLP module in Swin Transformer block, $Conv$ indicates a 7×7 depthwise convolution, and ${h^l}$ represents the output of the entire ERAM.
\subsection{Information Exchange Between Paths}
After the construction of two paths, we exchange information between the two paths twice during the feature extraction process. The exchanges aim to merge specific detailed information into the dependency path and to incorporate profound semantic information from the dependency path into the remote-local path, thereby enhancing the representation capabilities of each path. During the exchange process, the feature maps from the two paths are fused via element-wise addition. Given the different resolutions and channel numbers of the feature maps intended for fusion from the two paths, it is essential to make adjustments to align the feature maps. The principle is that the feature map intended for integration into another path should be adjusted to match the target feature map in both channels and resolution.
Specifically, the approach involves utilizing a 1x1 convolution, followed by normalization, to modify the channels. To adjust the resolution, integrating a low-resolution feature map into a high-resolution feature map requires the application of bilinear interpolation. Conversely, the integration of a high-resolution feature map into a low-resolution feature map requires downsampling the high-resolution feature map by employing several convolution operations with a stride of 2, until its resolution matches that of the low-resolution feature map. 
To reduce computational complexity, the channel number of each output feature map from the remote-local path is set to 128.

\subsection{Feature Aggregation Module}
The dependency path offers lower resolution and abstract semantic information, while the remote-local path contains a wide range of contextual information and rich detailed information. The contents and resolutions of the two paths demonstrate differences. We aim to fully utilize the output features from both paths to improve the segmentation results. To achieve this, we have devised a FAM to integrate the features outputted by the two paths, as shown in Fig. 6.
\begin{figure}
	\centering
	\includegraphics[width=1.0\columnwidth]{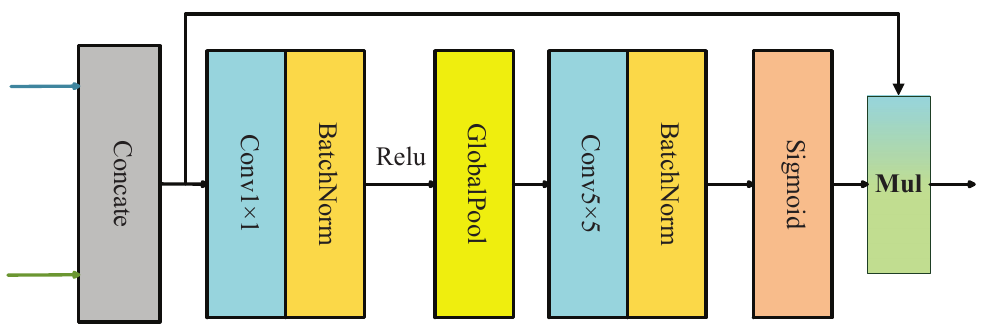}
	\caption{Feature aggregation module}
	\label{fig6}
\end{figure}

For the output of two paths, $low$ and $high$, where $low$ represents the feature map of the final output of the dependency path with lower resolution, and $high$ represents the feature map of the final output of the remote-local path with higher resolution. We first use 1x1 convolution to align the channel of the $low$ with that of the $high$. Subsequently, we upsample the output to match the resolution of $high$, resulting in the transformed output $low'$.
\begin{equation}
	low' = {U_4}(con{v_{1 \times 1}}(low)),
\end{equation}
where, ${U_4}$ represents the bilinear interpolation upsampling operation, with an upsampling factor of 4.

FAM first concatenates the feature maps $low'$ and $high$, which possess the same channel numbers and resolutions, to generate the feature map $FuseFeat$. Subsequently, the $FuseFeat$ undergoes a linear transformation, global pooling, and a 5×5 convolution. The sigmoid function is used to obtain the weights of the channels. Then channel weights are multiplied by $FuseFeat$, which is equivalent to the process of selecting and combining different channels. Similar to the channel attention \cite{hu2018squeeze} , FAM better utilizes the distinct features generated by the two paths. The specific process is shown in Eq. 10 and Eq. 11.
\begin{equation}
	FuseFeat = Concat(low',high),
\end{equation}
\begin{equation}
	\begin{split}
		OutPut = f(BN(con{v_{5 \times 5}}(GAP(W * FuseFeat))))\\
		\otimes FuseFeat,
	\end{split}
\end{equation}
where $high$ represents the output feature map of the remote-local path, $Concat$ represents the concatenation operation applied to the feature map along the channel dimension, $FuseFeat$ represents the concatenated feature map, $W$ stands for the linear transformation, $GAP$ denotes the global average pooling operation, $con{v_{5 \times 5}}$ refers to the convolution with the kernel size of 5, $BN$ is the batch normalization, and $f$ represents the sigmoid function.

\subsection{Loss Function}
In RS datasets, a common occurrence is the phenomenon of class imbalance, characterized by a significant variation in the number of samples across different categories. The overall segmentation performance is frequently suboptimal as a result of imbalanced categories. Categories with fewer samples often struggle to attain satisfactory segmentation results. The dice loss \cite{li2019dice} has been shown to be effective in preventing overfitting on categories with a higher number of samples while ensuring adequate attention is given to categories with fewer samples. We use a hybrid loss function that integrates cross-entropy (CE) loss and dice loss.
The hybrid loss enables the model to focus more on categories that are challenging to segment, consequently enhancing the model's performance on imbalanced datasets. Eq. 12 and Eq. 13 present the CE loss and dice loss, respectively. Eq. 14 presents the hybrid loss employed for network training. 
\begin{equation}
	\begin{split}
		{L_{CE}} =  - \frac{1}{N}\sum\nolimits_{i = 1}^N \sum\nolimits_{c = 1}^C {y_c^{(i)}\log \hat y_c^{(i)}}\\
			+ (1 - y_c^{(i)})\log (1 - \hat y_c^{(i)}),
	\end{split}
\end{equation}
\begin{equation}
	{L_{dice}} = 1 - \frac{2}{N}\sum\nolimits_{i = 1}^N {\sum\nolimits_{c = 1}^C {\frac{{\hat y_c^{(i)}y_c^{(i)}}}{{\hat y_c^{(i)} + y_c^{(i)}}}} },
\end{equation}
\begin{equation}
	L = {L_{CE}} + {L_{dice}},
\end{equation}
among them, $N$ represents the number of samples, $C$ represents the number of categories, ${y^{(i)}}$ represents the one-hot encoding of labels, ${\hat y^{(i)}}$ represents the softmax output of the segmentation network, ${\hat y_c^{(i)}}$ is the confidence of sample $i$ belonging to the category $c$.

\section{Experiments}
\subsection{Experimental Setup}
\subsubsection{Datasets} 
There are two datasets used in the experiment, namely Vaihingen and Potsdam.

Vaihingen: The Vaihingen dataset consists of 33 orthorectified top images, each of which has three multispectral bands (red, green, near-infrared) and a digital surface model (DSM). The ground sampling distance (GSD) is 9 centimeters, and the average size of the images is 2494x2064 pixels. This dataset includes five foreground categories (impervious surface, building, low vegetation, tree, car) and one clutter. As with previous work, we used 17 images for testing, IDs: 2, 4, 6, 8, 10, 12, 14, 16, 20, 22, 24, 27, 29, 31, 33, 35, 38. The remaining 16 images are used for training. During the experiment, we did not use DSM.

Potsdam: The Potsdam dataset consists of 38 orthorectified top images, each of which has four multispectral bands (red, green, blue, near-infrared) as well as DSM and standardized DSM. The GSD is 5 centimeters, and the size of each image is 6000x6000 pixels. The data category of this dataset is the same as that of the Vaihingen dataset. We used 14 images for testing, IDs: 2$\_$13, 2$\_$14, 3$\_$13, 3$\_$14, 4$\_$13, 4$\_$14, 4$\_$15, 5$\_$13, 5$\_$14, 5$\_$15, 6$\_$13, 6$\_$14, 6$\_$15, 7$\_$13. After removing images $7\_10$ with incorrect annotations, the remaining 23 images were used for training. During the experiment, only the red, green, and blue channels were used.

\subsubsection{Evaluation indicators}
The performance of the model on the dataset is evaluated through two aspects. In terms of segmentation performance, it is measured by the overall accuracy (OA), mean intersection over union (mIoU), and mean F1 score calculated on the cumulative confusion matrix. In terms of model complexity, it is measured by the number of parameters and floating-point operations (FLOPs).
\begin{equation}
	{\rm{OA = }}\frac{{\sum\nolimits_{k = 1}^N {T{P_k}} }}{{{{\sum\nolimits_{k = 1}^N {TP} }_k} + F{P_k} + T{N_k} + F{N_k}}},
\end{equation}
where ${\rm{T}}{{\rm{P}}_k}$, ${\rm{F}}{{\rm{P}}_k}$, ${\rm{T}}{{\rm{N}}_k}$, and ${\rm{F}}{{\rm{N}}_k}$ represent the true positive, false positive, true negative, and false negative, respectively, for object indexed as class $k$.
\begin{equation}
	mIoU = \frac{1}{N}\sum\nolimits_{k = 1}^N {\frac{{T{P_k}}}{{T{P_k} + F{P_k} + F{N_k}}}},
\end{equation}
\begin{equation}
	F1 = 2 \times \frac{{precision \times recall}}{{precision + recall}},
\end{equation}
the precision and recall are as follows:
\begin{equation}
	precision = \frac{1}{N}\sum\nolimits_{{\rm{k}} = 1}^{\rm{N}} {\frac{{T{P_k}}}{{T{P_k} + F{P_k}}}}
\end{equation}
\begin{equation}
	recall = \frac{1}{N}\sum\nolimits_{{\rm{k}} = 1}^{\rm{N}} {\frac{{T{P_k}}}{{T{P_k} + F{N_k}}}}
\end{equation}

\subsubsection{Comparison methods}
The comparison methods include efficient lightweight segmentation models and non-lightweight high-performance models. The lightweight segmentation models include networks with encoder-decoder structure: SwiftNet \cite{orsic2019defense}, ShelfNet \cite{zhuang2019shelfnet}, UNetFormer \cite{wang2022unetformer}, and networks with bilateral structure: BiseNet \cite{yu2018bisenet}, ABCNet \cite{li2021abcnet}, BANet \cite{wang2021transformer}, DDRNet \cite{pan2022deep}. And two non-lightweight models, FT-UNetFormer \cite{wang2022unetformer} and DC-Swin \cite{wang2022novel}, which are two state-of-the-art methods on Vaihingen and Potsdam. 

\subsubsection{Implementation details}
All models in the experiment were implemented using the PyTorch framework and executed on a single NVIDIA GTX 3090 GPU. The optimizer used is AdamW, with an initial learning rate set to 2e-4 and weight decay set to 1e-4. The learning rate is adjusted using cosine strategy.

For two datasets, the large image patches were cropped to 512x512 pixels. During the training process, enhancement techniques such as random scale ([0.5, 0.75, 1.0, 1.25, 1.5]), random vertical flipping, random horizontal flipping, and random 90 degree rotation were used, with a training epoch of 60 and a batch size of 16. During the testing phase, Test-time augmentation (TTA) techniques were used, such as multi-scale ([0.5, 0.75, 1.0, 1.25, 1.5]) and random flip augmentation.
\subsection{Experimental Results}
\begin{table*}
	\caption{Segmentation Results on the Vaihingen Dataset\label{tab:table1}}
	\centering
	\begin{tabular}{|c||c||c||c||c||c||c||c||c||c||c|}
		\hline
		Method & Parameters & FLOPs & Imp. Surf. & Building & Low Veg.& Tree & Car & mean F1 & OA($\%$) &mIoU($\%$) \\
		\hline
		BiseNet \cite{yu2018bisenet} & 13.4M & 14.8G & 92.24 & 95.21 & 84.07 & 90.00 & 82.37 & 88.78& 90.50& 80.16  \\
		\hline
		SwiftNet \cite{orsic2019defense} & 11.8M & 13.0G & 91.73 & 94.35 &83.85 & 89.24& 78.82
		&87.59 & 89.80 & 78.36 \\
		\hline
		ShelfNet \cite{zhuang2019shelfnet} & 14.6M & 12.3G & 92.66 & 95.46 & 84.89 & 89.82 & 83.79 & 89.32 & 90.78 & 81.00 \\
		\hline
		BANet \cite{wang2021transformer} &12.7M&15.2G&92.50&95.21&84.43&89.84&85.65&89.53&90.63	&81.29\\
		\hline
		ABCNet \cite{li2021abcnet} &14.0M&15.7G&92.49&95.48&83.91&90.14&85.56&89.51&90.73&81.29\\
		\hline
		DDRNet \cite{pan2022deep} &20.3M&17.9G&93.10&95.21&84.08&\pmb{90.27}&88.46&90.22&90.90&82.41\\
		\hline
		UNetFormer\cite{wang2022unetformer} &11.7M&\pmb{11.8G}&93.07&95.56&84.30&90.07&88.41&90.28&91.00&82.51\\
		\hline
		BAFNet (ours)&\pmb{6.4M}&12.3G&\pmb{93.24}&\pmb{95.90}&\pmb{84.75}&89.98&\pmb{89.66}&\pmb{90.70}&\pmb{91.16}&\pmb{83.20}\\
		\hline
		DC-Swin \cite{wang2022novel} &66.9M&72.2G&93.35&95.93&84.90&90.17&88.76&90.62&91.29&83.07\\
		\hline
		FT-UNetFormer \cite{wang2022unetformer} &96M&128.4G&93.17&96.03&84.92&90.14&90.61&90.97&91.26&83.65\\
		\hline
	\end{tabular}
\end{table*}

Table 1 shows the segmentation results of various comparison methods on the Vaihingen dataset, where the best value for each item between the lightweight network is marked in bold. Our proposed BAFNet demonstrates superior performance compared to other lightweight networks in terms of mean F1, OA, and mIoU while keeping network parameters and FLOPs at a low level. It is noteworthy that, in comparison to the current state-of-the-art lightweight model UNetFormer on the Vaihingen dataset, BAFNet has shown varying degrees of improvement in Imp. Surf., Building, Low veg., and Car categories, especially with an increase of over 1.2$\%$ in the Car category, indicating that the BAFNet we proposed performs best on smaller-scale targets.
When compared to the two best performing non-lightweight models on the Vaihingen dataset, namely DC-Swin and FT-UNetFormer, it is evident that DC-Swin employs a stronger Swin-Small \cite{liu2021swin} as feature extractor, which possesses 10 times more parameters and 6 times more FLOPs than BAFNet. BAFNet even exhibits a marginal advantage in OA and mIoU. For the FT-UNetFormer that employs Swin-Base as the encoder, the number of parameters and FLOPs is 15 and 10 times greater than that of BAFNet, respectively. 
However, BAFNet is very close to it in terms of various accuracy metrics, with an overall accuracy lag of only 0.1$\%$ and mIoU lagging by less than 0.5$\%$. This suggests that BAFNet, employing a lightweight encoder along with meticulously designed efficient remote attention module, multi-scale local attention module, and feature aggregation module, is capable of achieving performance levels comparable to models with considerably higher complexity.
The segmentation results of various comparison methods on the Vaihingen test set as shown in Fig. 7,
demonstrating the effectiveness of the proposed method.
\begin{figure*}
	\centering{}
	\begin{minipage}[t]{0.092\linewidth}
		\centering
		\includegraphics[scale=0.094]{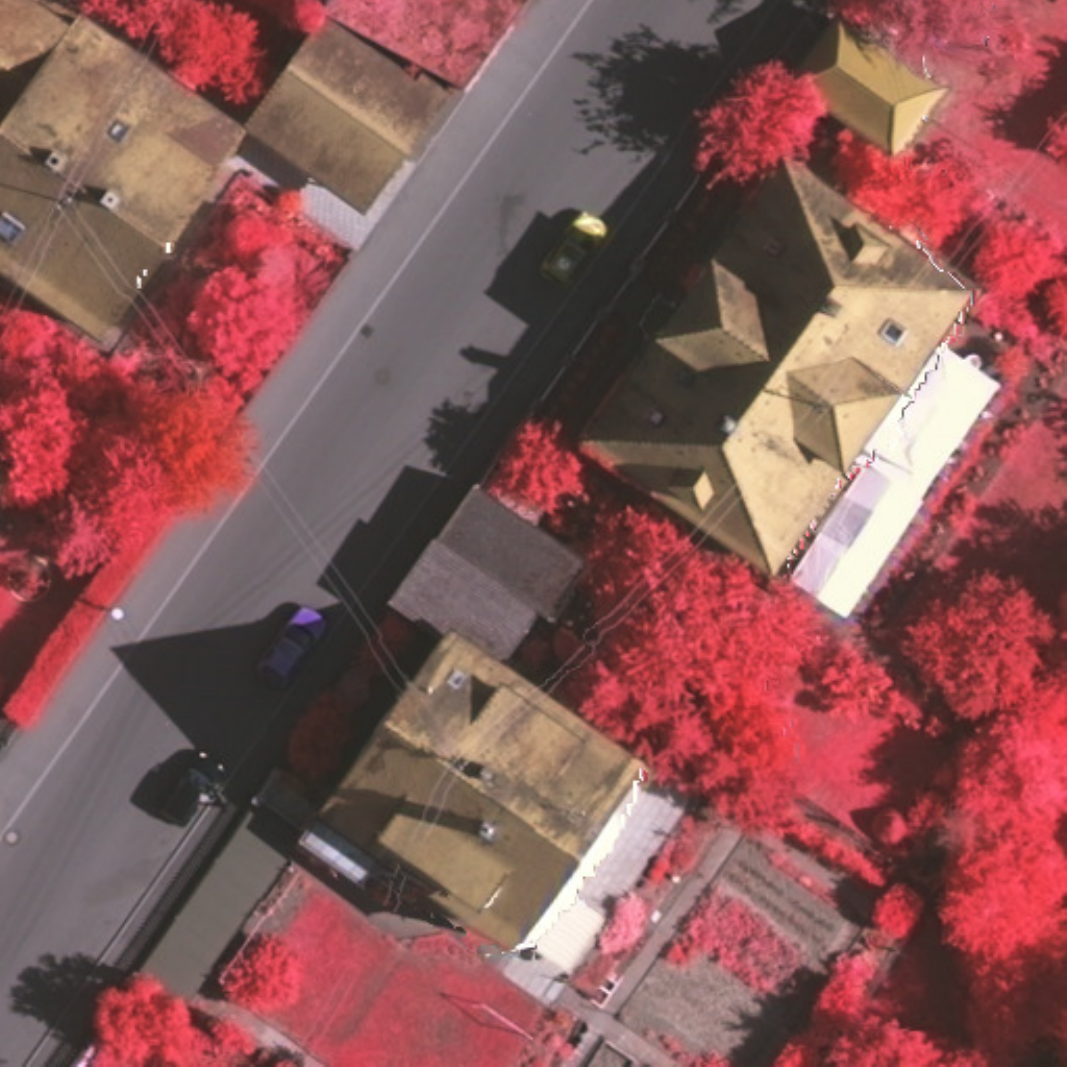}
		
		\vspace{1mm}
		
		\includegraphics[scale=0.094]{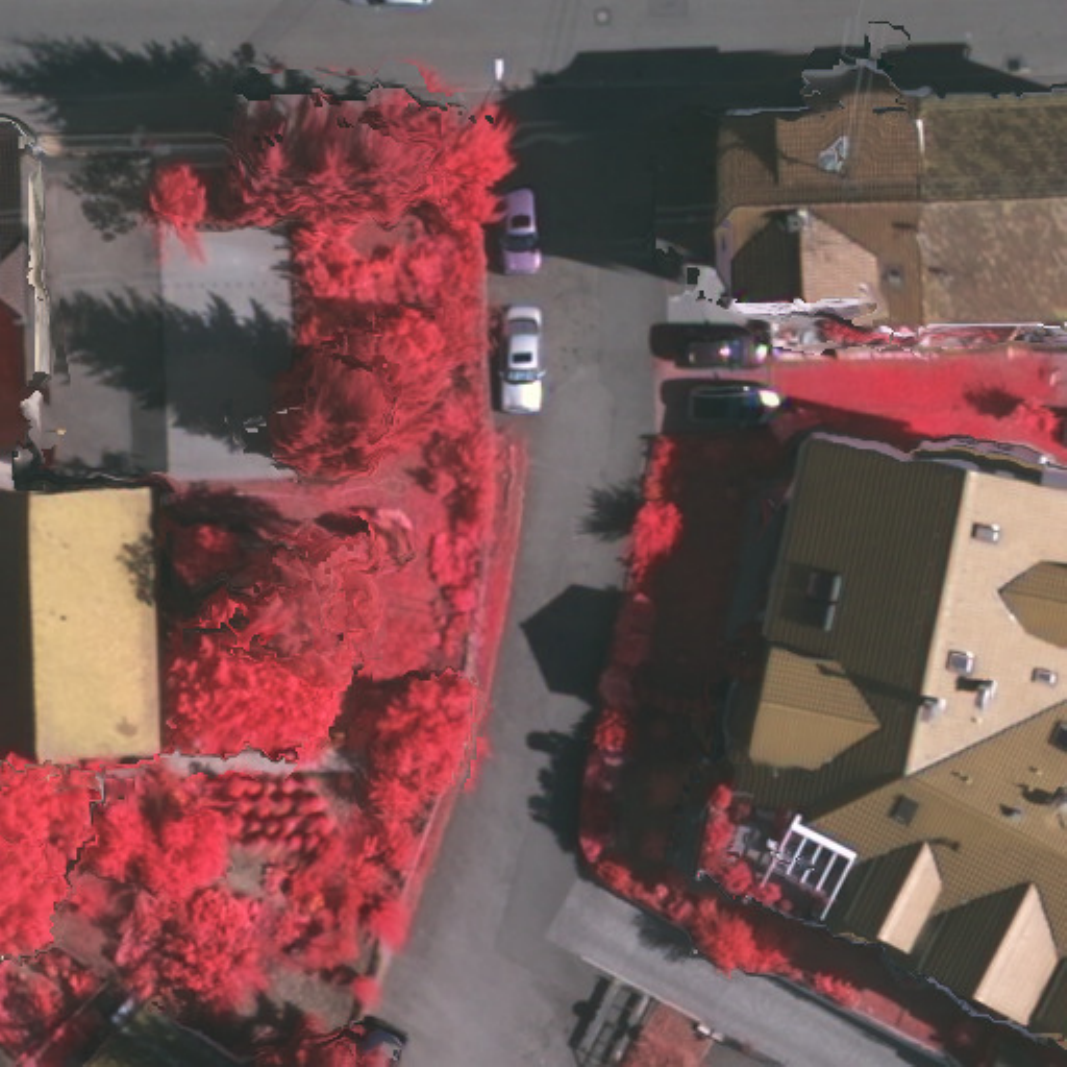}
		
		\vspace{1mm}
		
		\includegraphics[scale=0.094]{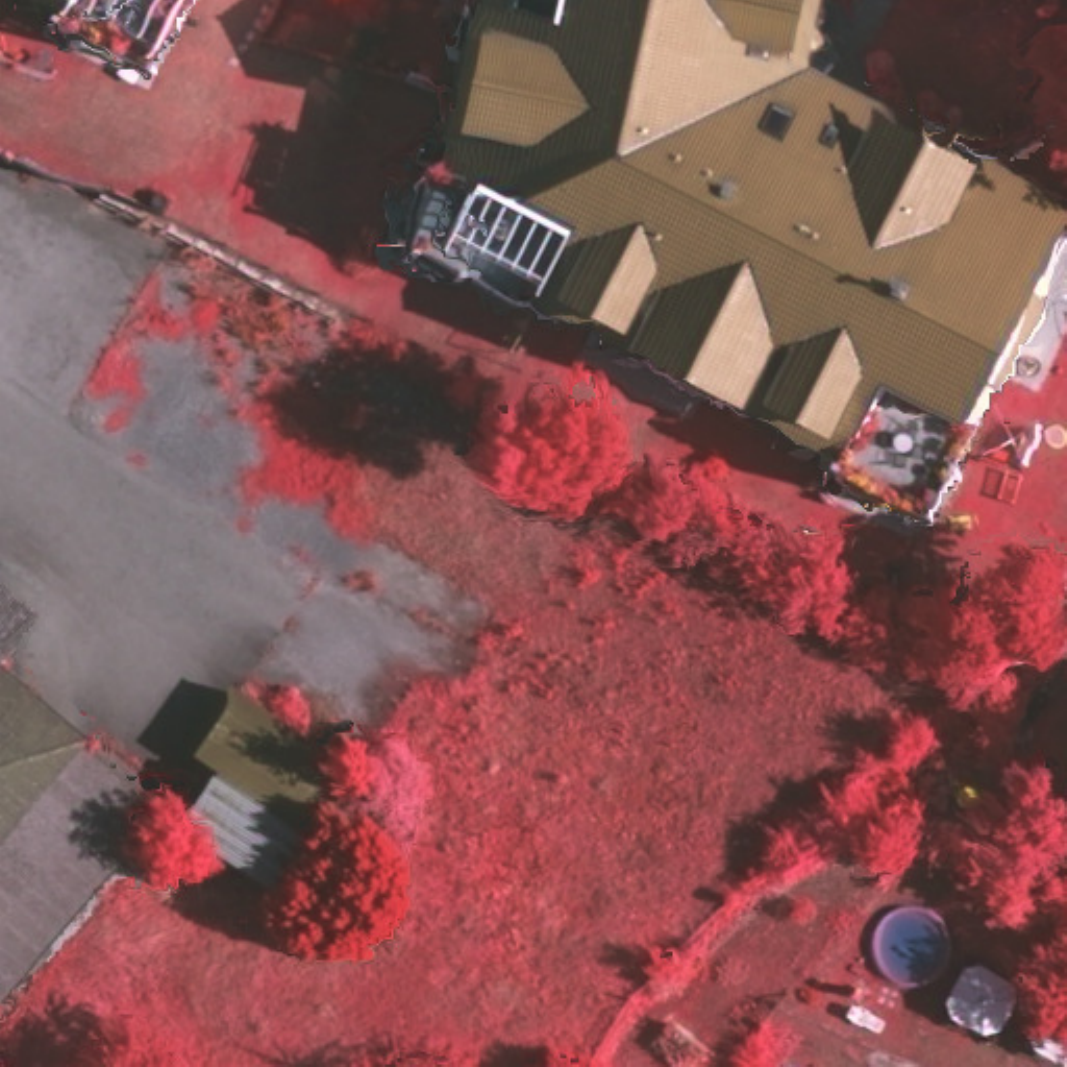}
		
		\vspace{1mm}
		
		\includegraphics[scale=0.094]{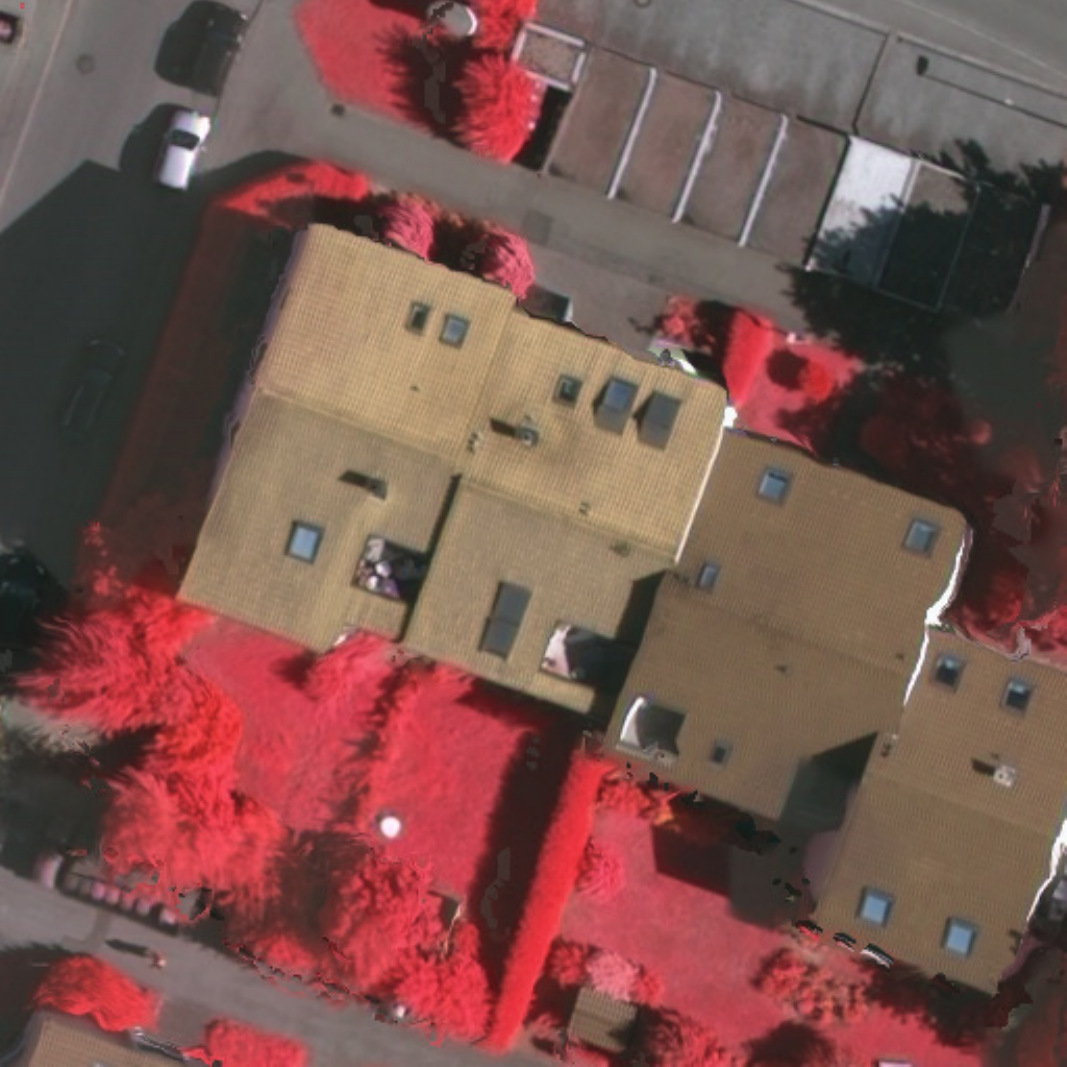}
		
		\vspace{1mm}
		
		\includegraphics[scale=0.094]{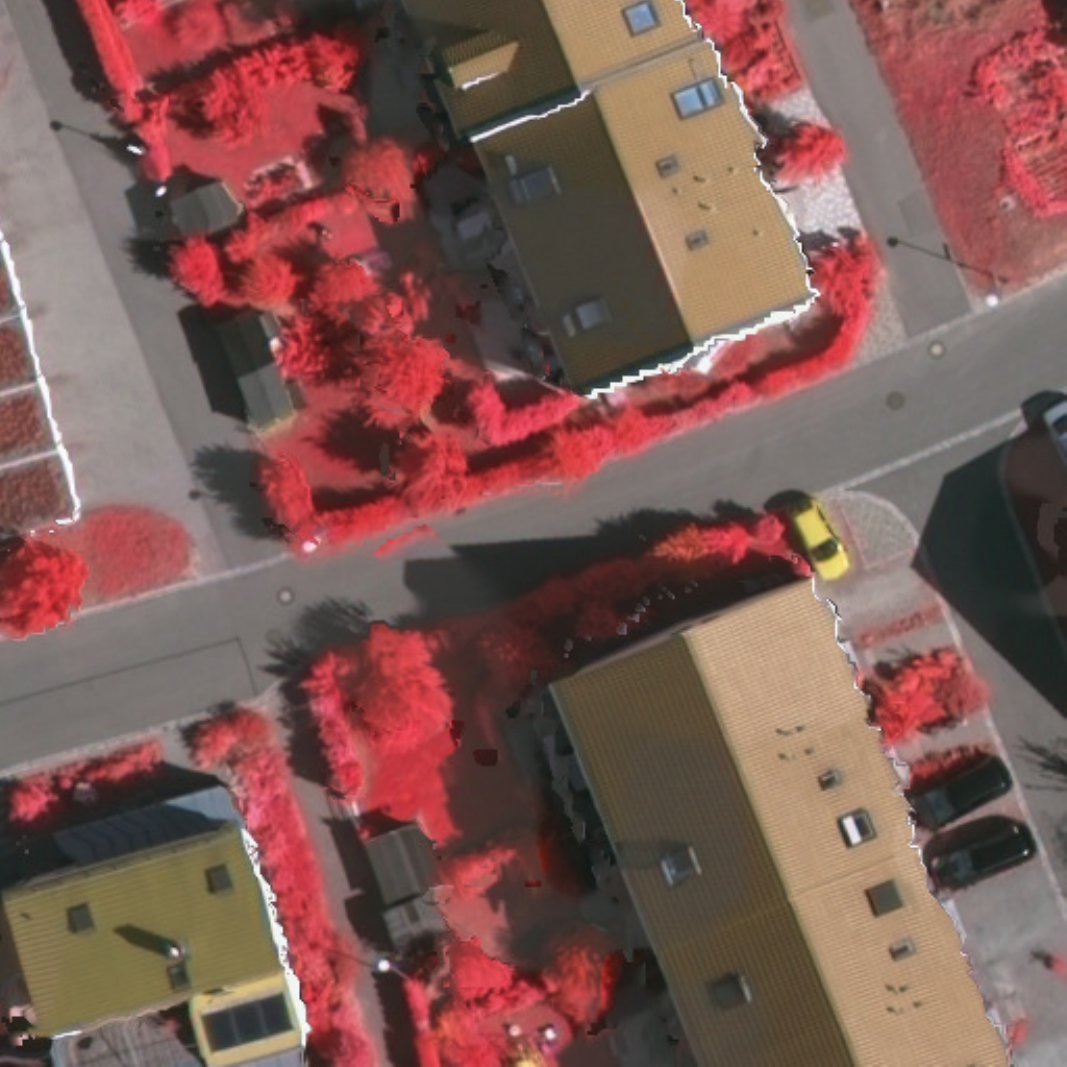}
		\centerline{(a)}
	\end{minipage}
	\begin{minipage}[t]{0.092\linewidth}
		\centering
		\includegraphics[scale=0.094]{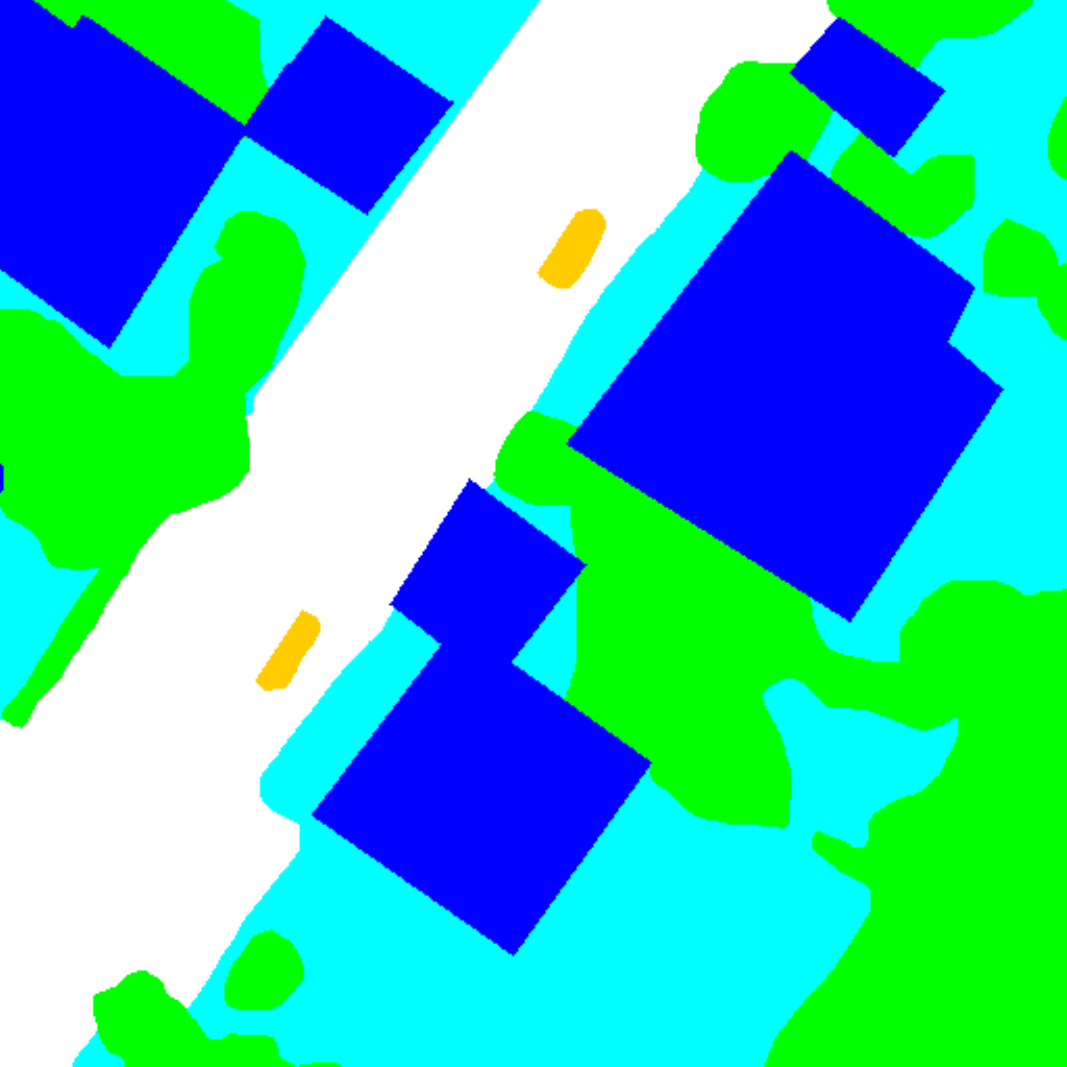}
		
		\vspace{1mm}
		
		\includegraphics[scale=0.094]{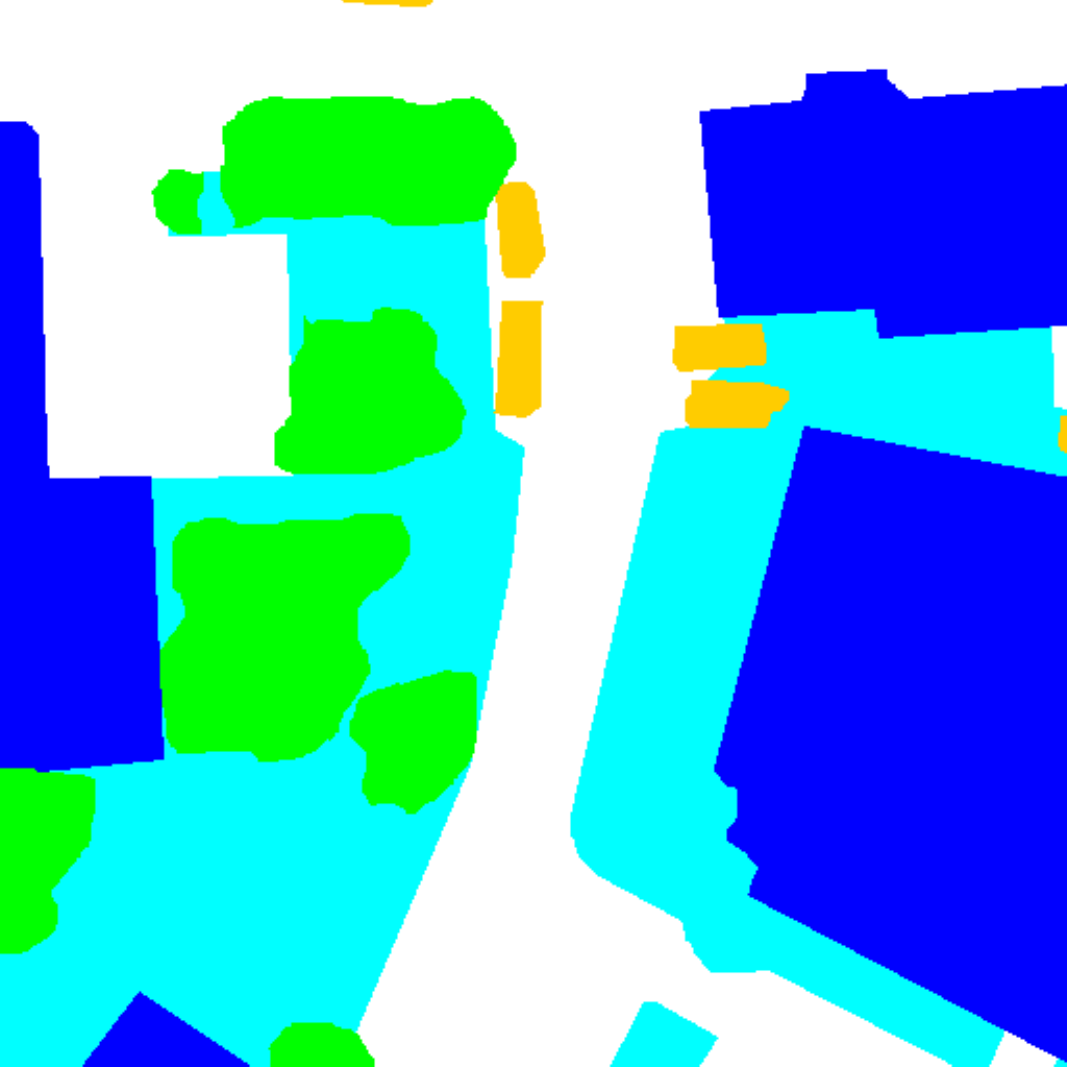}
		
		\vspace{1mm}
		
		\includegraphics[scale=0.094]{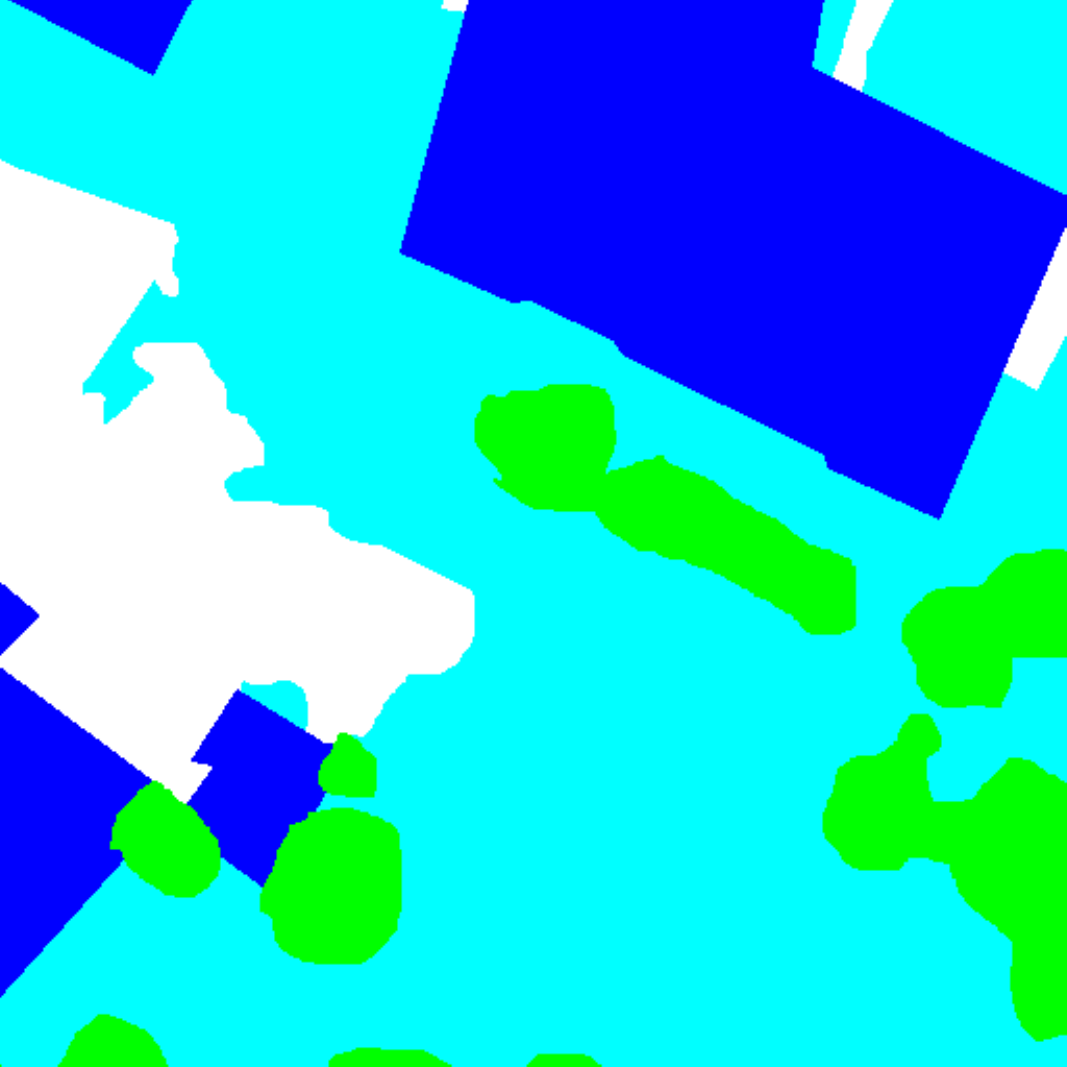}
		
		\vspace{1mm}
		
		\includegraphics[scale=0.094]{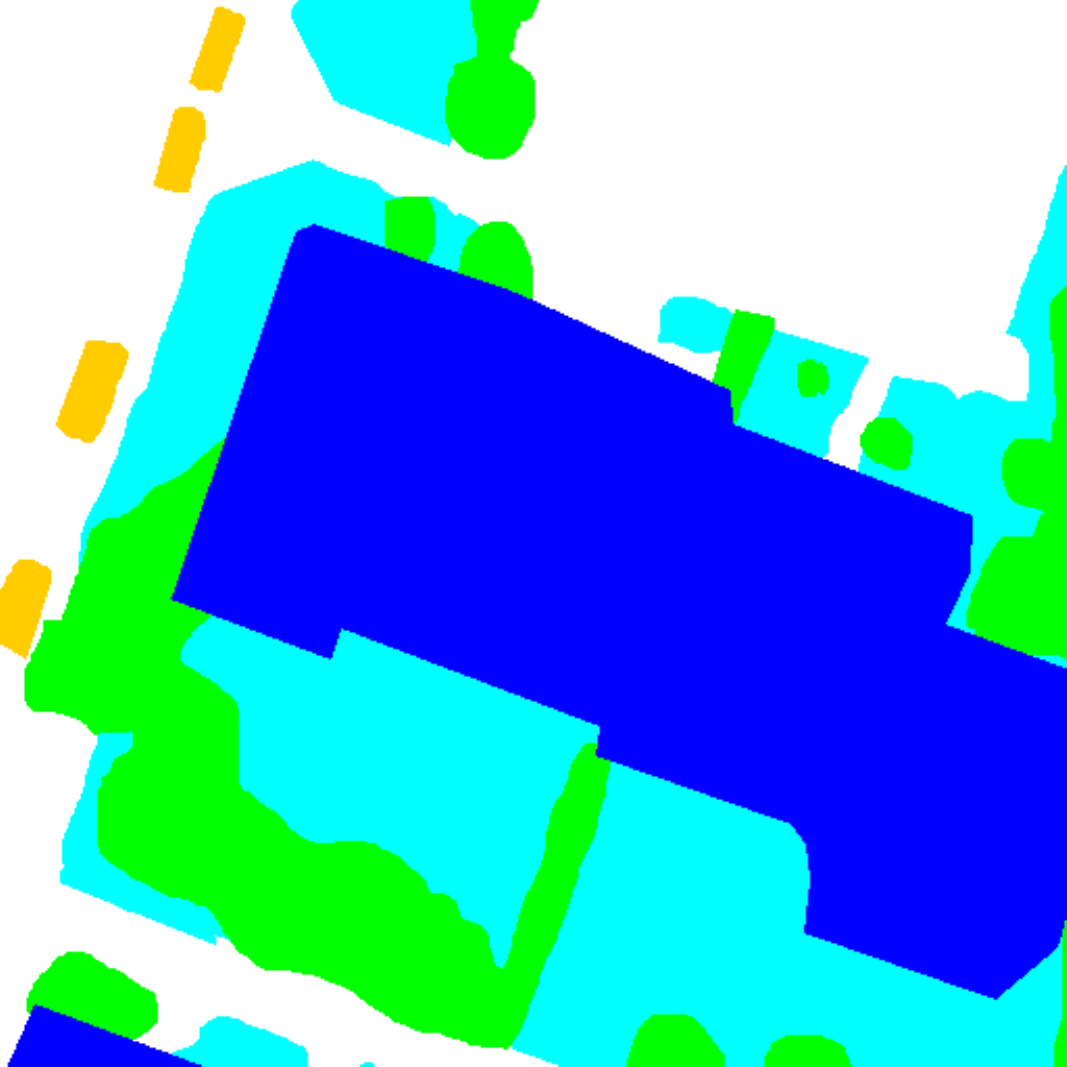}
		
		\vspace{1mm}
		
		\includegraphics[scale=0.094]{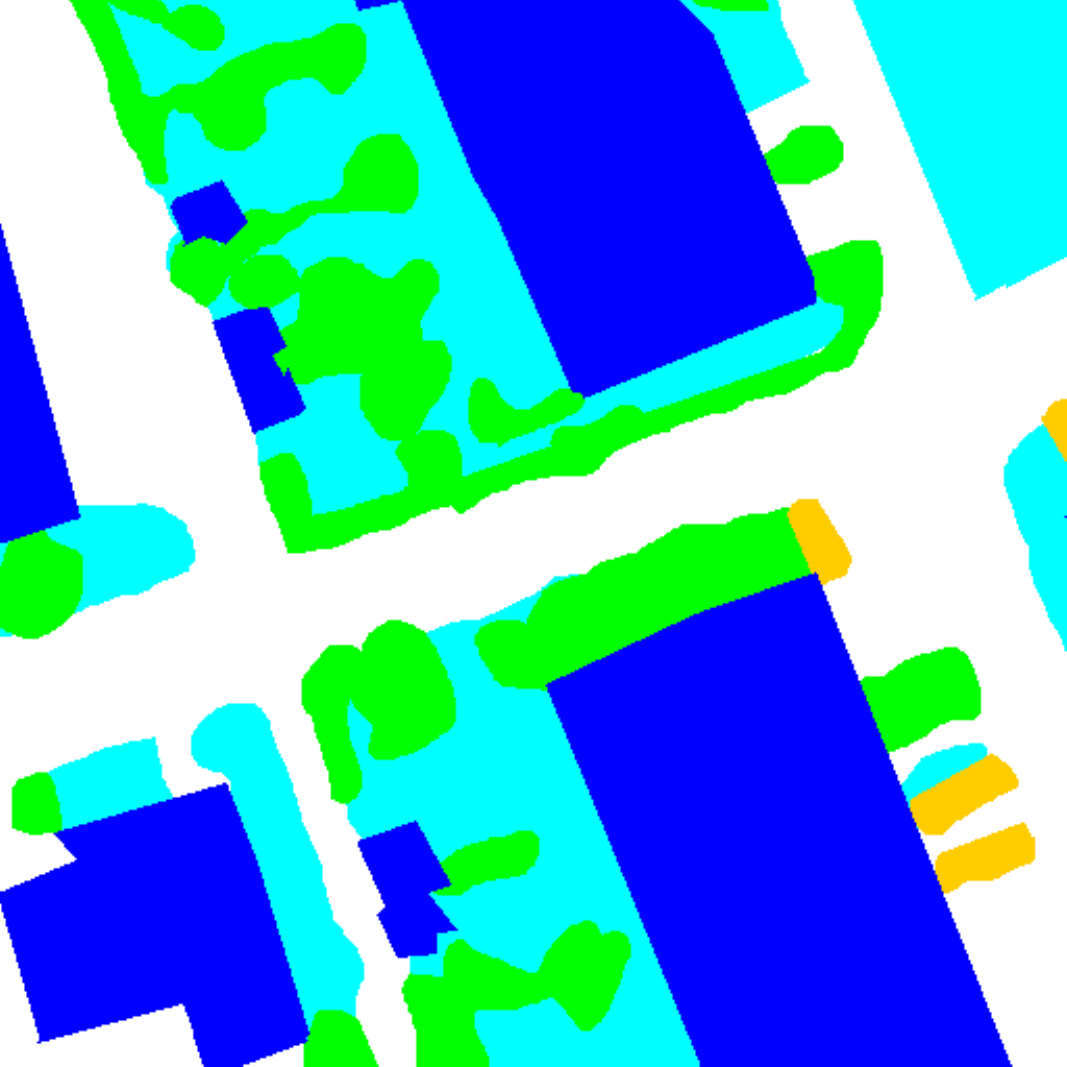}
		\centerline{(b)}
	\end{minipage}
	\begin{minipage}[t]{0.092\linewidth}
		\centering
		\includegraphics[scale=0.094]{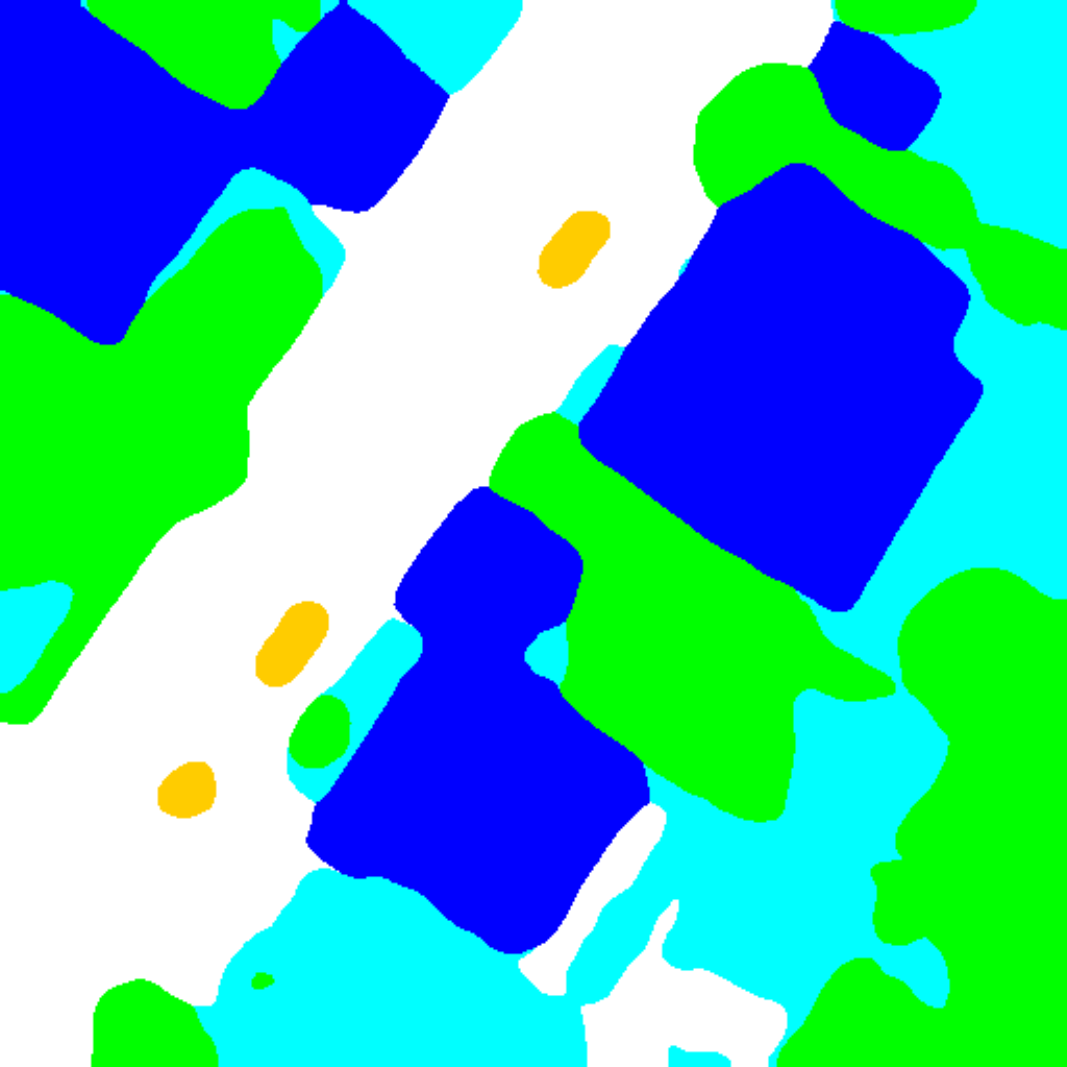}
		
		\vspace{1mm}
		
		\includegraphics[scale=0.094]{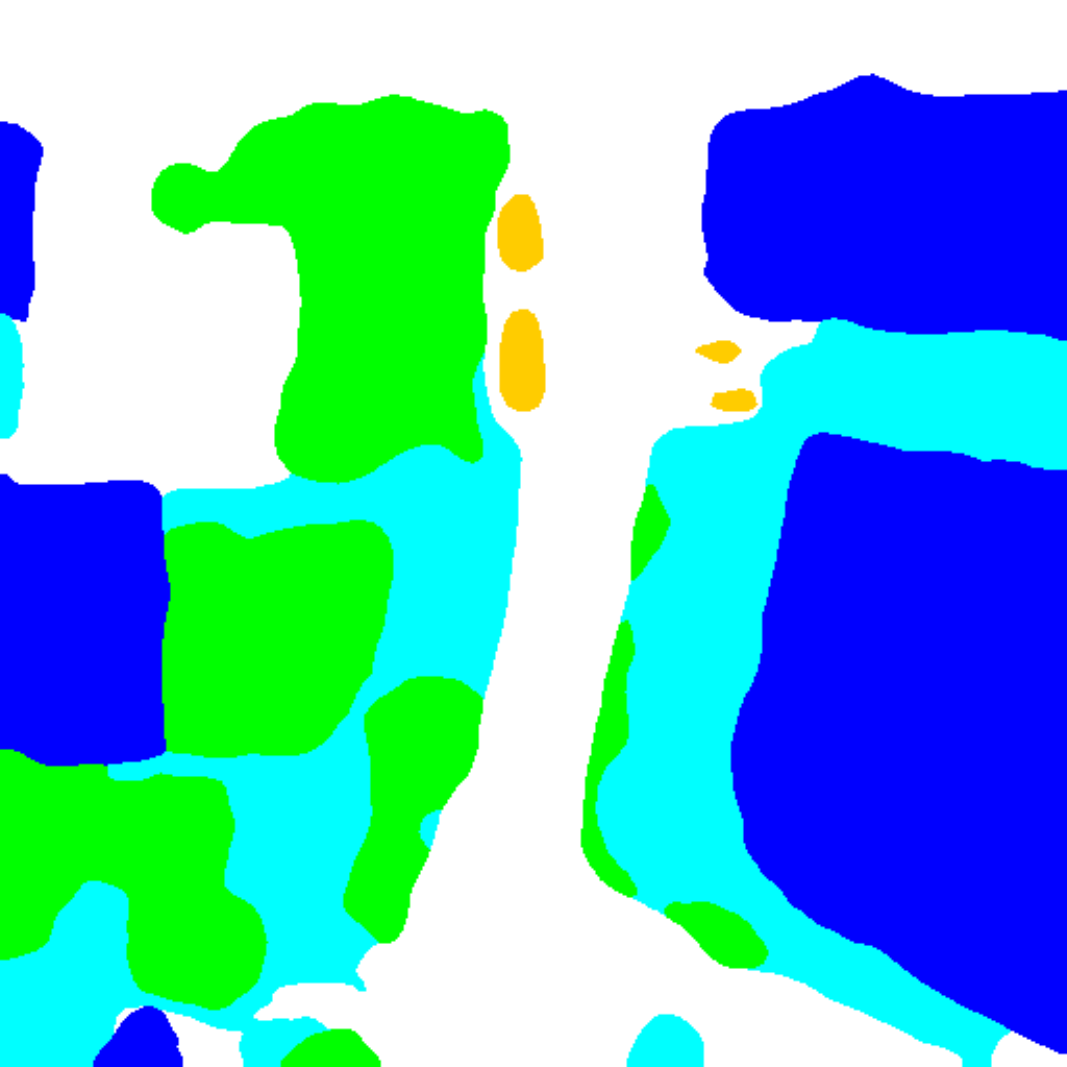}
		
		\vspace{1mm}
		
		\includegraphics[scale=0.094]{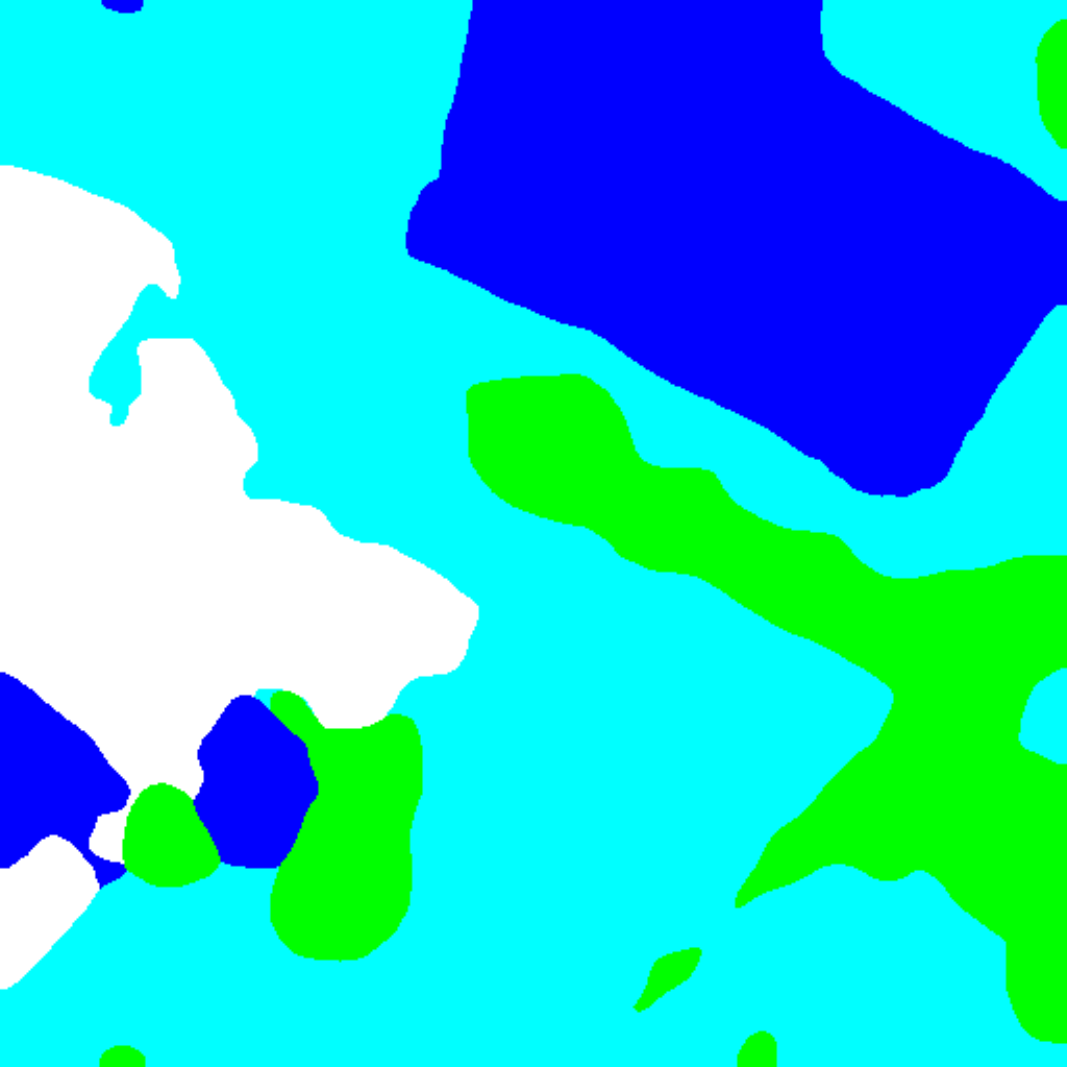}
		
		\vspace{1mm}
		
		\includegraphics[scale=0.094]{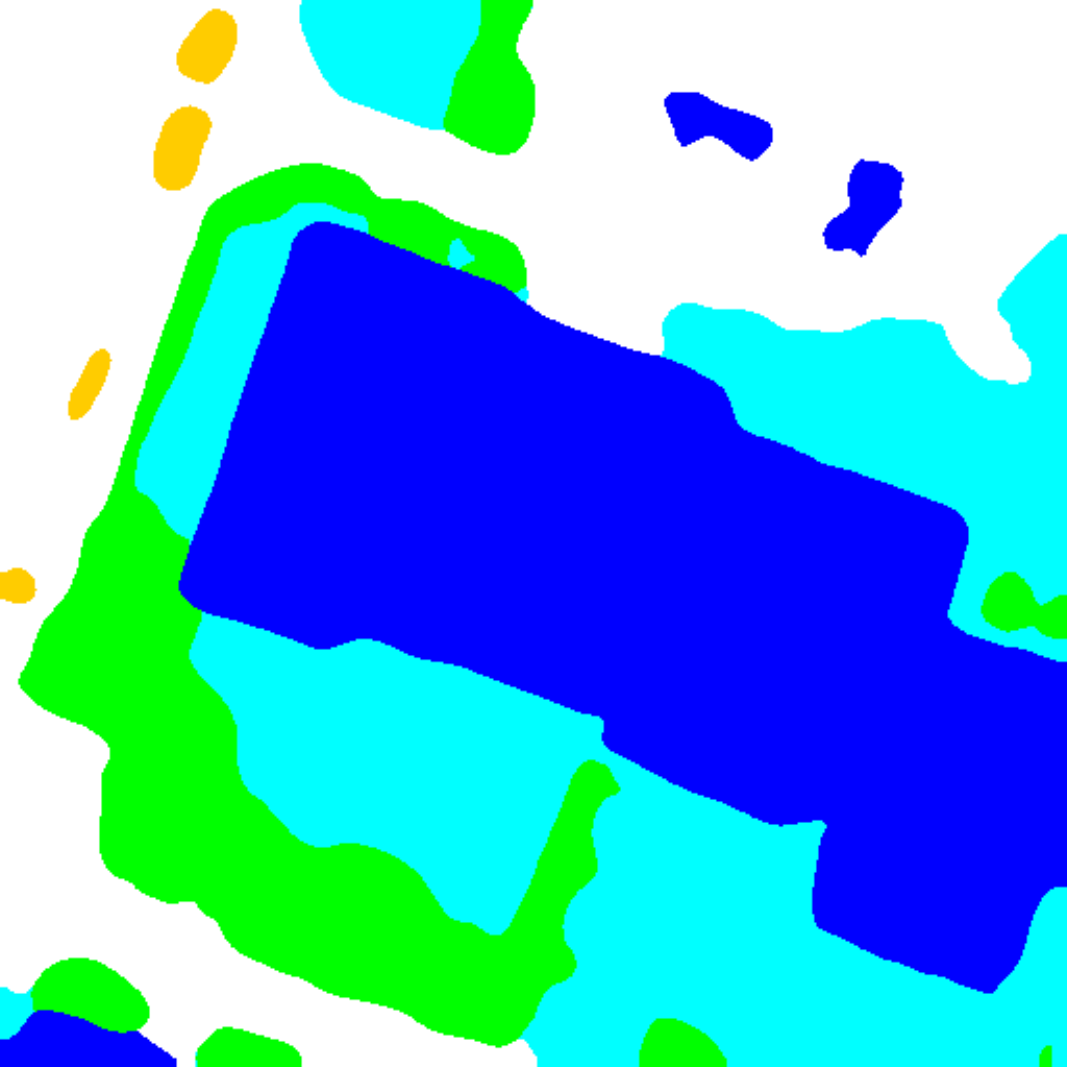}
		
		\vspace{1mm}
		
		\includegraphics[scale=0.094]{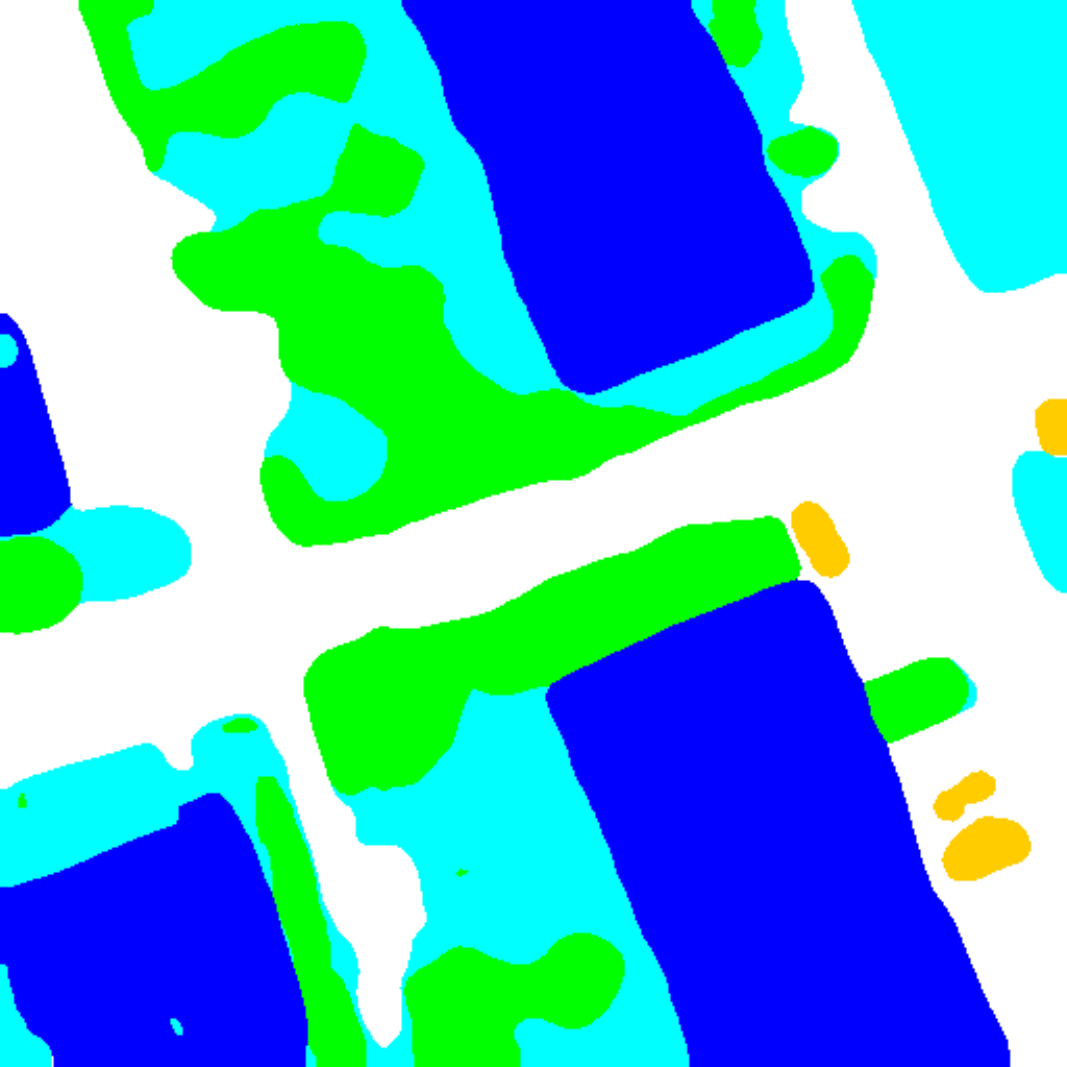}
		\centerline{(c)}
	\end{minipage}
	\begin{minipage}[t]{0.092\linewidth}
		\centering
		\includegraphics[scale=0.094]{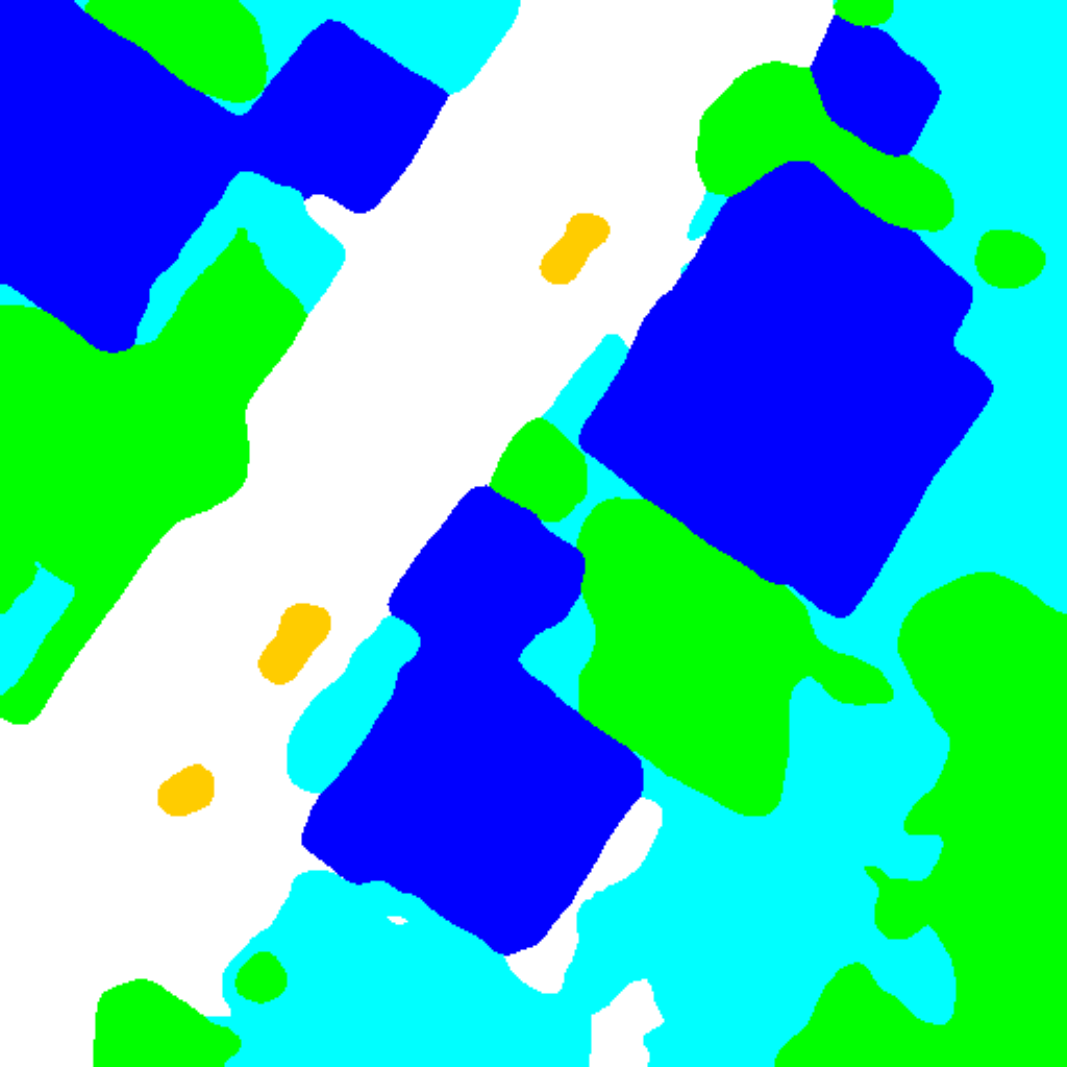}
		
		\vspace{1mm}
		
		\includegraphics[scale=0.094]{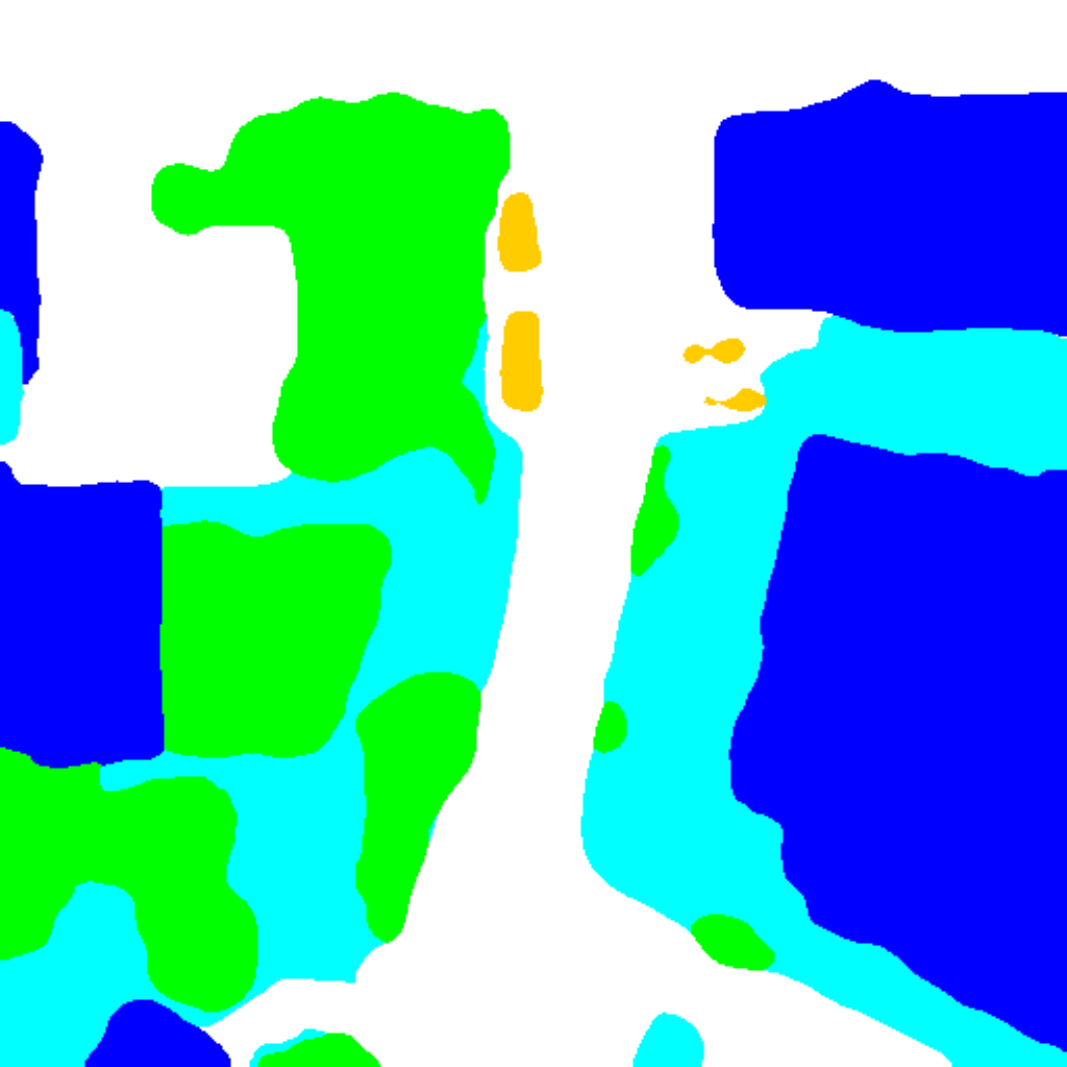}
		
		\vspace{1mm}
		
		\includegraphics[scale=0.094]{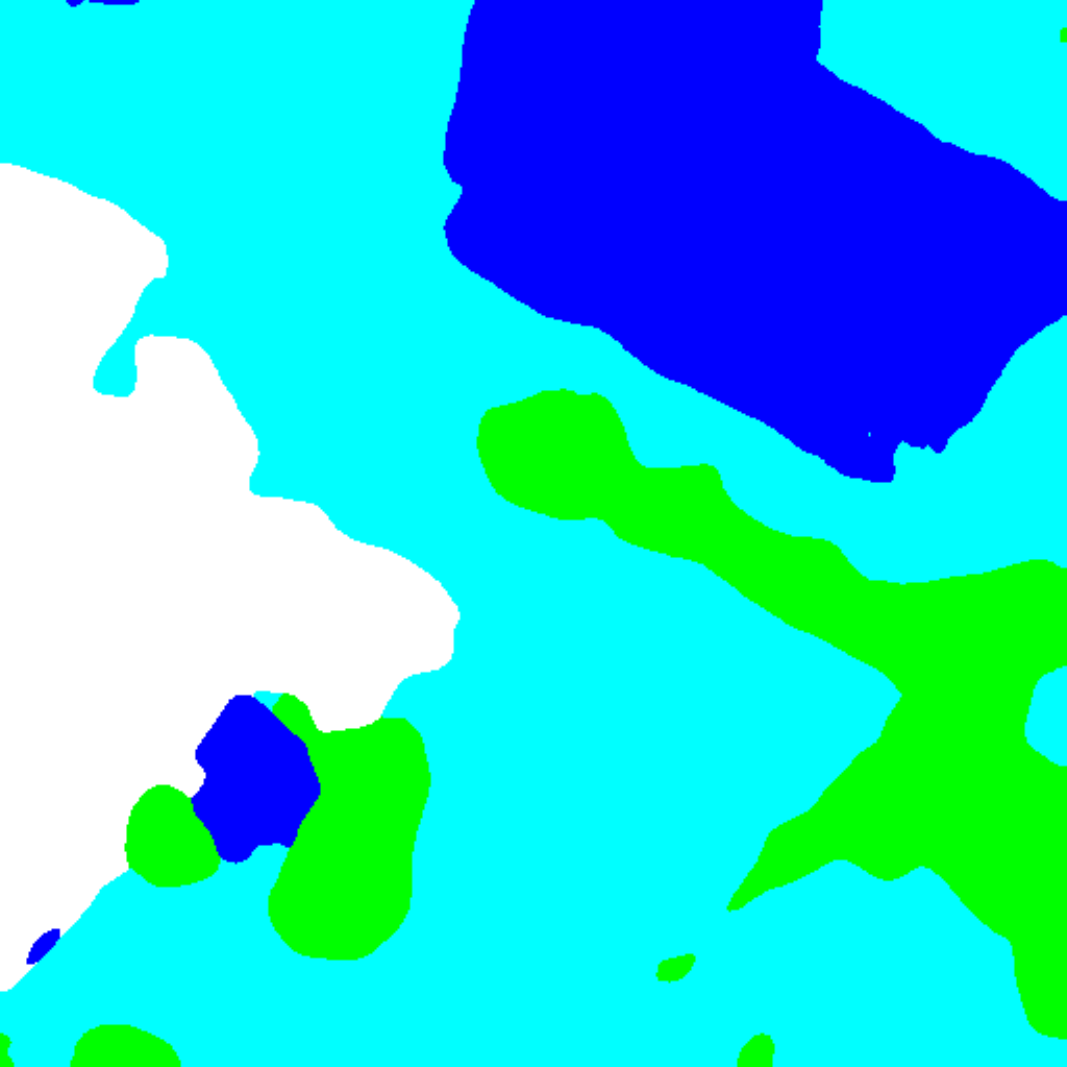}
		
		\vspace{1mm}
		
		\includegraphics[scale=0.094]{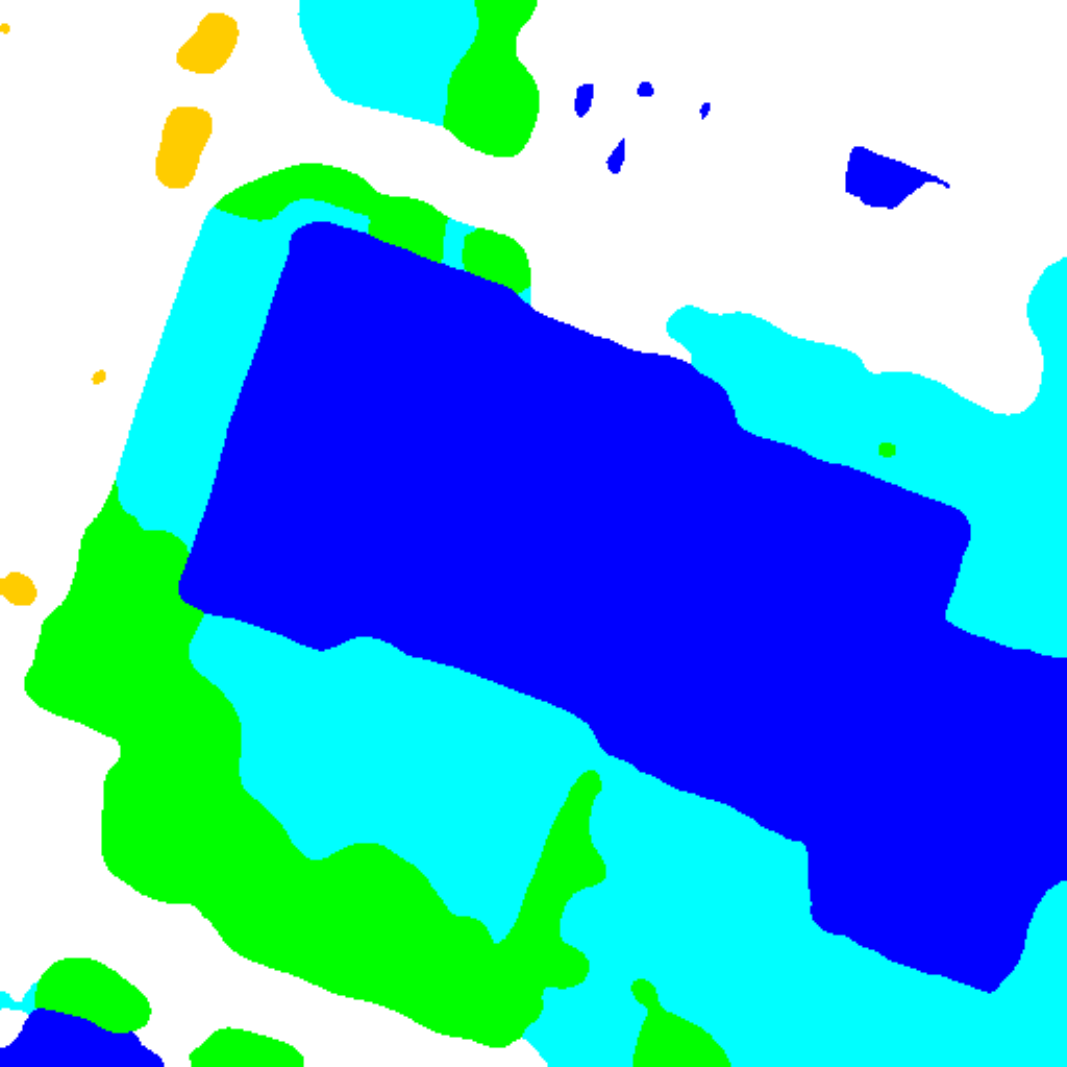}
		
		\vspace{1mm}
		
		\includegraphics[scale=0.094]{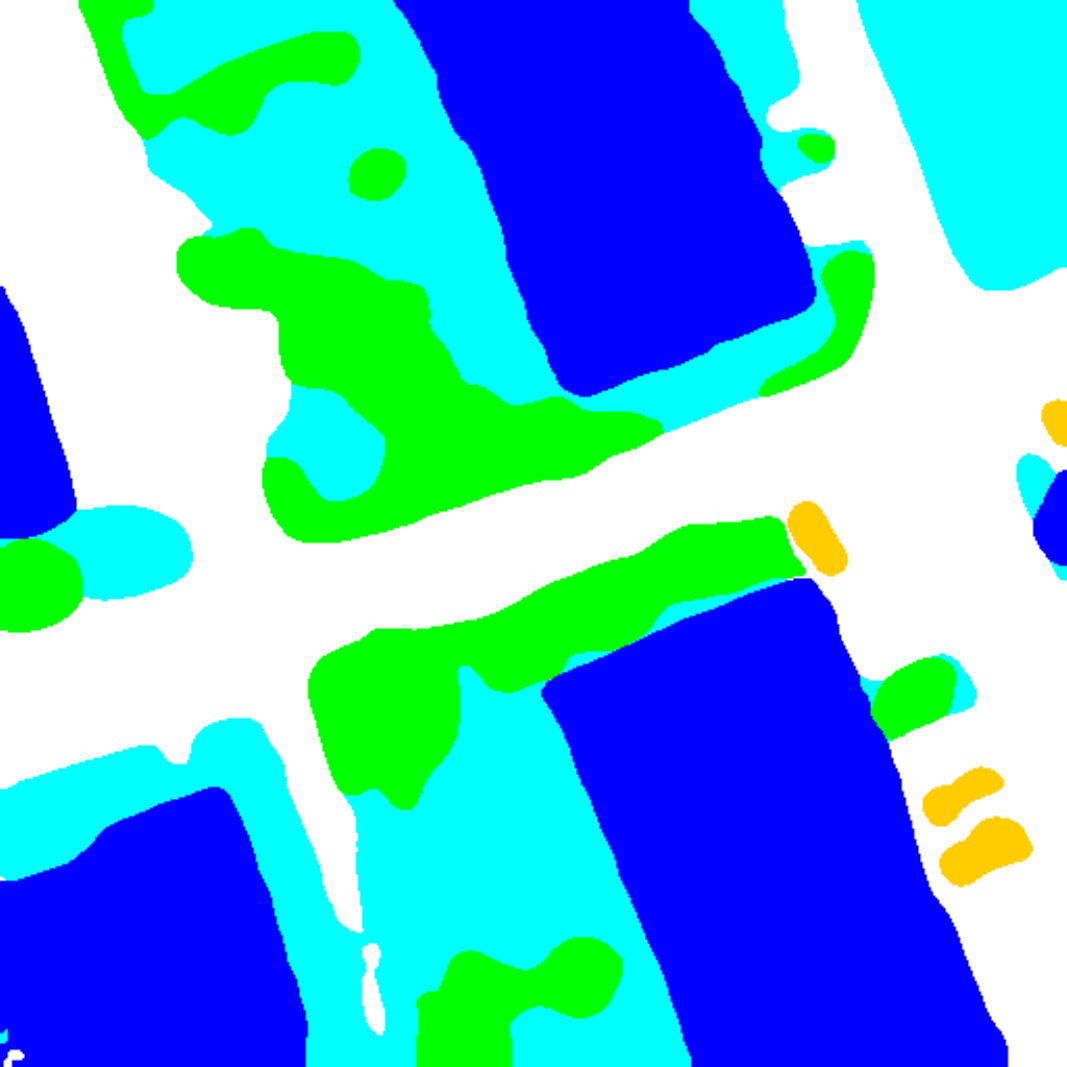}
		\centerline{(d)}
	\end{minipage}
	\begin{minipage}[t]{0.092\linewidth}
		\centering
		\includegraphics[scale=0.094]{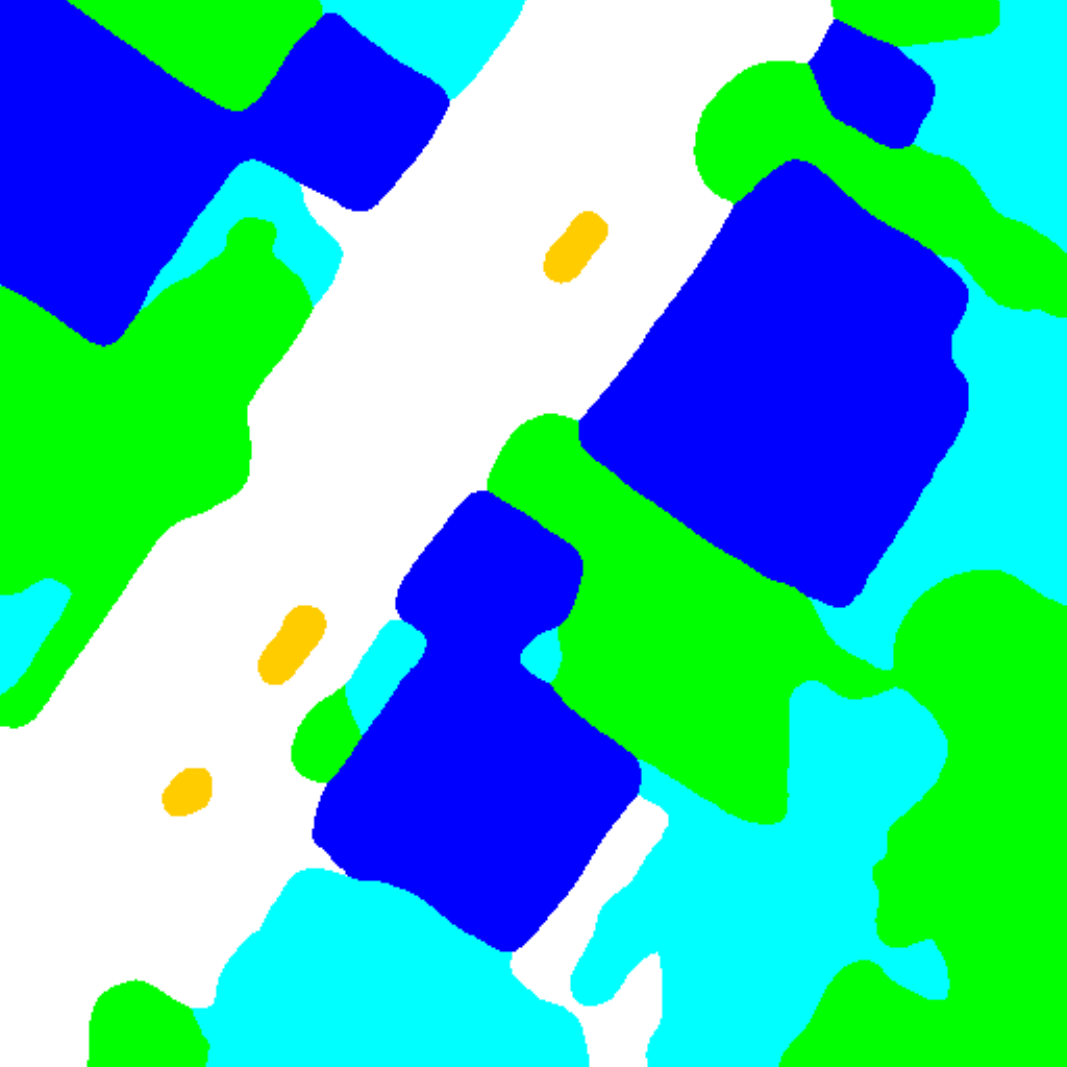}
		
		\vspace{1mm}
		
		\includegraphics[scale=0.094]{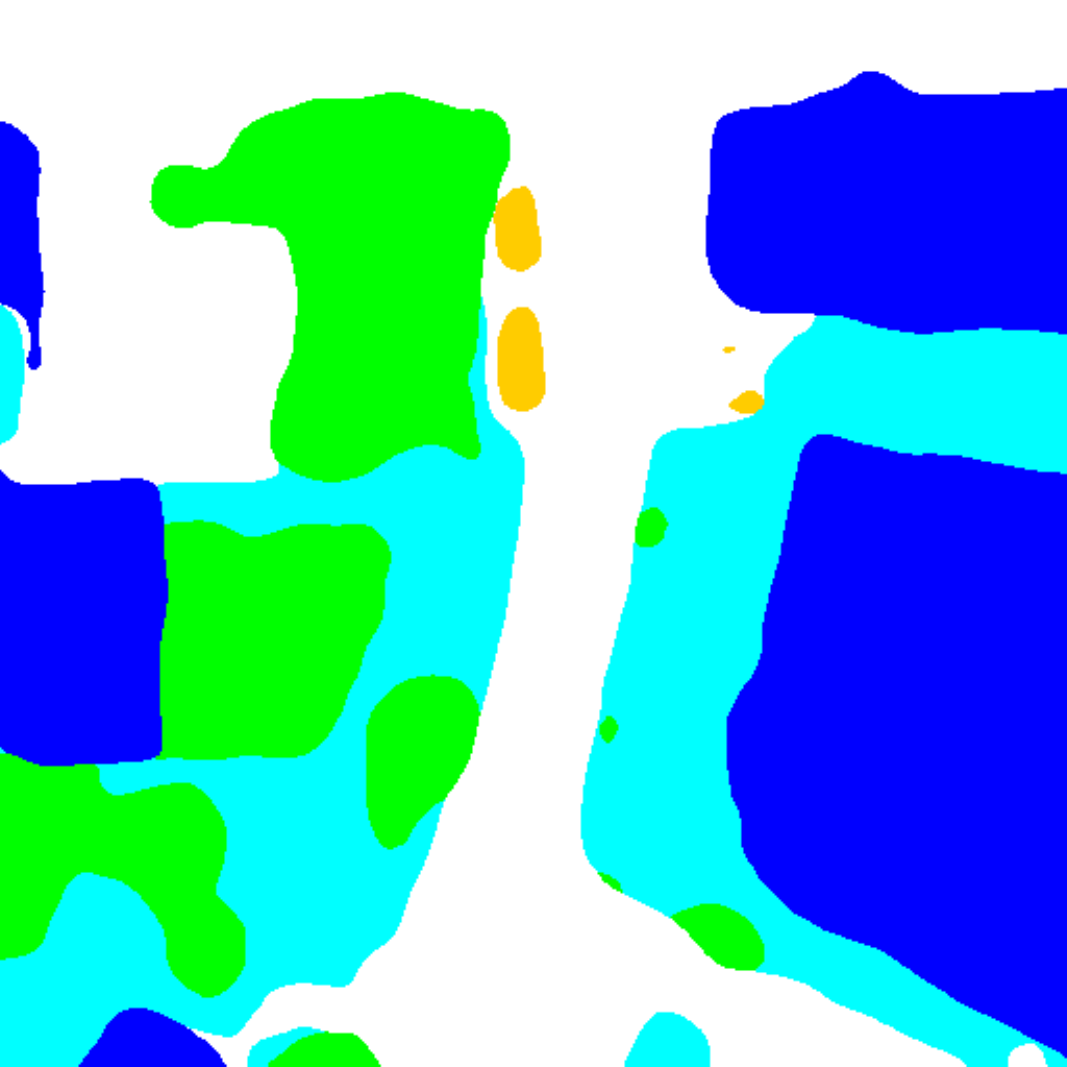}
		
		\vspace{1mm}
		
		\includegraphics[scale=0.094]{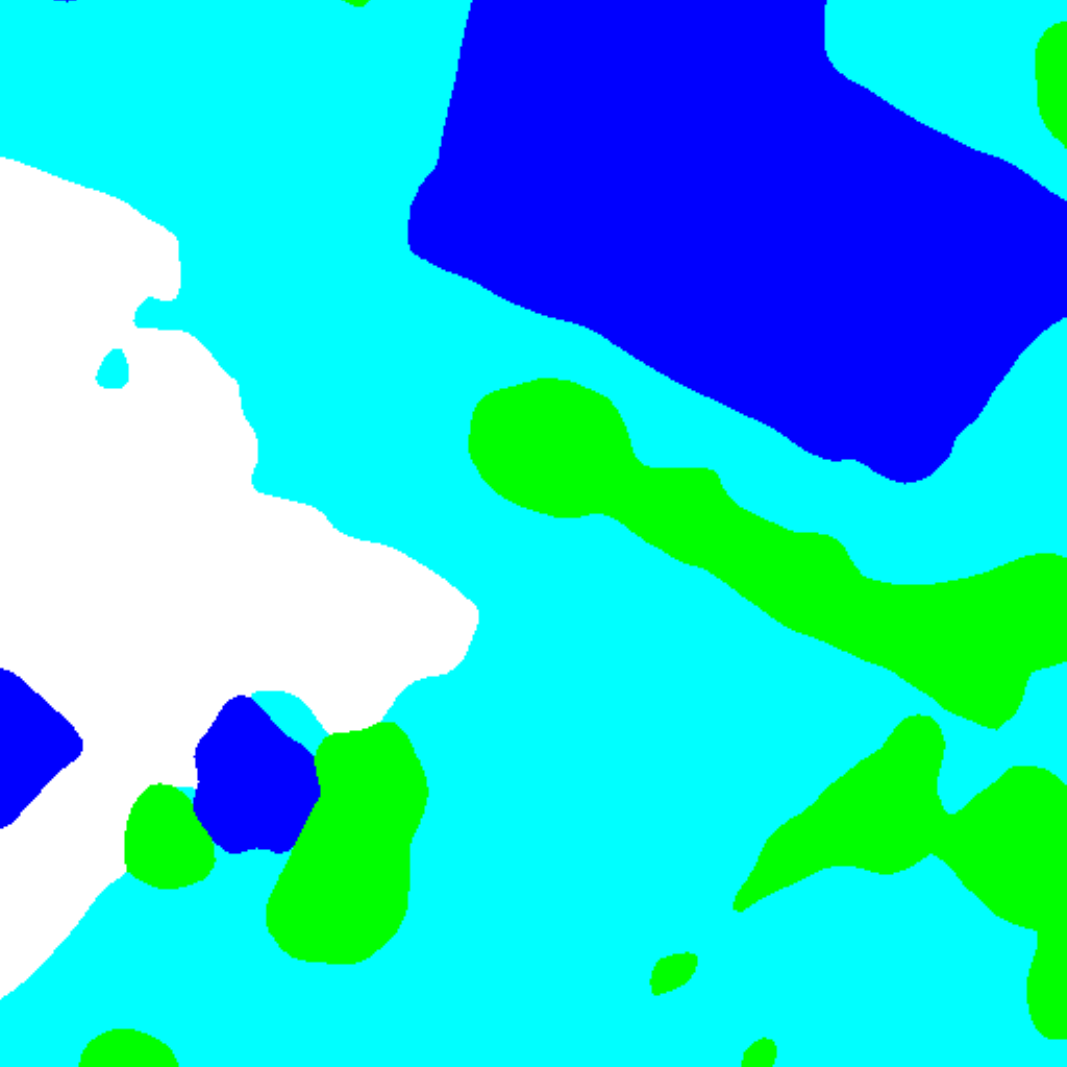}
		
		\vspace{1mm}
		
		\includegraphics[scale=0.094]{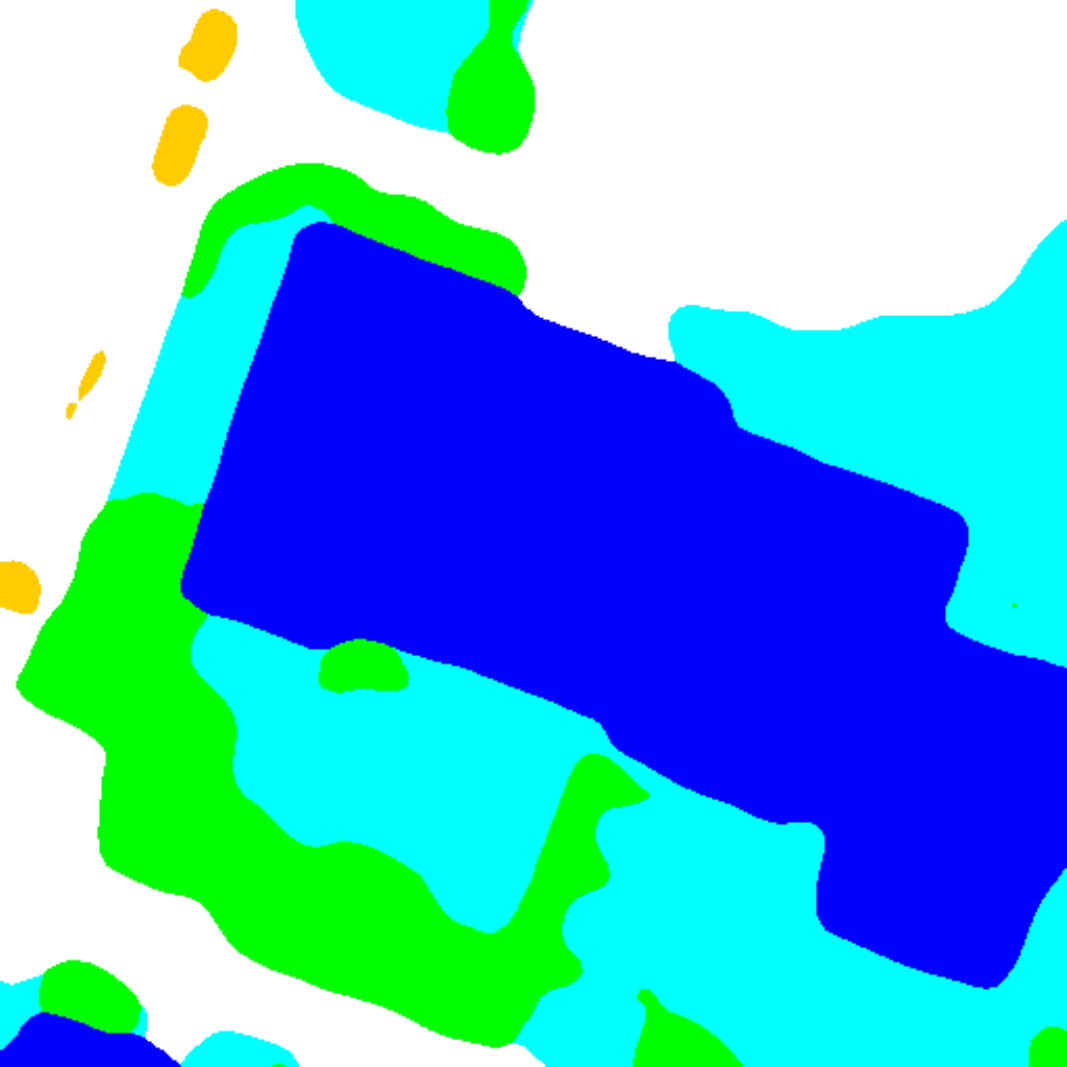}
		
		\vspace{1mm}
		
		\includegraphics[scale=0.094]{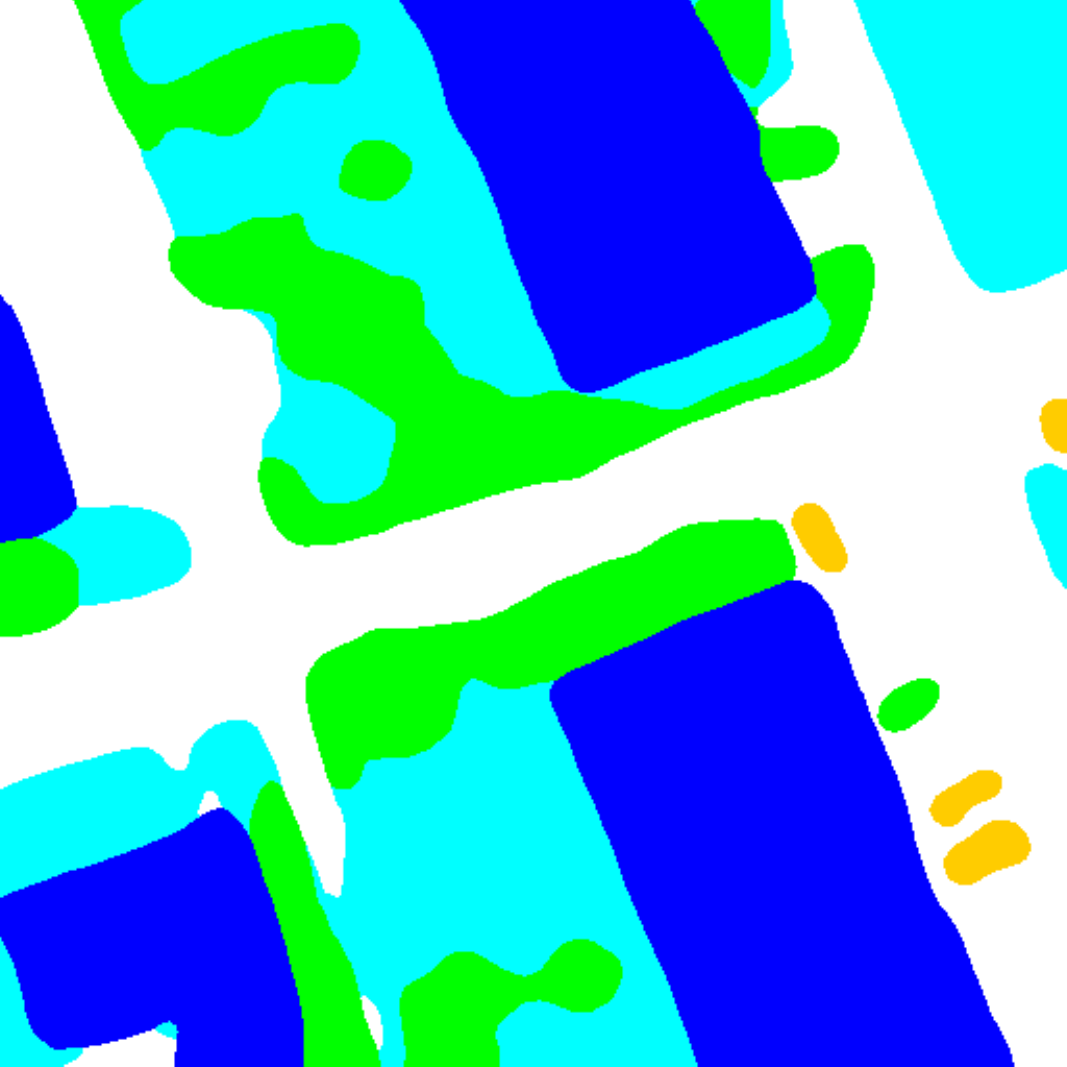}
		\centerline{(e)}
	\end{minipage}
	\begin{minipage}[t]{0.092\linewidth}
		\centering
		\includegraphics[scale=0.094]{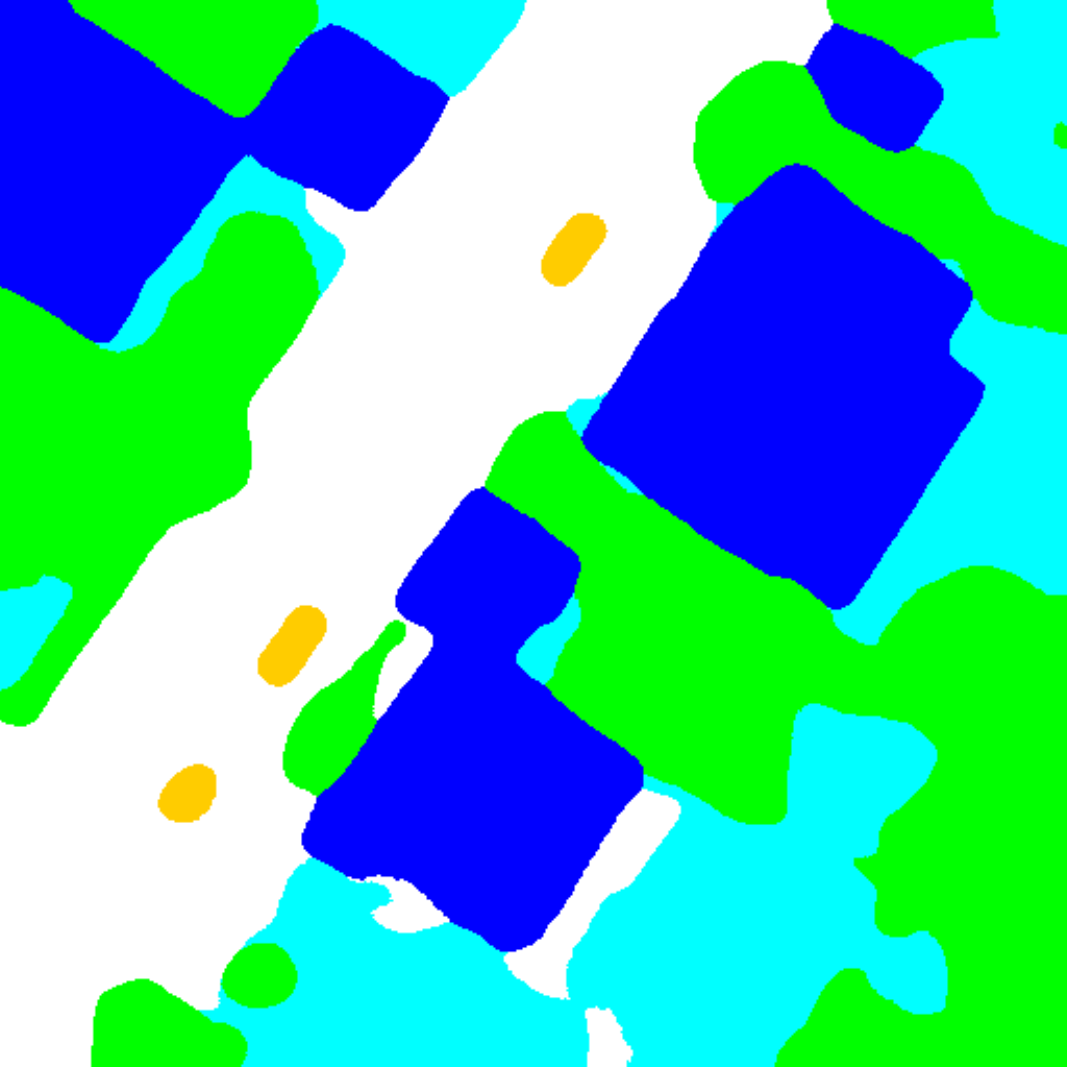}
		
		\vspace{1mm}
		
		\includegraphics[scale=0.094]{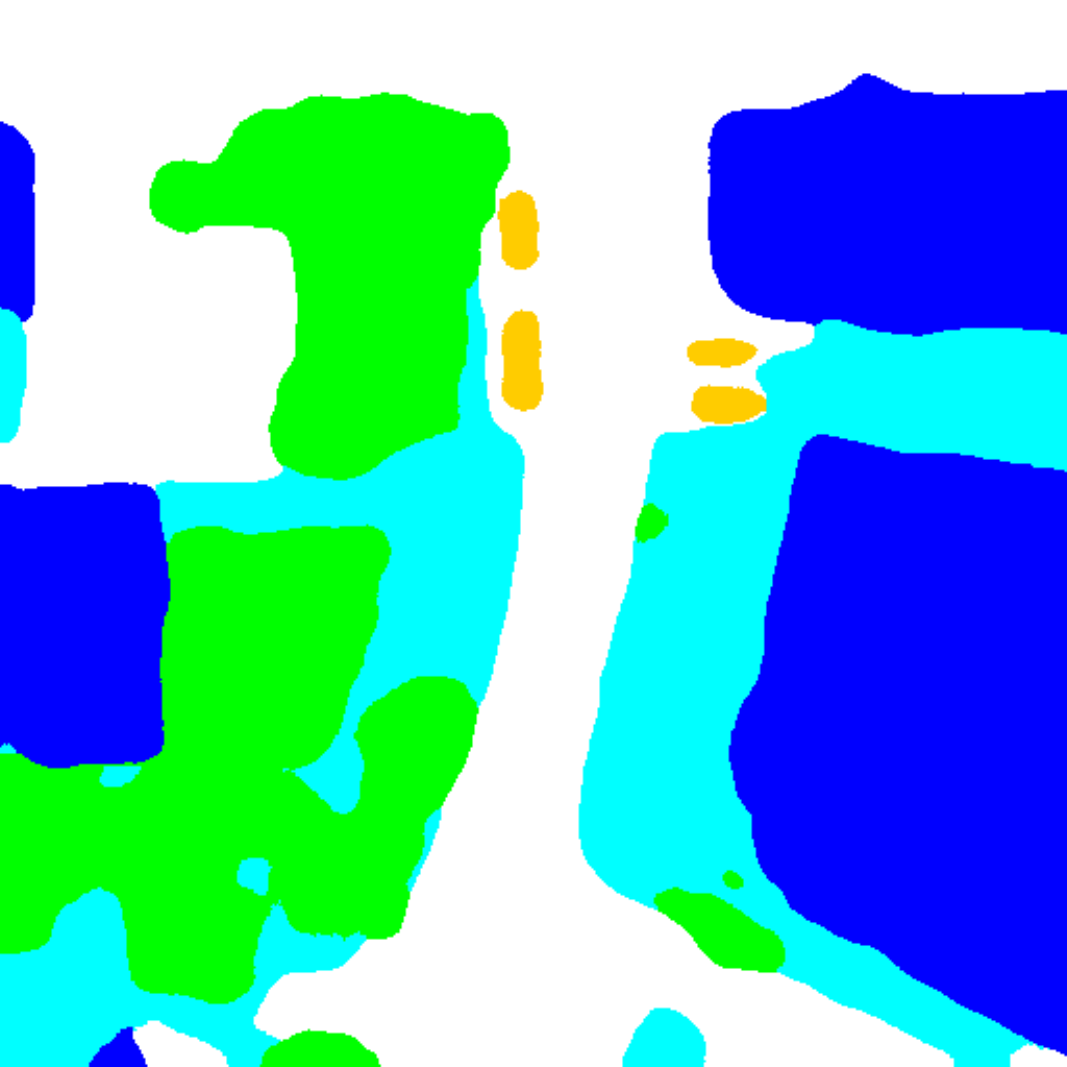}
		
		\vspace{1mm}
		
		\includegraphics[scale=0.094]{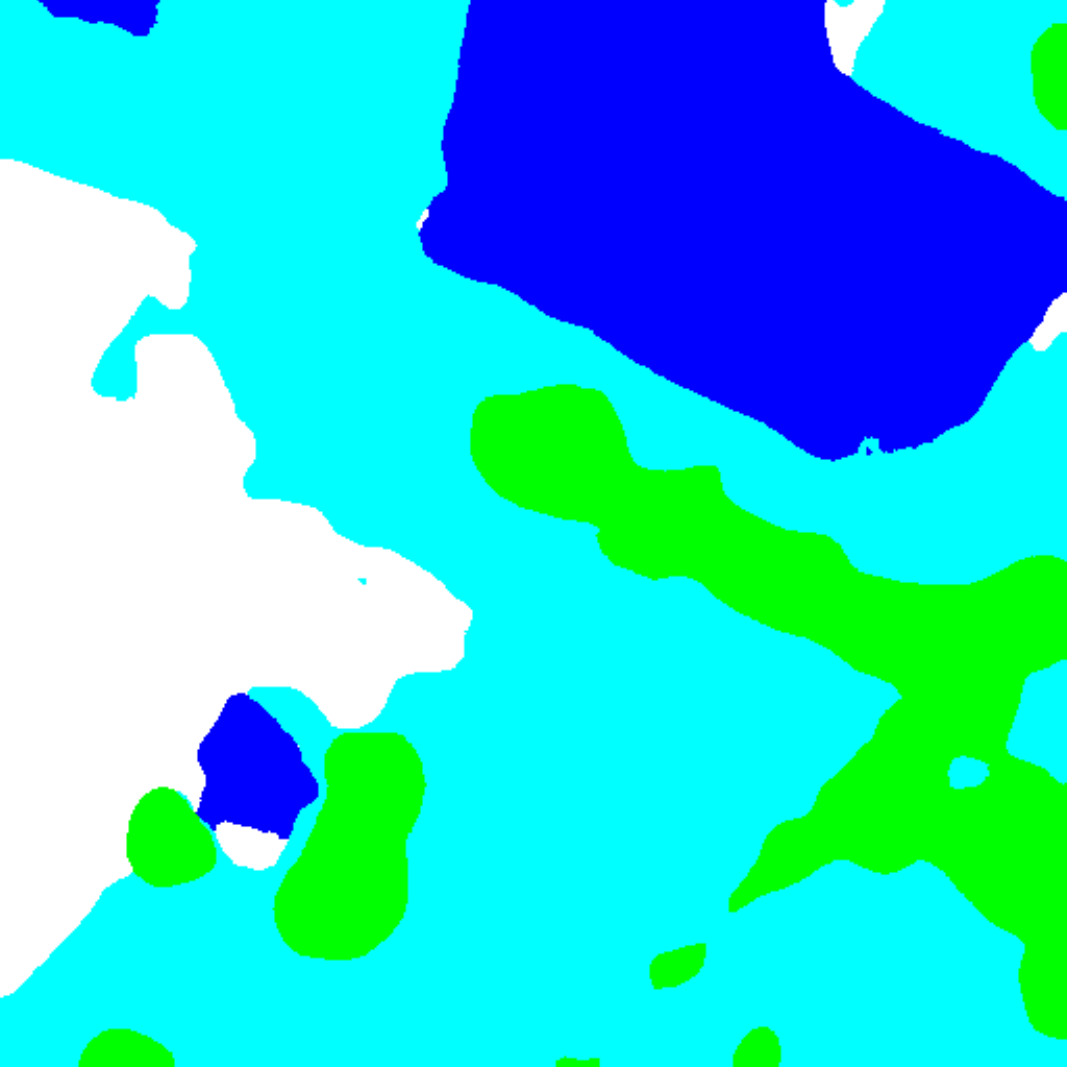}
		
		\vspace{1mm}
		
		\includegraphics[scale=0.094]{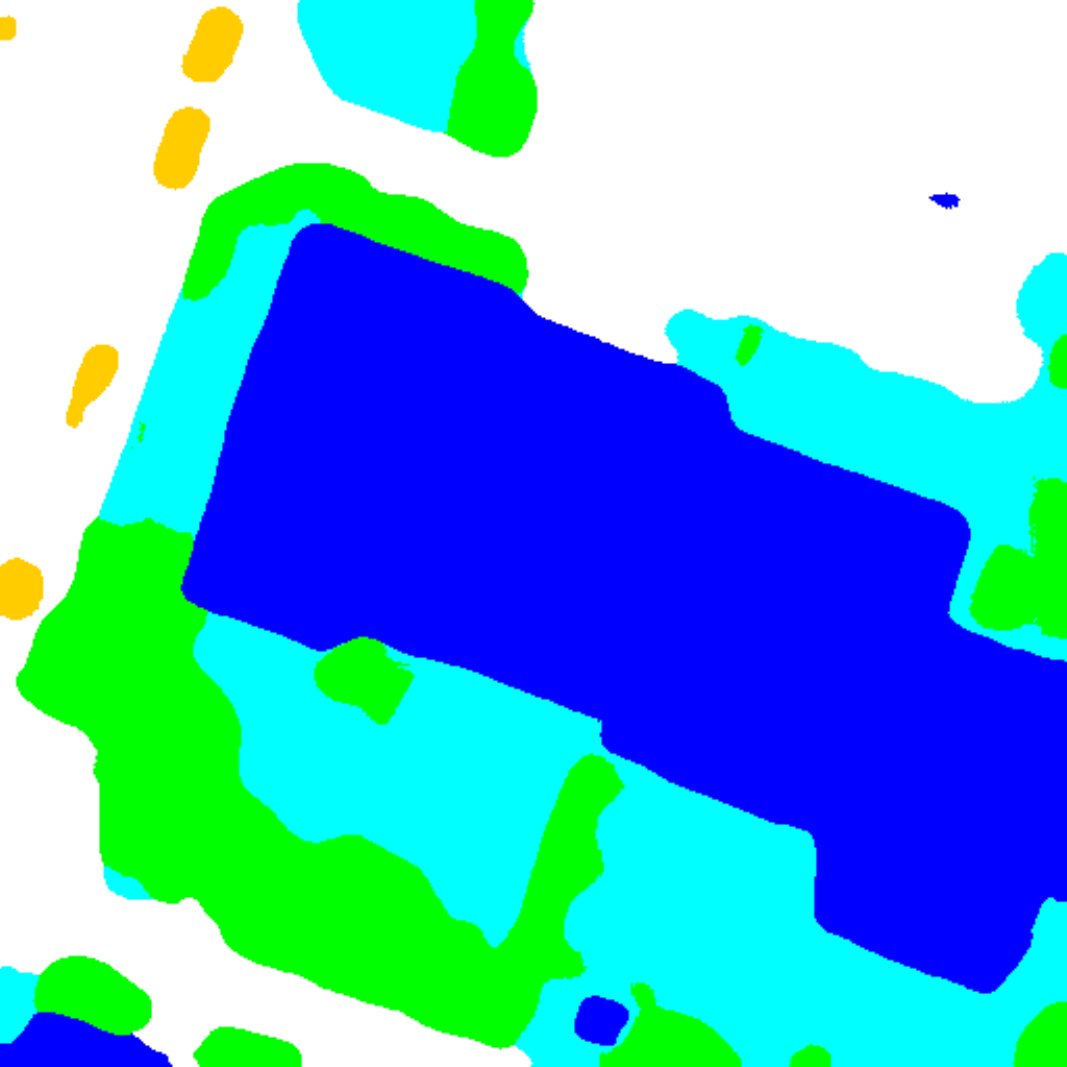}
		
		\vspace{1mm}
		
		\includegraphics[scale=0.094]{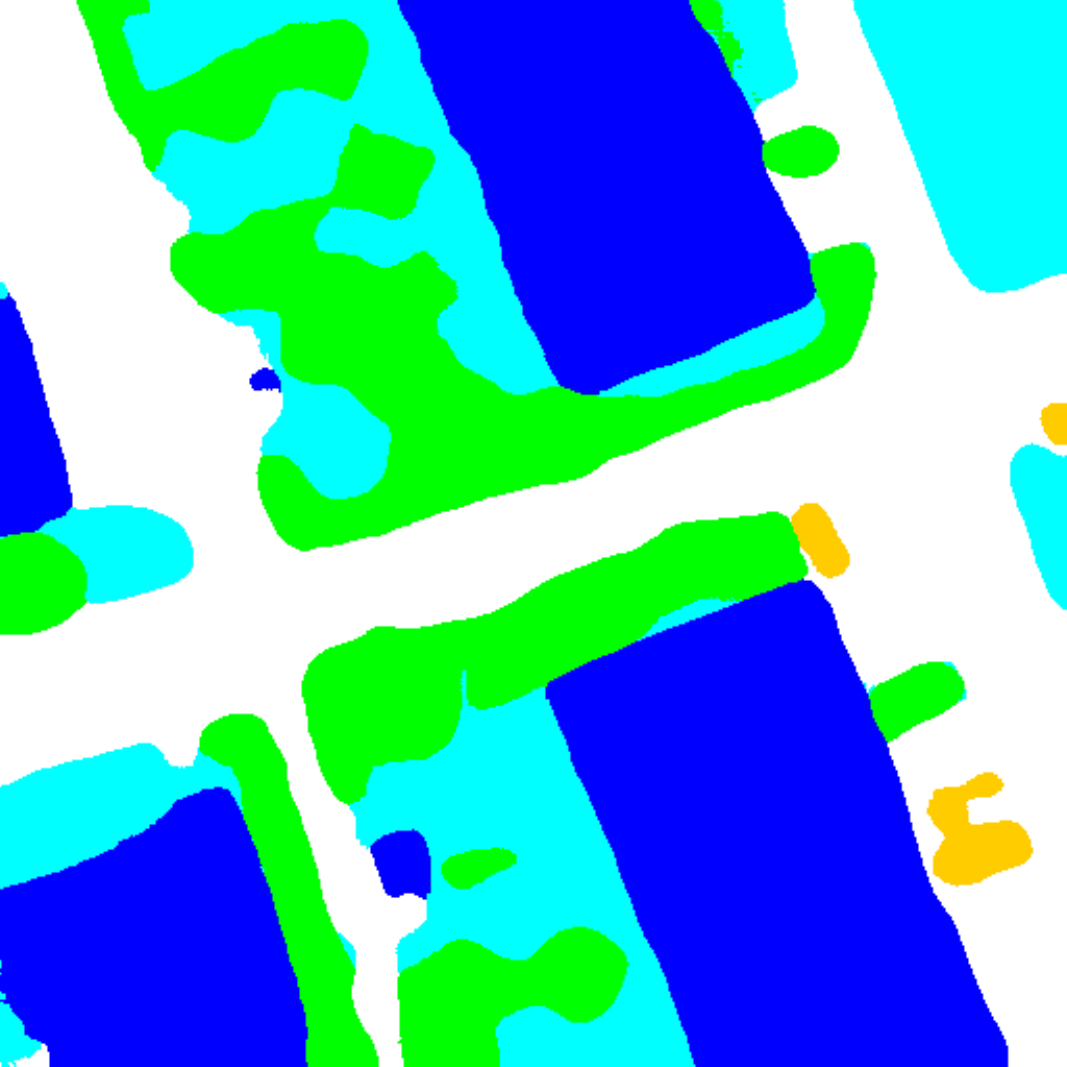}
		\centerline{(f)}
	\end{minipage}
	\begin{minipage}[t]{0.092\linewidth}
		\centering
		\includegraphics[scale=0.094]{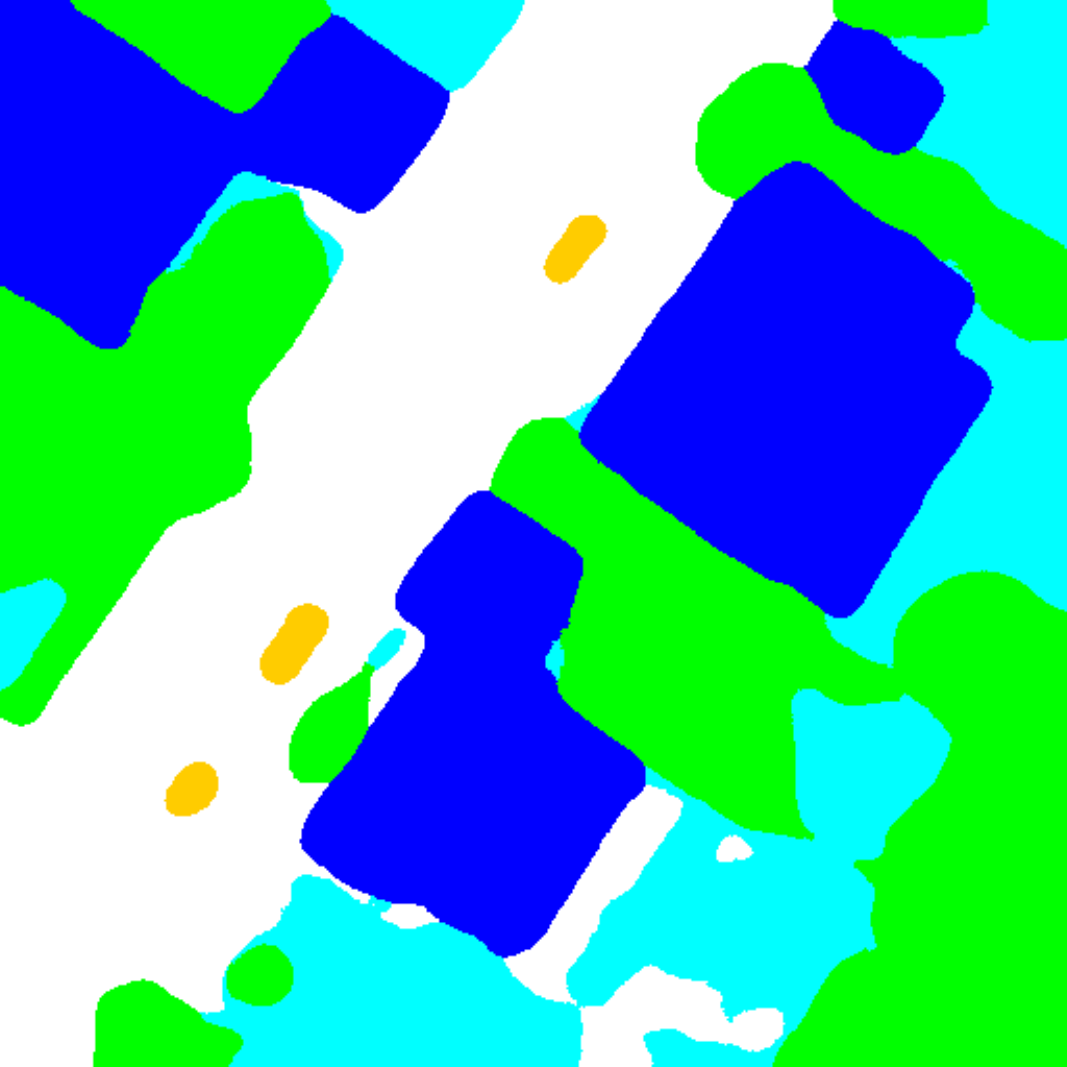}
		
		\vspace{1mm}
		
		\includegraphics[scale=0.094]{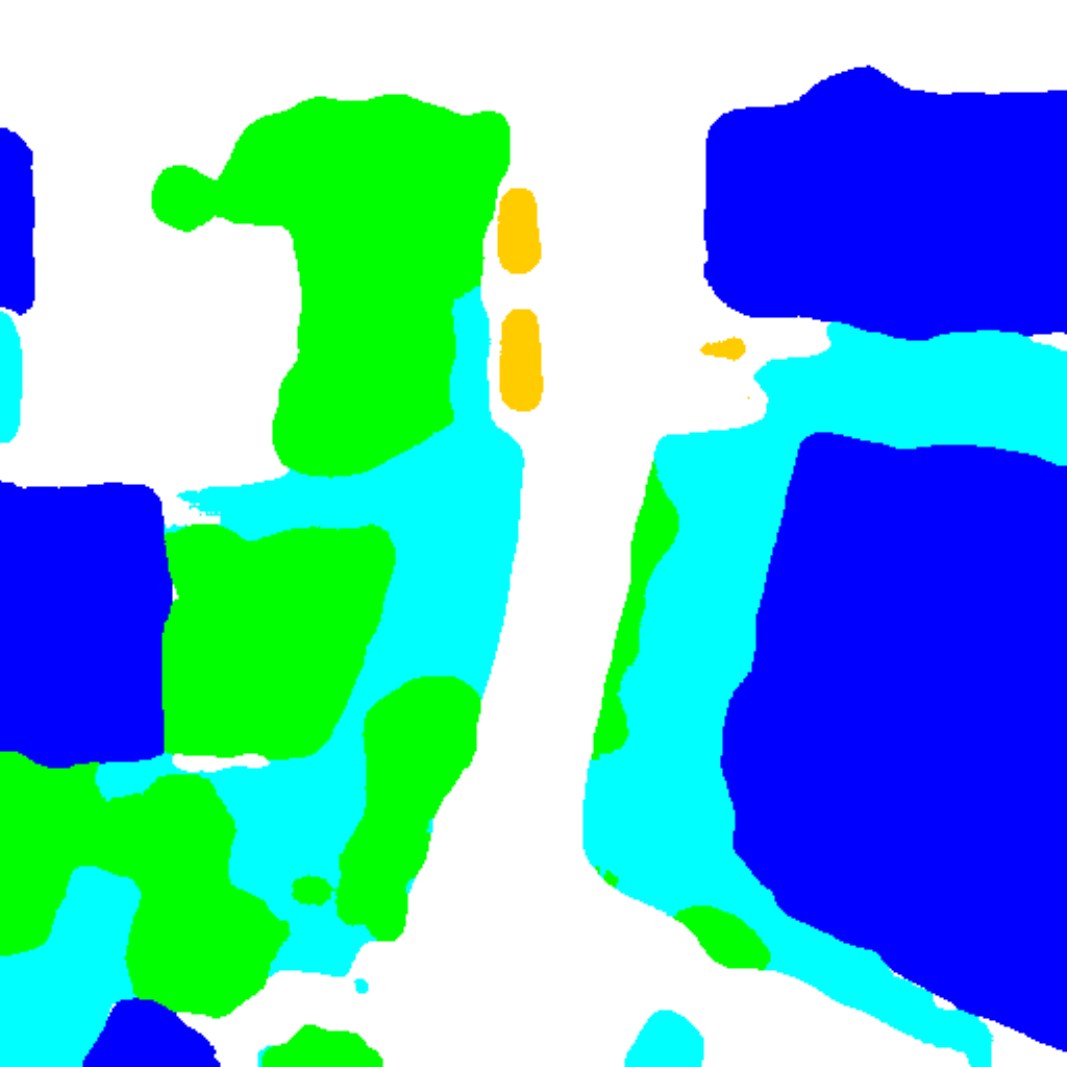}
		
		\vspace{1mm}
		
		\includegraphics[scale=0.094]{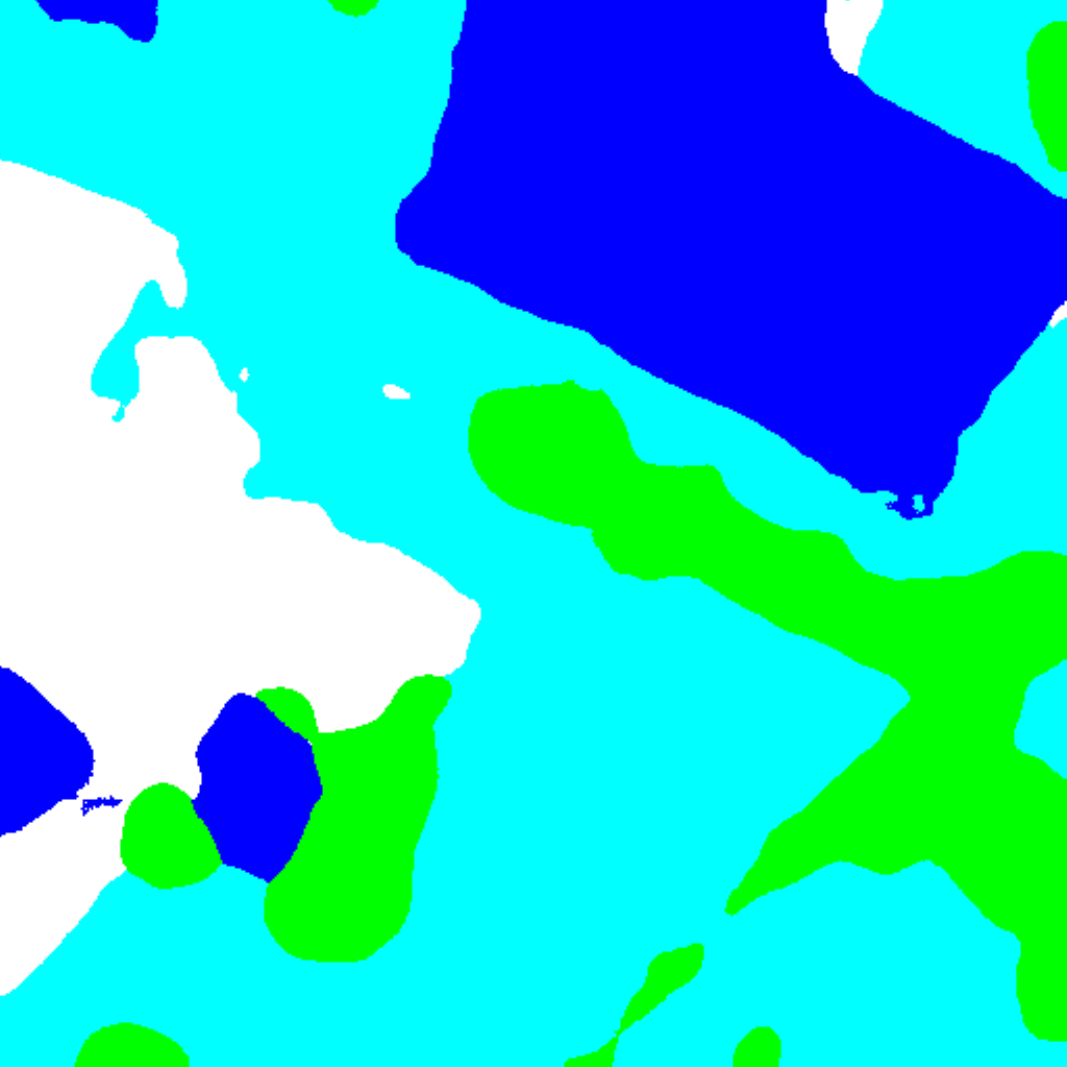}
		
		\vspace{1mm}
		
		\includegraphics[scale=0.094]{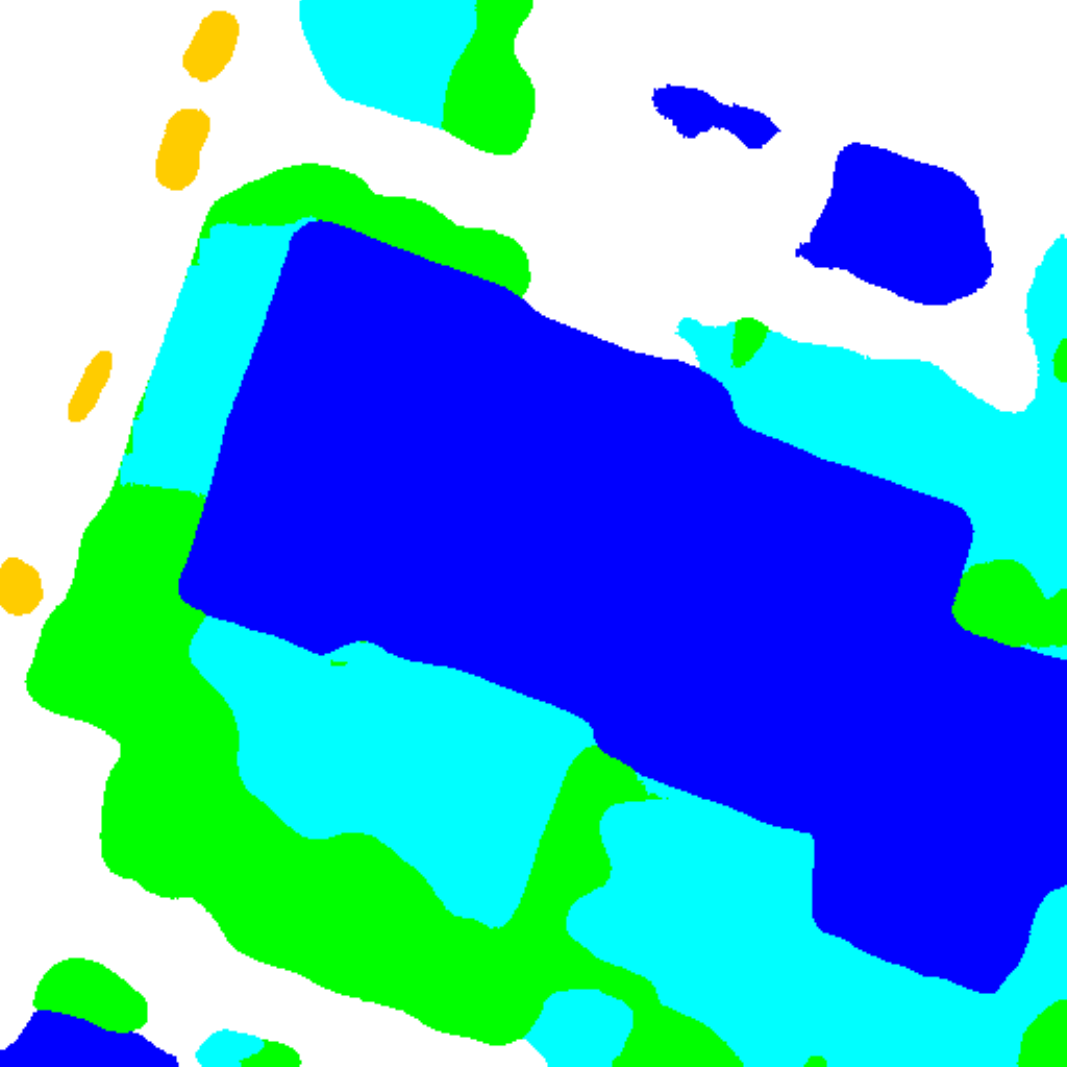}
		
		\vspace{1mm}
		
		\includegraphics[scale=0.094]{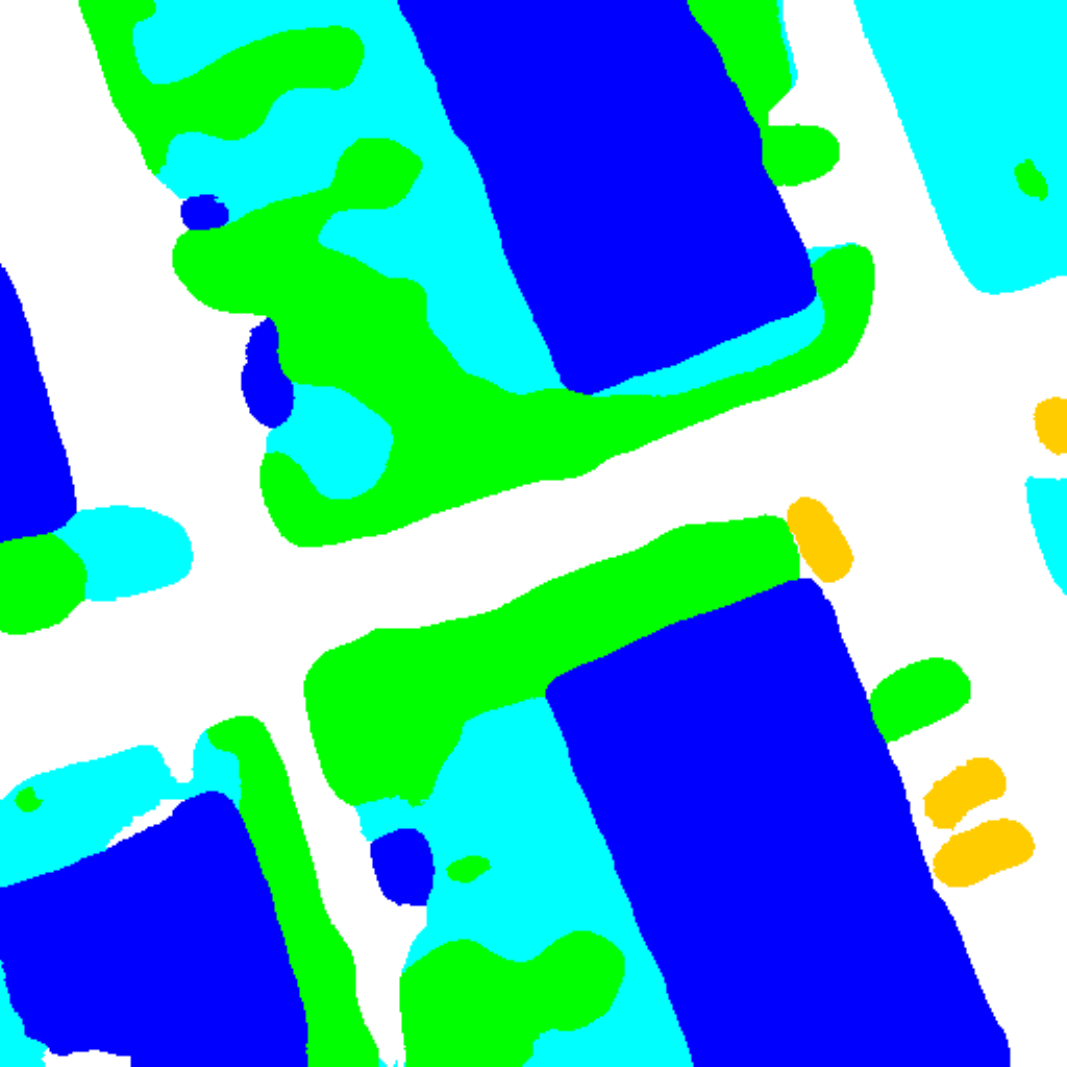}
		\centerline{(g)}
	\end{minipage}
	\begin{minipage}[t]{0.092\linewidth}
		\centering
		\includegraphics[scale=0.094]{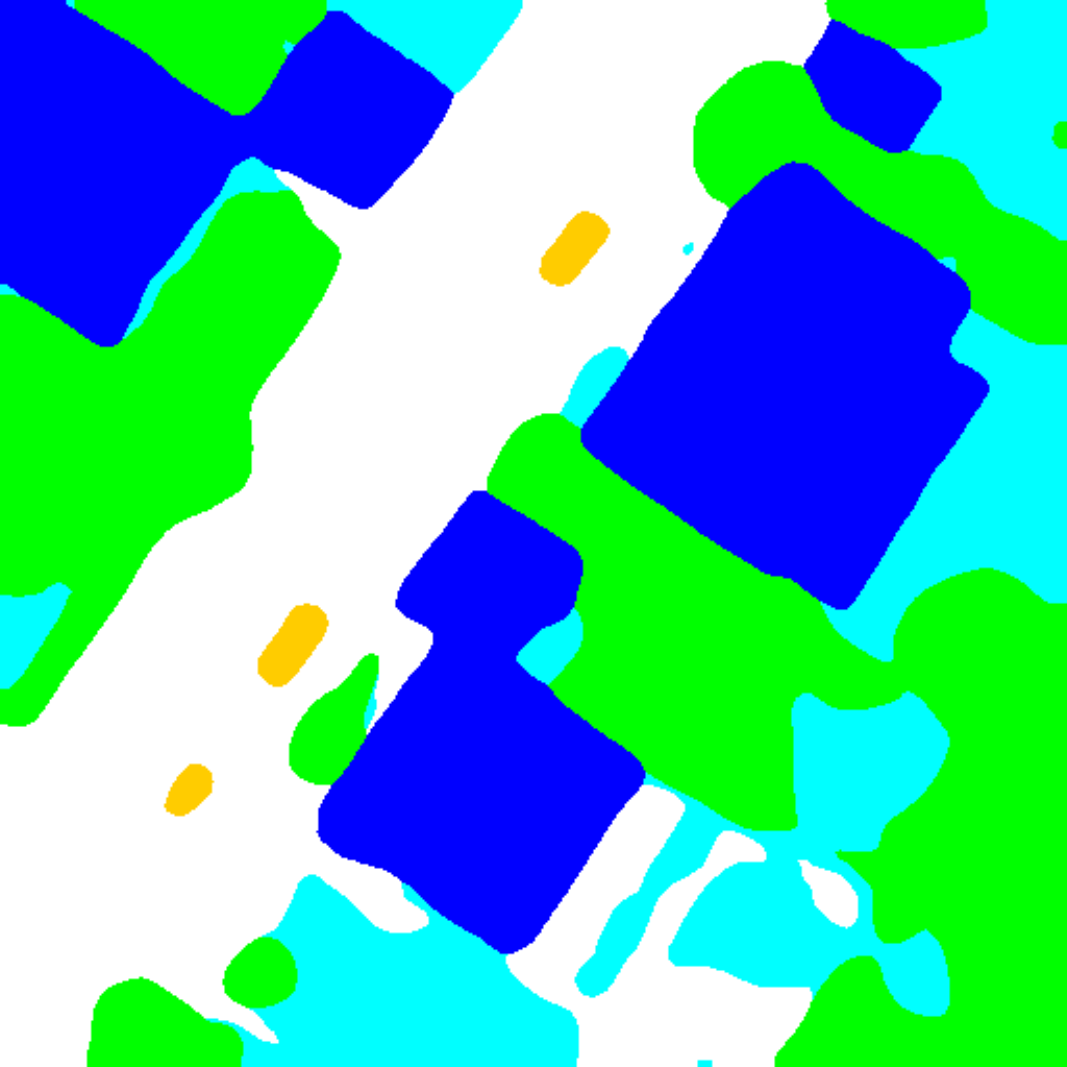}
		
		\vspace{1mm}
		
		\includegraphics[scale=0.094]{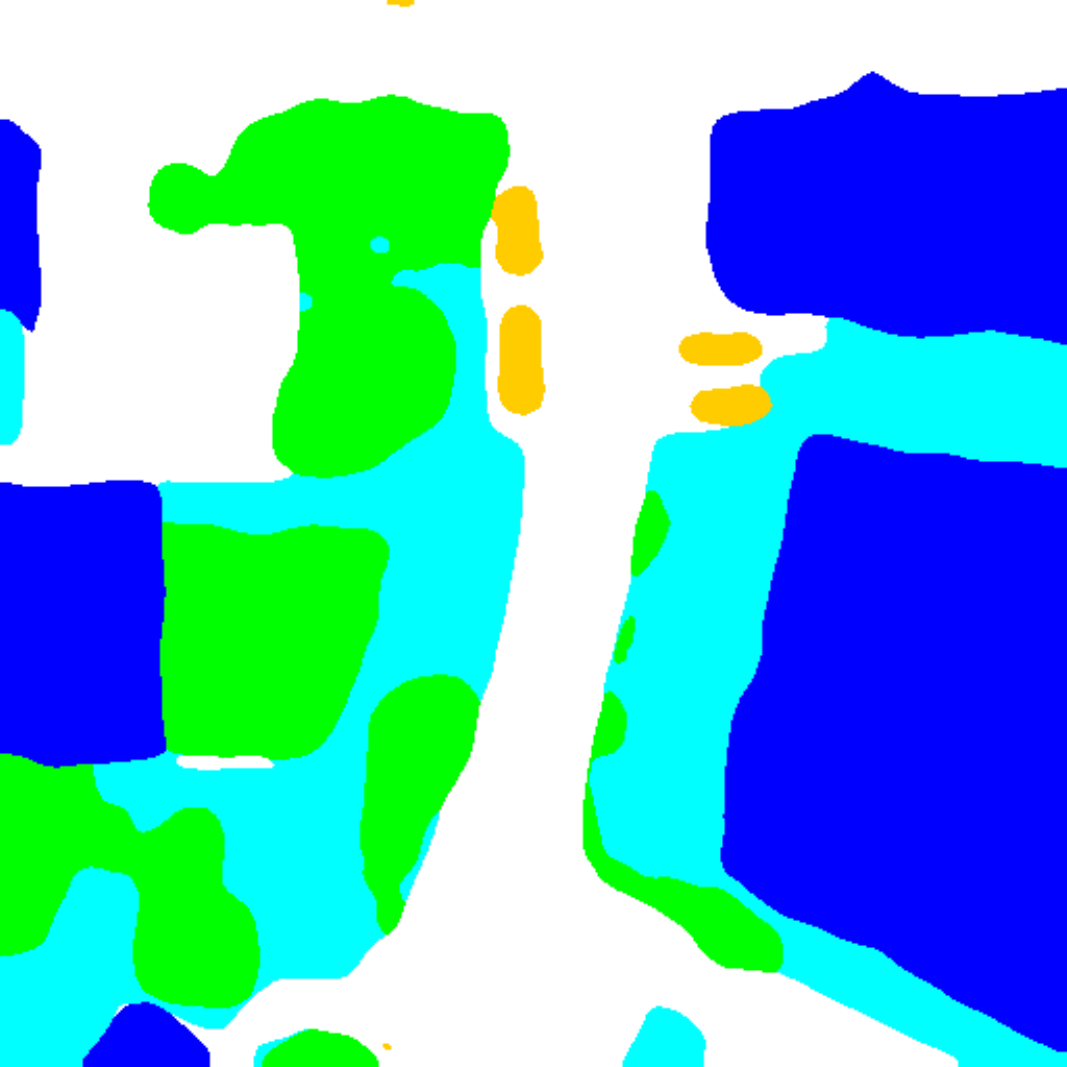}
		
		\vspace{1mm}
		
		\includegraphics[scale=0.094]{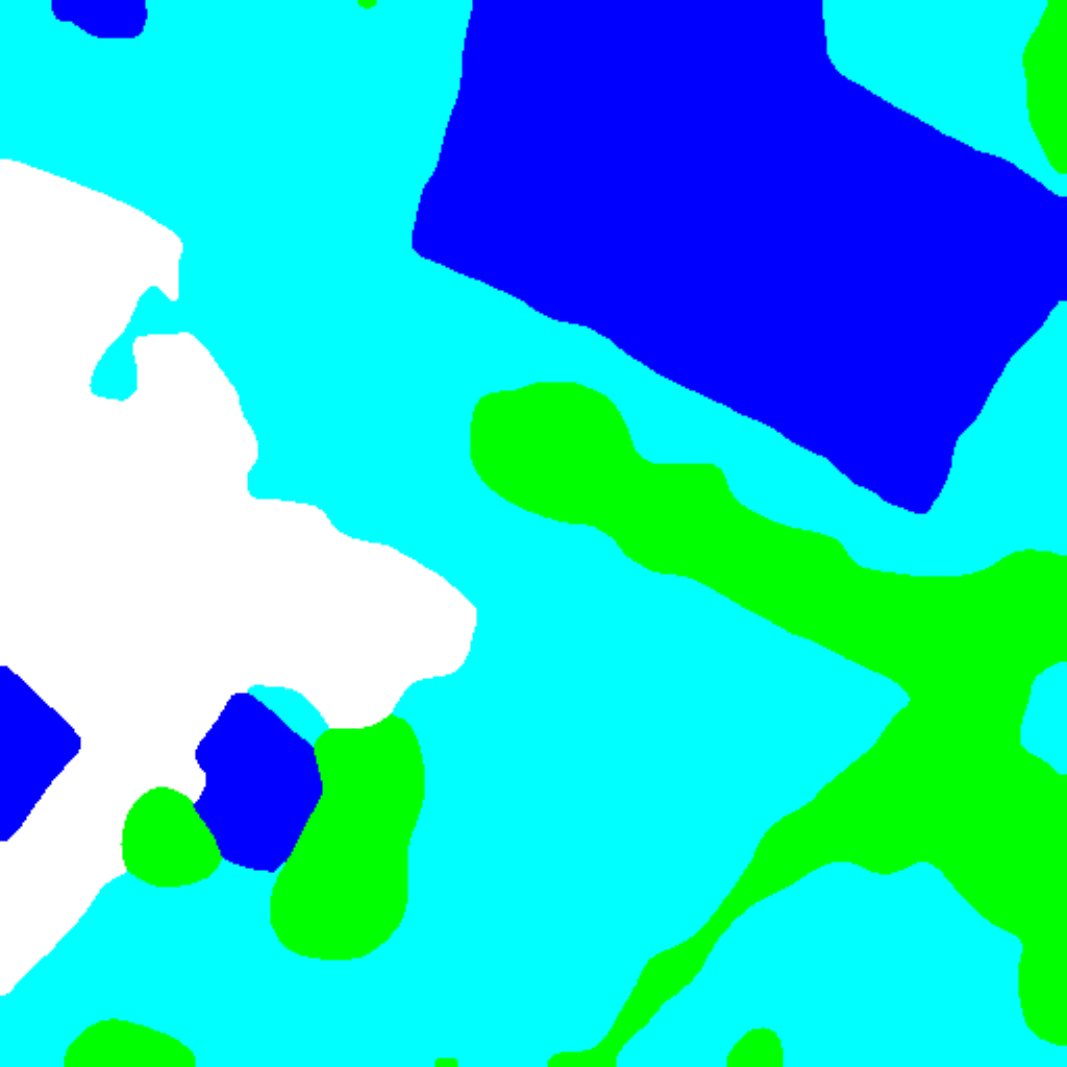}
		
		\vspace{1mm}
		
		\includegraphics[scale=0.094]{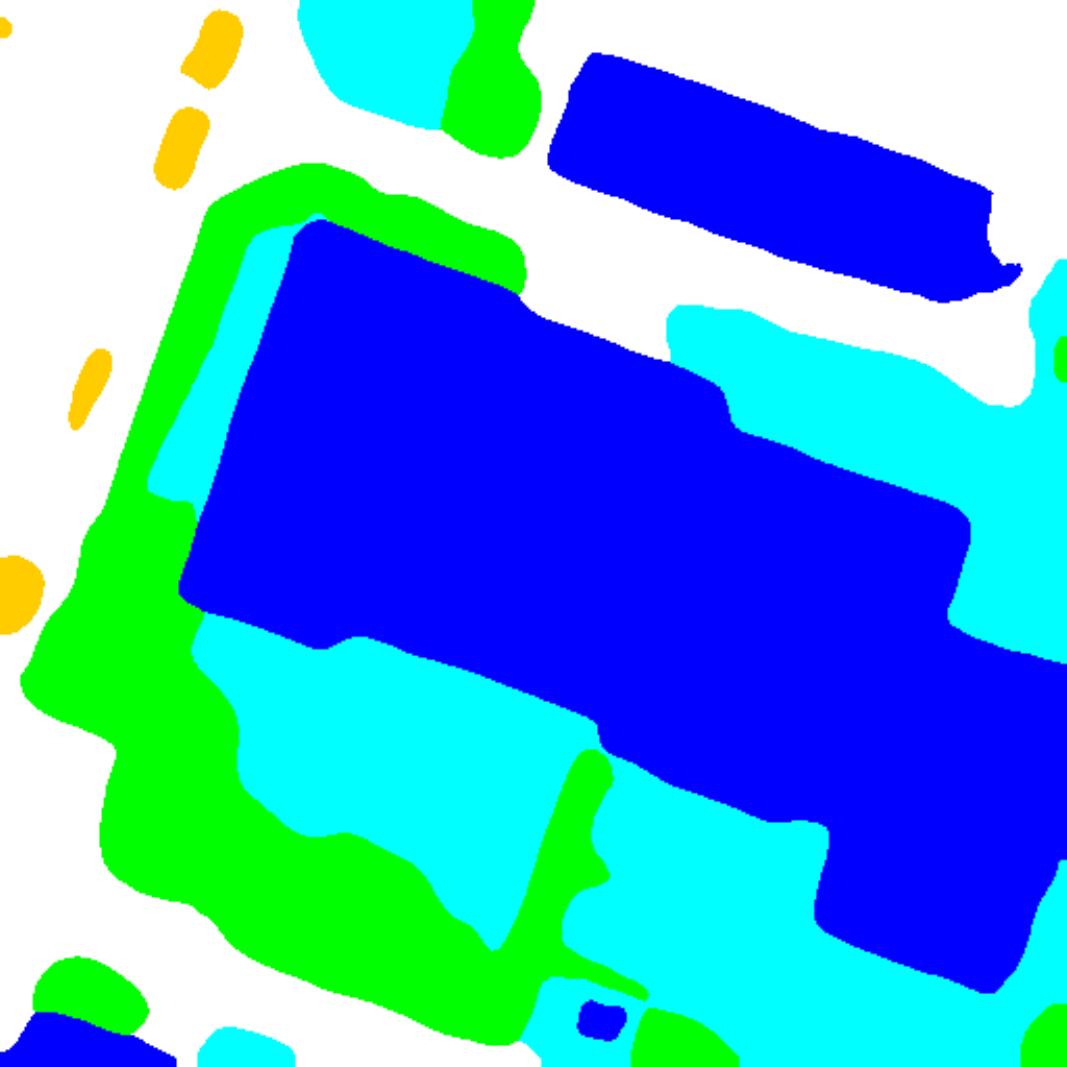}
		
		\vspace{1mm}
		
		\includegraphics[scale=0.094]{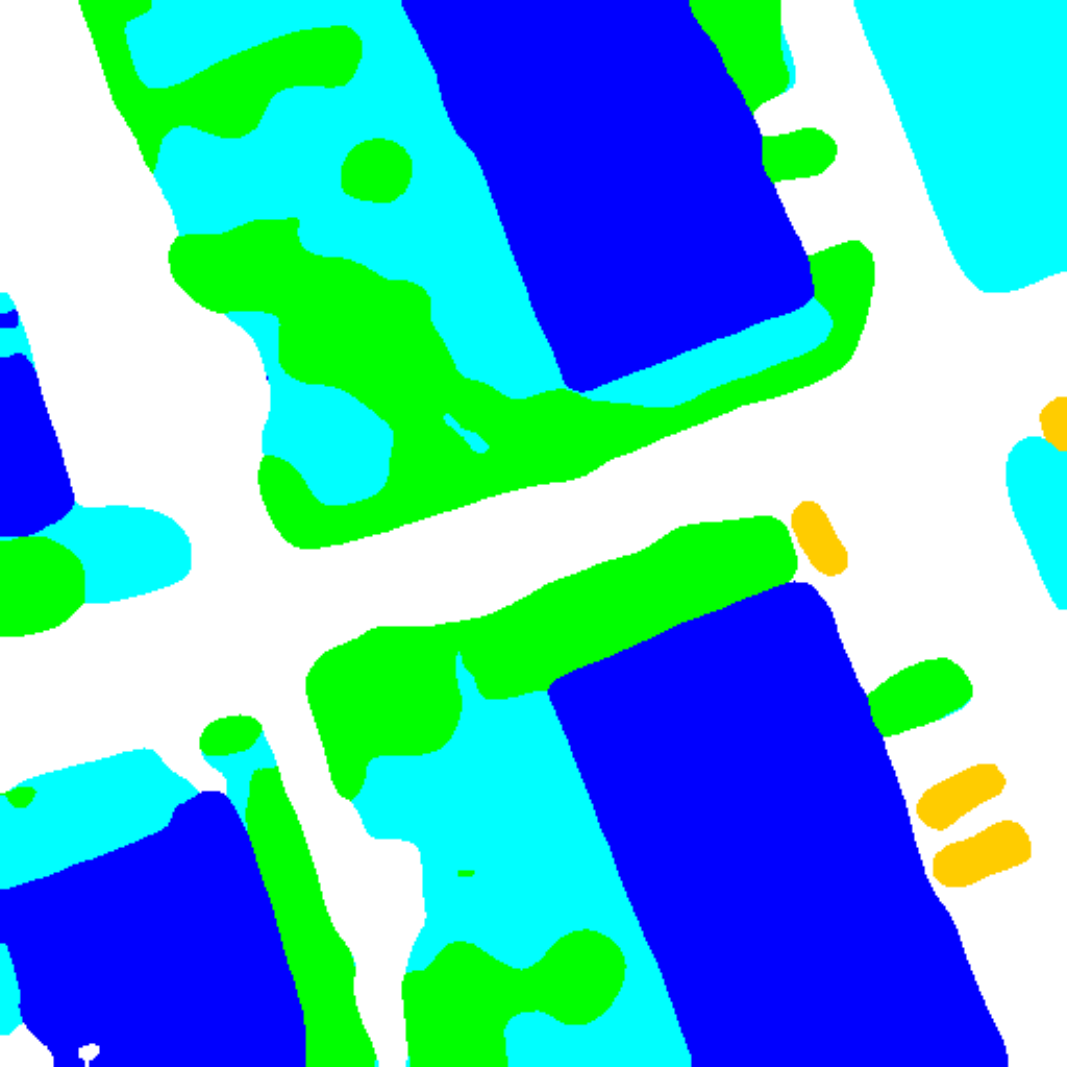}
		\centerline{(h)}
	\end{minipage}
	\begin{minipage}[t]{0.092\linewidth}
		\centering
		
		\includegraphics[scale=0.094]{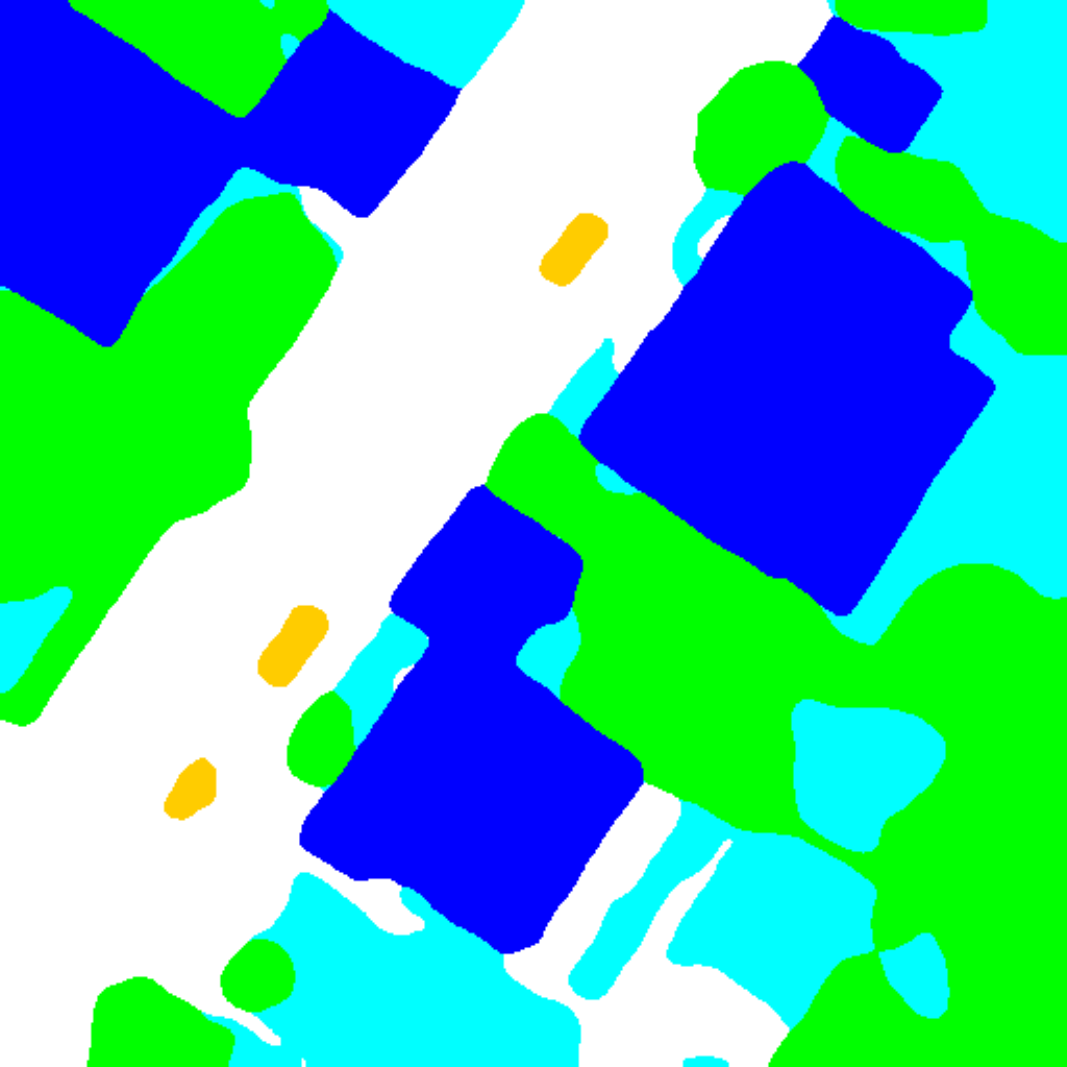}
		
		\vspace{1mm}
		
		\includegraphics[scale=0.094]{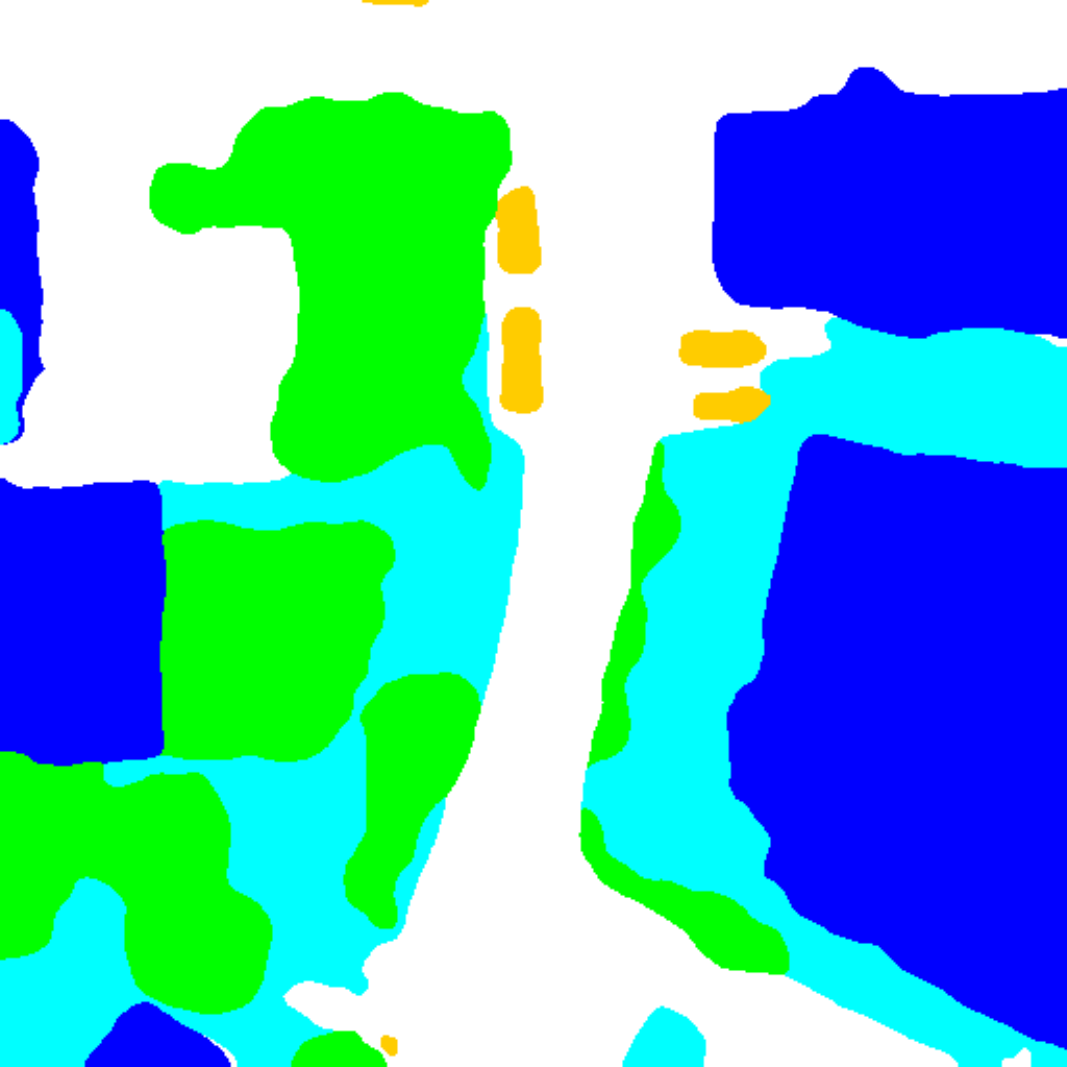}
		
		\vspace{1mm}
		
		\includegraphics[scale=0.094]{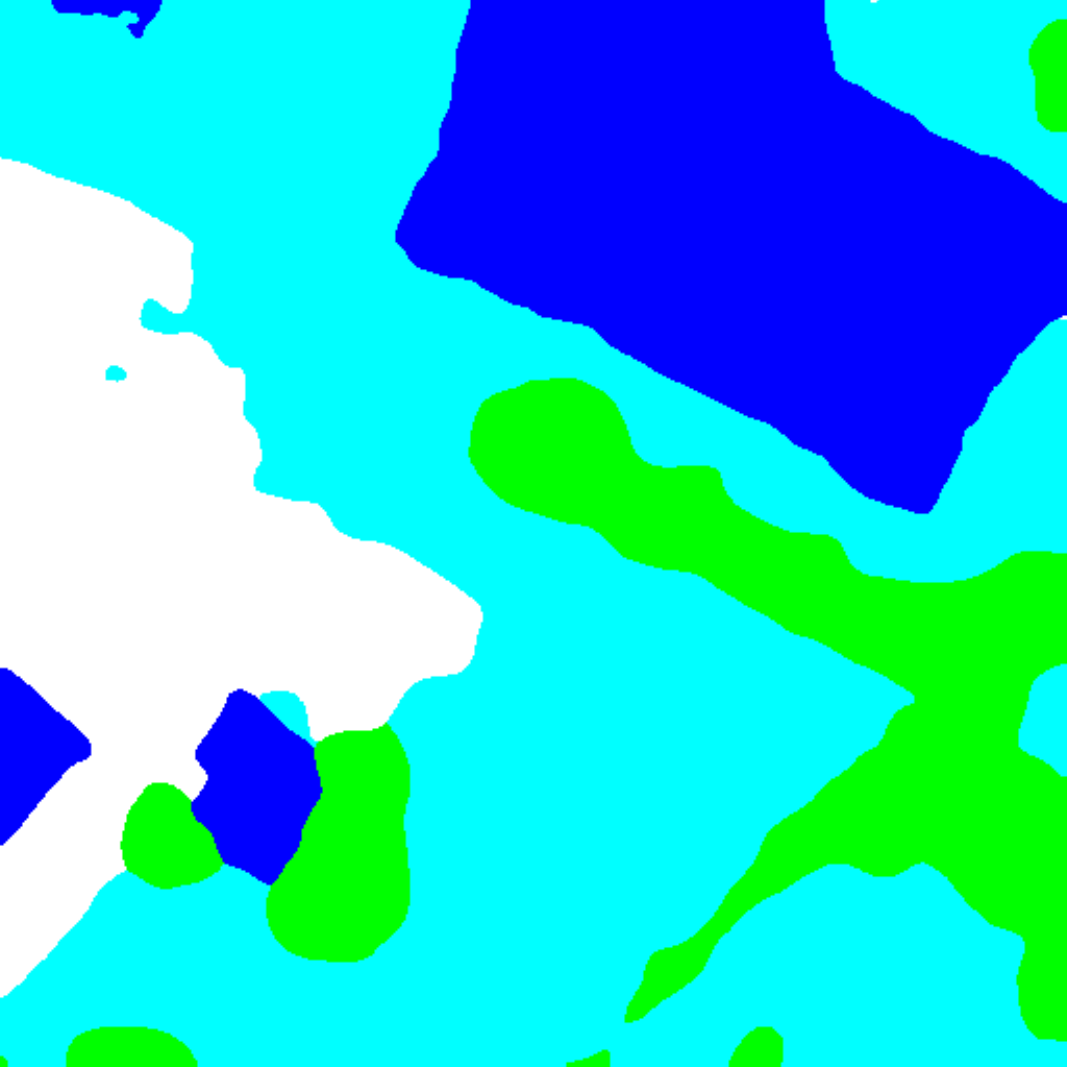}
		
		\vspace{1mm}
		
		\includegraphics[scale=0.094]{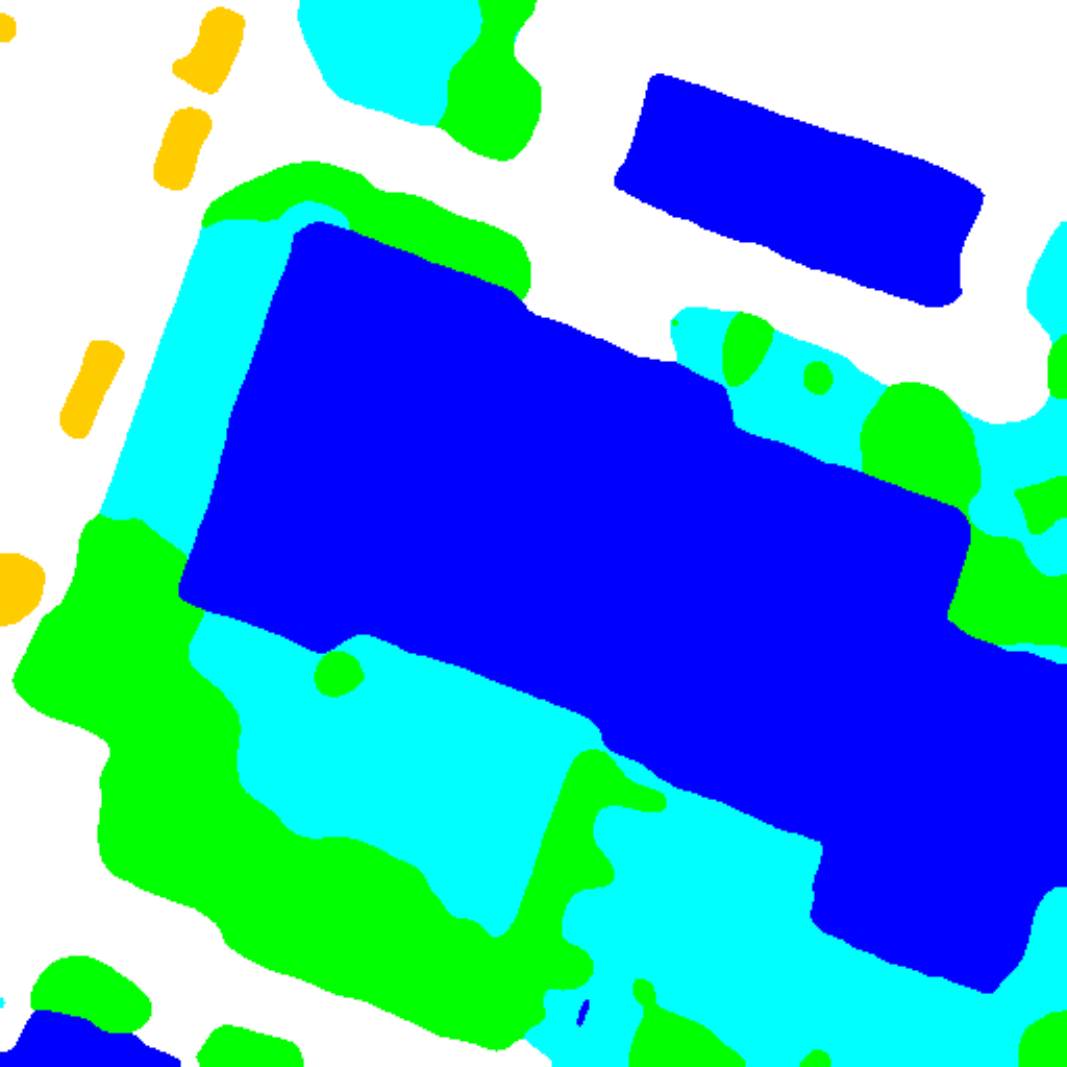}
		
		\vspace{1mm}
		
		\includegraphics[scale=0.094]{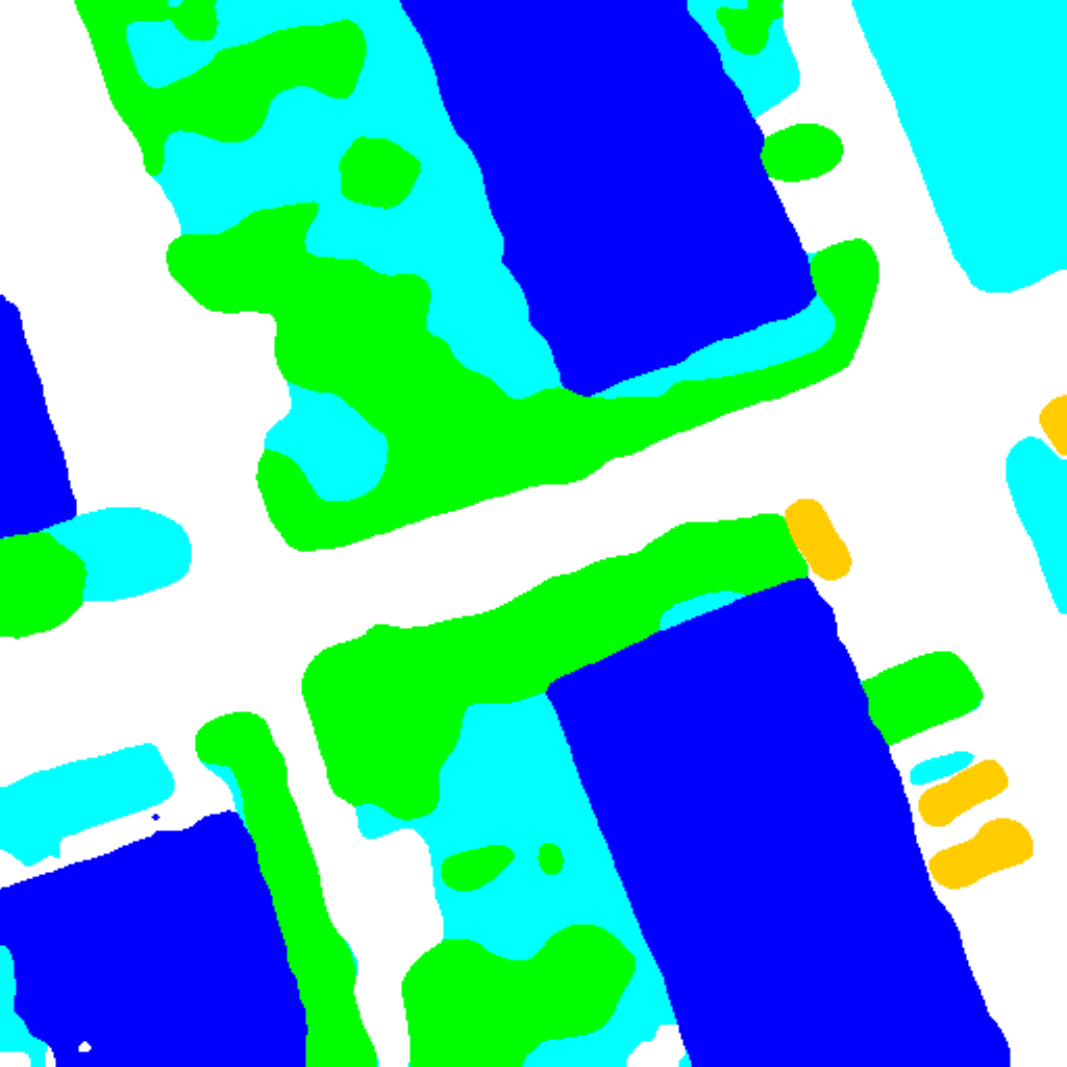}
		\centerline{(i)}
	\end{minipage}
	\begin{minipage}[t]{0.092\linewidth}
		\centering
		\includegraphics[scale=0.094]{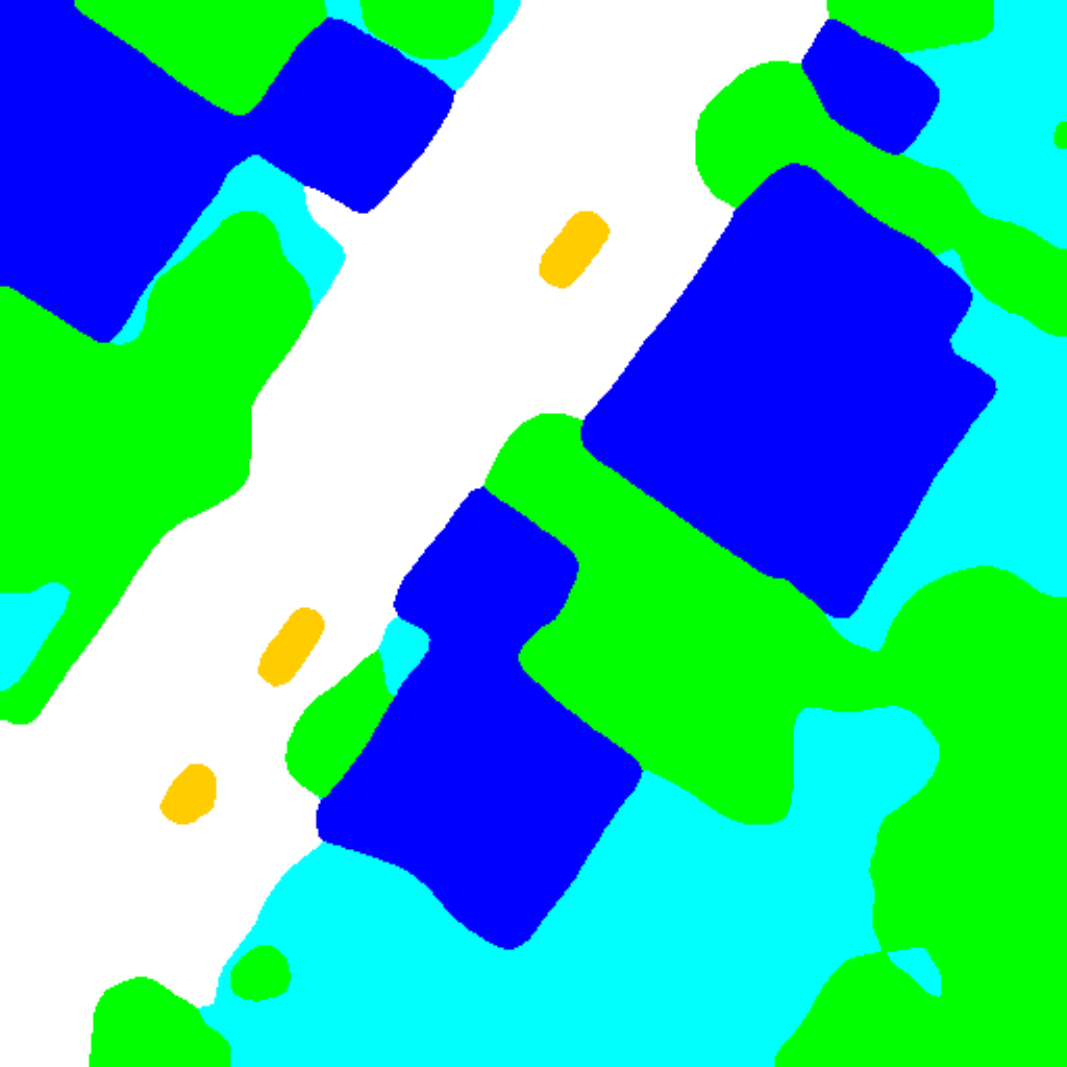}
		
		\vspace{1mm}
		
		\includegraphics[scale=0.094]{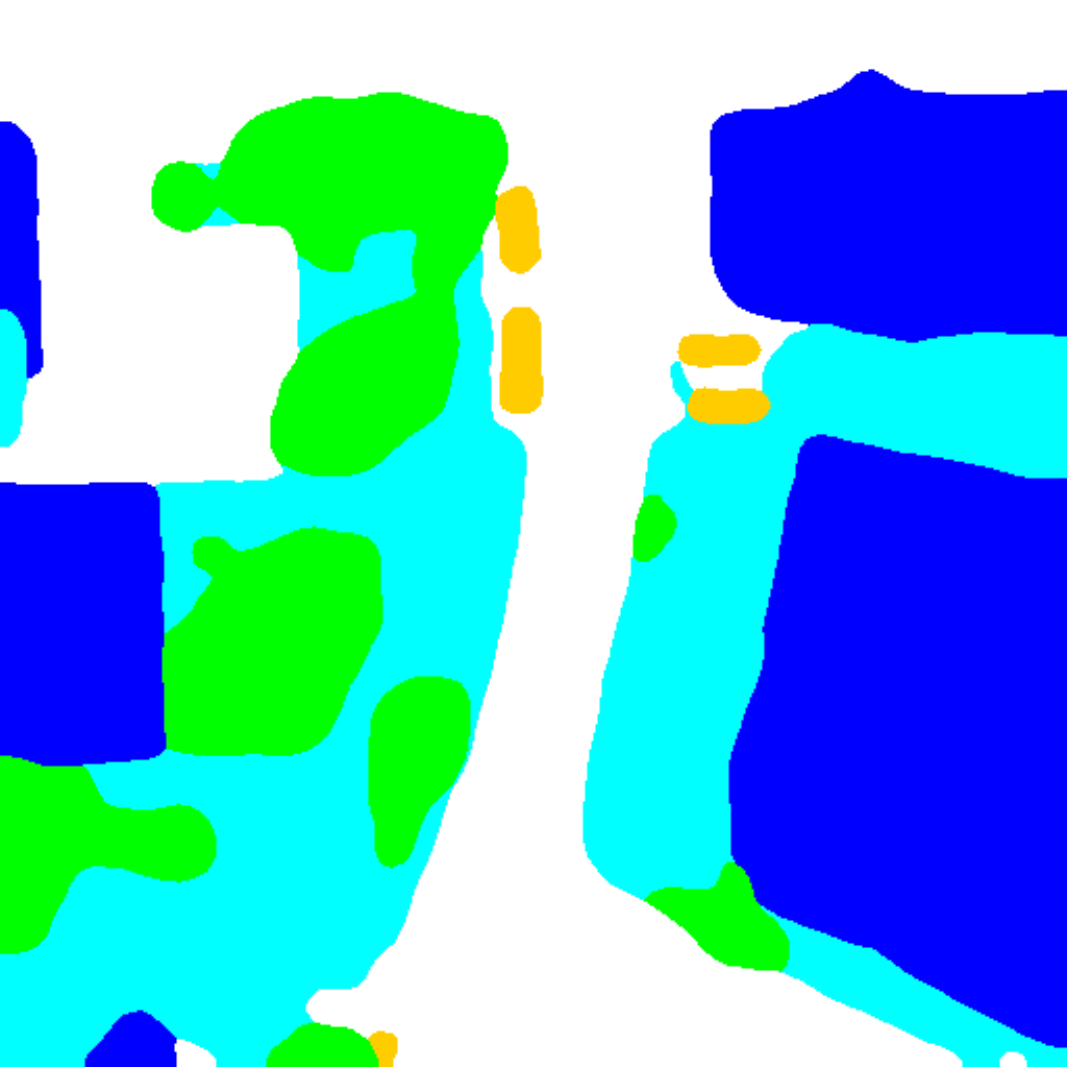}
		
		\vspace{1mm}
		
		\includegraphics[scale=0.094]{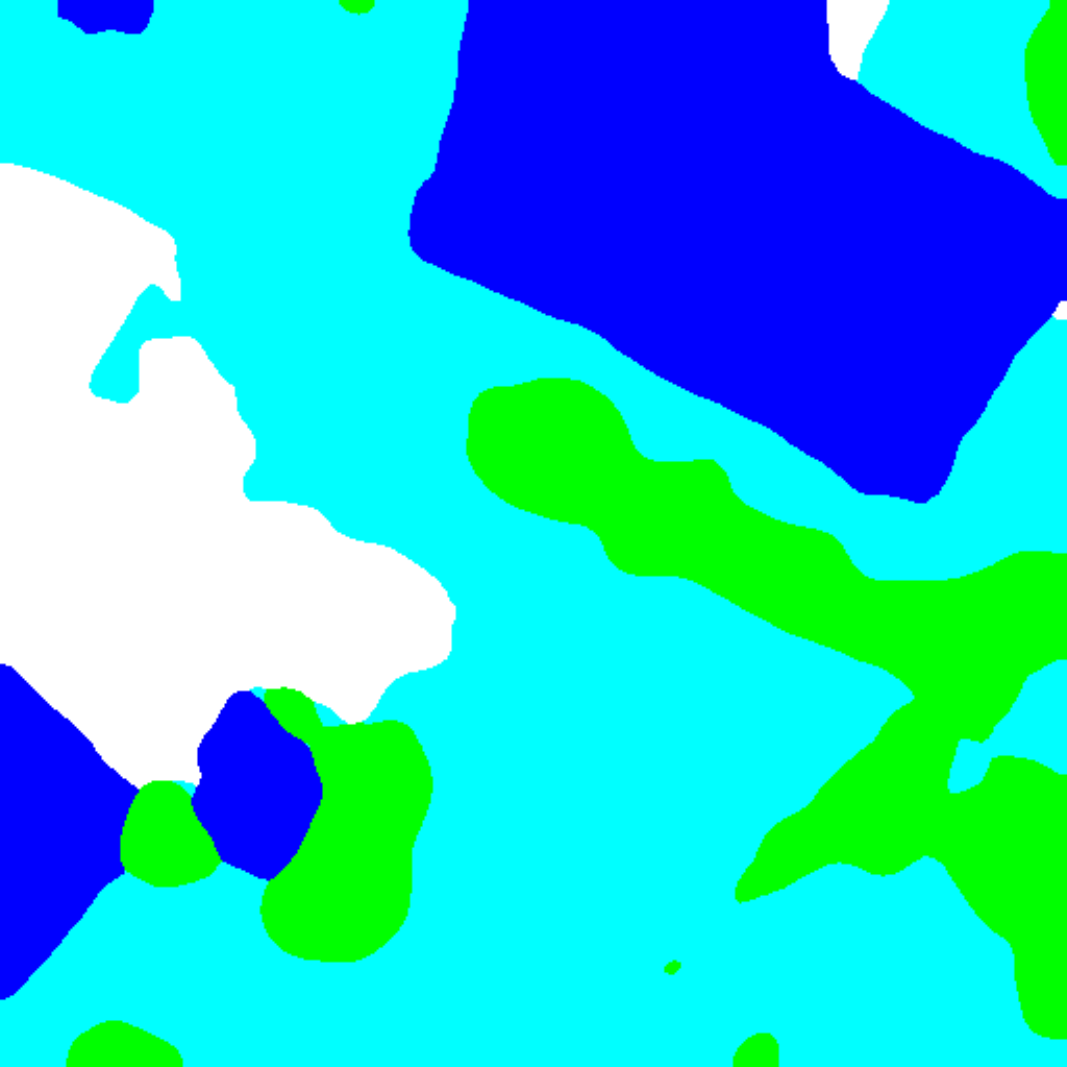}
		
		\vspace{1mm}
		
		\includegraphics[scale=0.094]{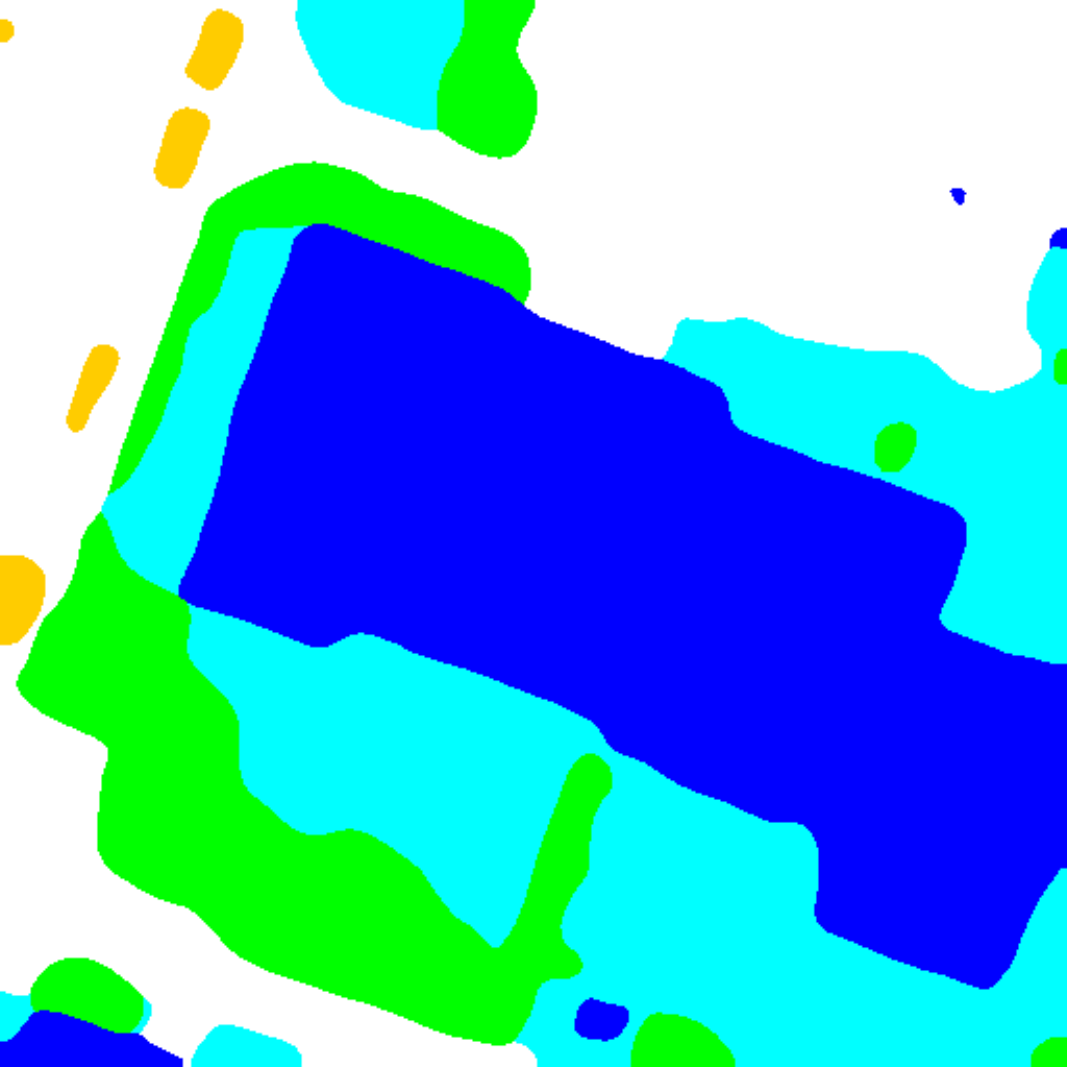}
		
		\vspace{1mm}
		
		\includegraphics[scale=0.094]{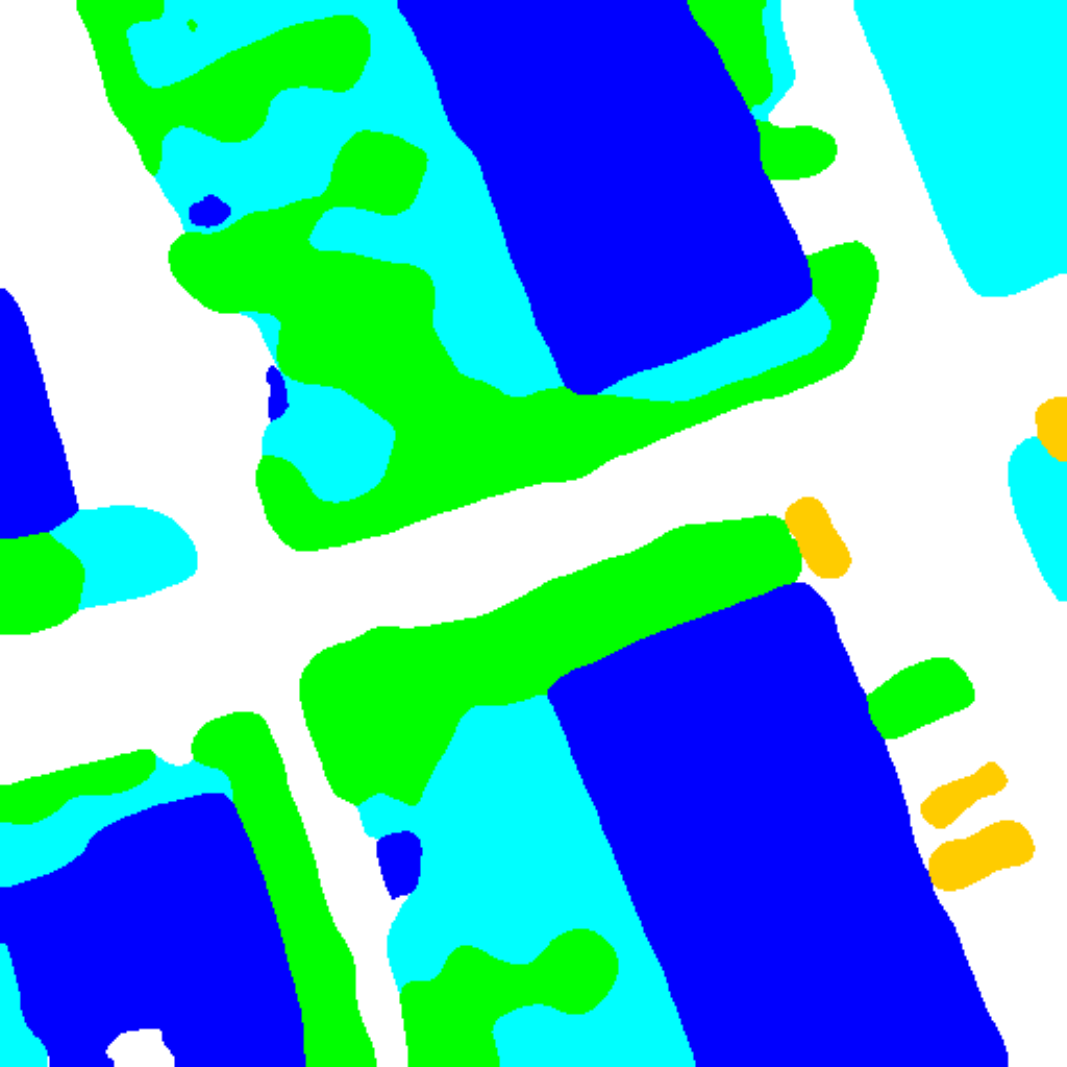}
		\centerline{(j)}
	\end{minipage}
	\begin{minipage}{\linewidth}
	\hfill
	\includegraphics[width=\textwidth]{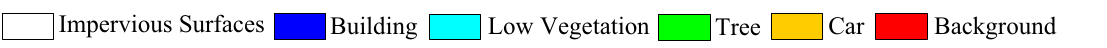}
	\end{minipage}
	\caption{Visualization segmentation results of different methods on the Vaihingen dataset ((a) Input Image, (b) Ground Truth, (c) BiseNet, (d) SwiftNet, (e) ShelfNet, (f) BANet, (g) ABCNet, (h) DDRNet, (i) UNetFormer, (j) BAFNet)} 
\label{fig7}
\end{figure*}

\begin{table*}
	\caption{Segmentation Results on the Potsdam Dataset\label{tab:table2}}
	\centering
	\begin{tabular}{|c||c||c||c||c||c||c||c||c||c||c|}
		\hline
		Method & Parameters & FLOPs & Imp. Surf & Building & Low Veg.& Tree & Car & mean F1 & OA($\%$) &mIoU($\%$) \\
		\hline
		BiseNet \cite{yu2018bisenet} &13.4M&14.8G&91.86&95.07&86.34&86.83&94.58&90.83&89.54&83.41\\
		\hline
		SwiftNet \cite{orsic2019defense}&11.8M&13.0G&91.84&95.43&86.55&87.83&94.89&	91.31&89.70&84.21\\
		\hline
		ShelfNet \cite{zhuang2019shelfnet}&14.6M&12.3G&92.59&95.97&87.41&88.57&95.24&	91.96&90.59&85.30\\
		\hline
		BANet \cite{wang2021transformer}&12.7M&15.2G&92.53&96.12&87.08&88.51&95.08&	91.86&90.48&85.15\\
		\hline
		ABCNet \cite{li2021abcnet} &14.0M&15.7G&92.89&96.35&86.80&87.66&95.62&	91.86&90.43&85.20\\	
		\hline
		DDRNet \cite{pan2022deep}&20.3M&17.9G&93.04&96.25&87.12&87.75&95.52&	91.94&90.53&85.31\\
		\hline
		UNetFormer \cite{wang2022unetformer} &11.7M&11.8G&92.73&96.20&87.36&88.82&	95.92&92.21&90.72&85.75\\
		\hline
		BAFNet (ours) &\pmb{6.4M}&12.3G&\pmb{93.12}&\pmb{96.73}&\pmb{87.98}&\pmb{89.35}&\pmb{96.14}&\pmb{92.66}&\pmb{91.28}&\pmb{86.53}\\
		\hline
		DC-Swin \cite{wang2022novel} &66.9M&72.2G&93.44&97.06&88.10&89.68&96.05&	92.87&91.48&86.88\\
		\hline
		FT-UNetFormer \cite{wang2022unetformer} &96M&128.4G&93.45&96.86&88.12&89.66&96.35&	92.89&91.39&86.92\\
		\hline
	\end{tabular}
\end{table*}

Table 2 shows the segmentation results of all comparison methods on the Potsdam dataset, highlighting the optimal value for each item achieved by the lightweight network in bold. Compared to all lightweight segmentation networks, our proposed BAFNet achieved the optimal F1 score for each category. The overall segmentation result of mean F1 score of 92.66, OA of 91.28$\%$, and mIoU of 86.53$\%$. It outperforms the best lightweight model on the Potsdam dataset, UNetFormer, by 0.8$\%$ in terms of mIoU.
When compared to the two non-lightweight state-of-the-art models based on Transformers, BAFNet's mean F1 score, OA, and mIoU are already very close. This fully demonstrates that our proposed BAFNet, despite employing a lightweight backbone, improves the network's capacity to capture long-range dependencies and detailed information by designing a novel bilateral network architecture. As a result, it achieves comparable performance to segmentation networks that utilize Swin-Base as the backbone.
Some visual segmentation results obtained from different comparison methods on the Potsdam test set are shown in Fig. 8, 
which can show the  accuracy of our network.

\begin{figure*}
	\centering{}
	\begin{minipage}[t]{0.092\linewidth}
		\centering
		\includegraphics[scale=0.094]{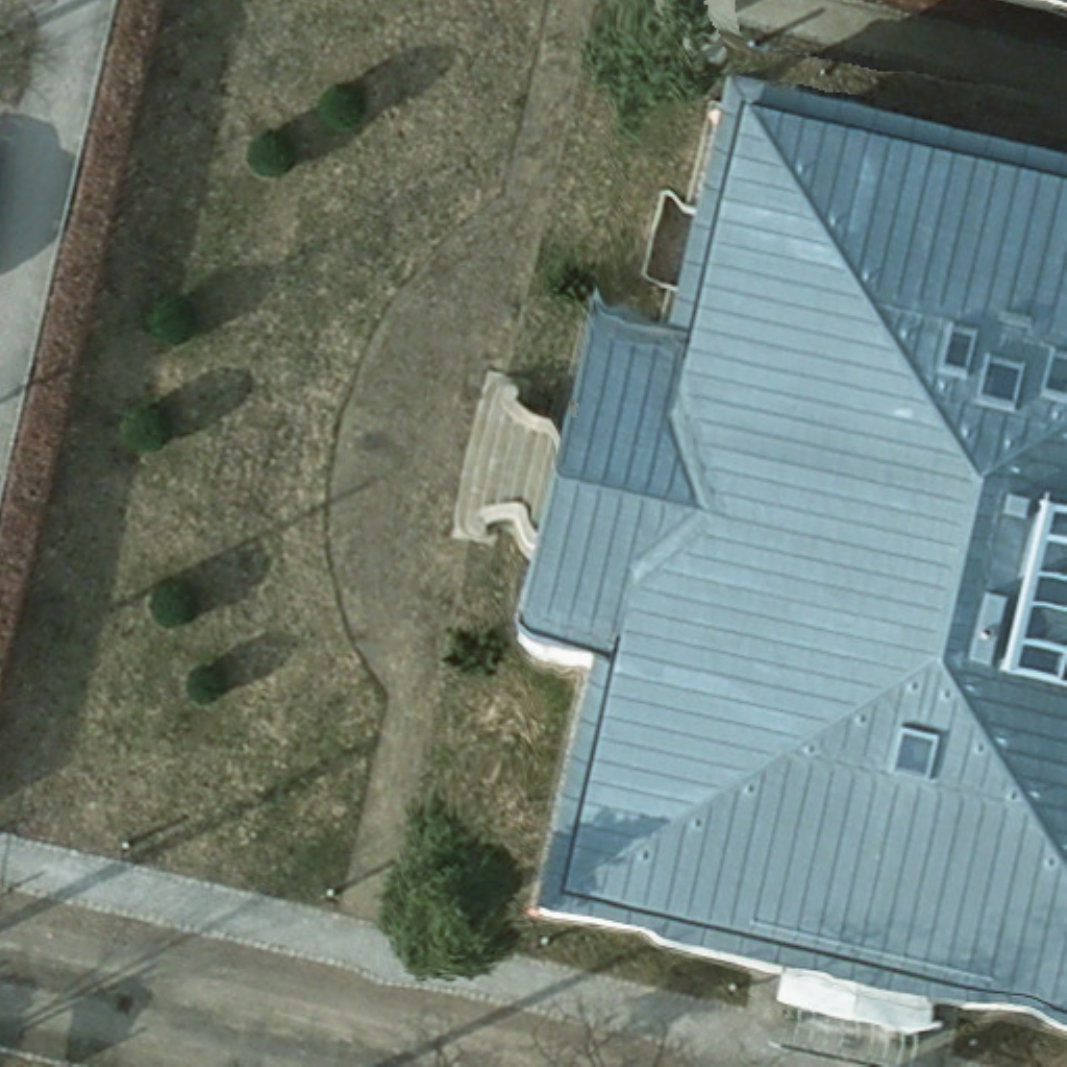}
		
		\vspace{1mm}
		
		\includegraphics[scale=0.094]{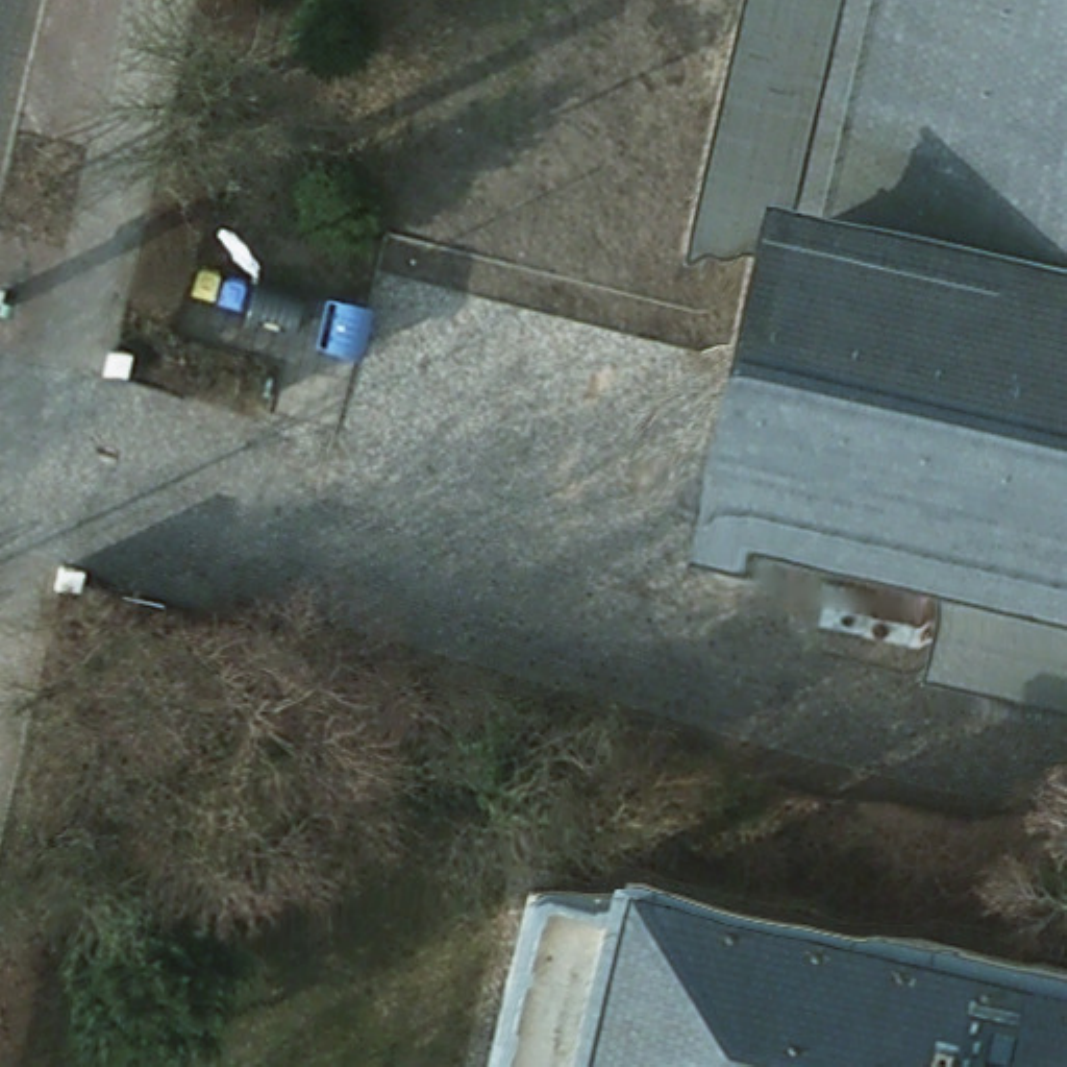}
		
		\vspace{1mm}
		
		\includegraphics[scale=0.094]{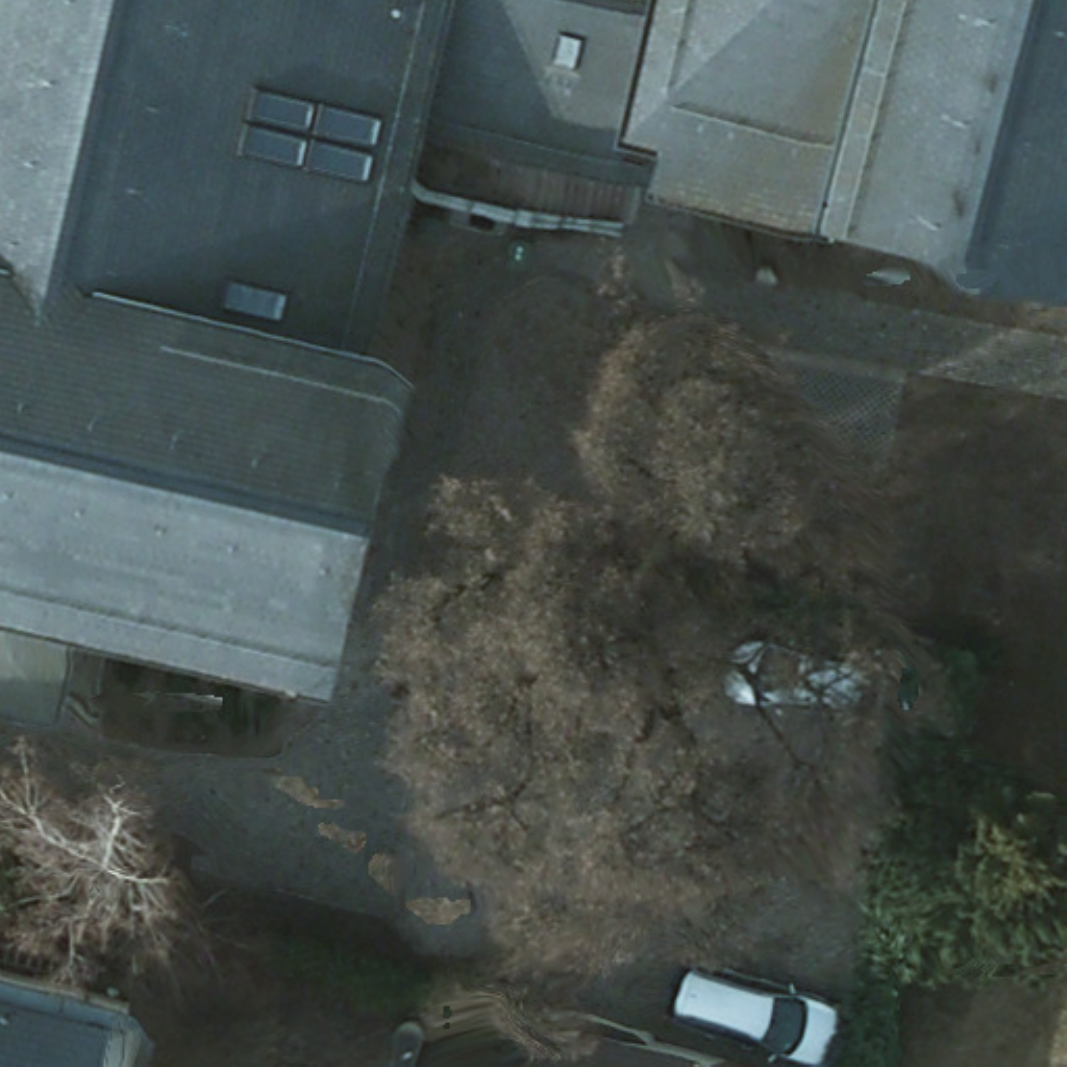}
		
		\vspace{1mm}
		
		\includegraphics[scale=0.094]{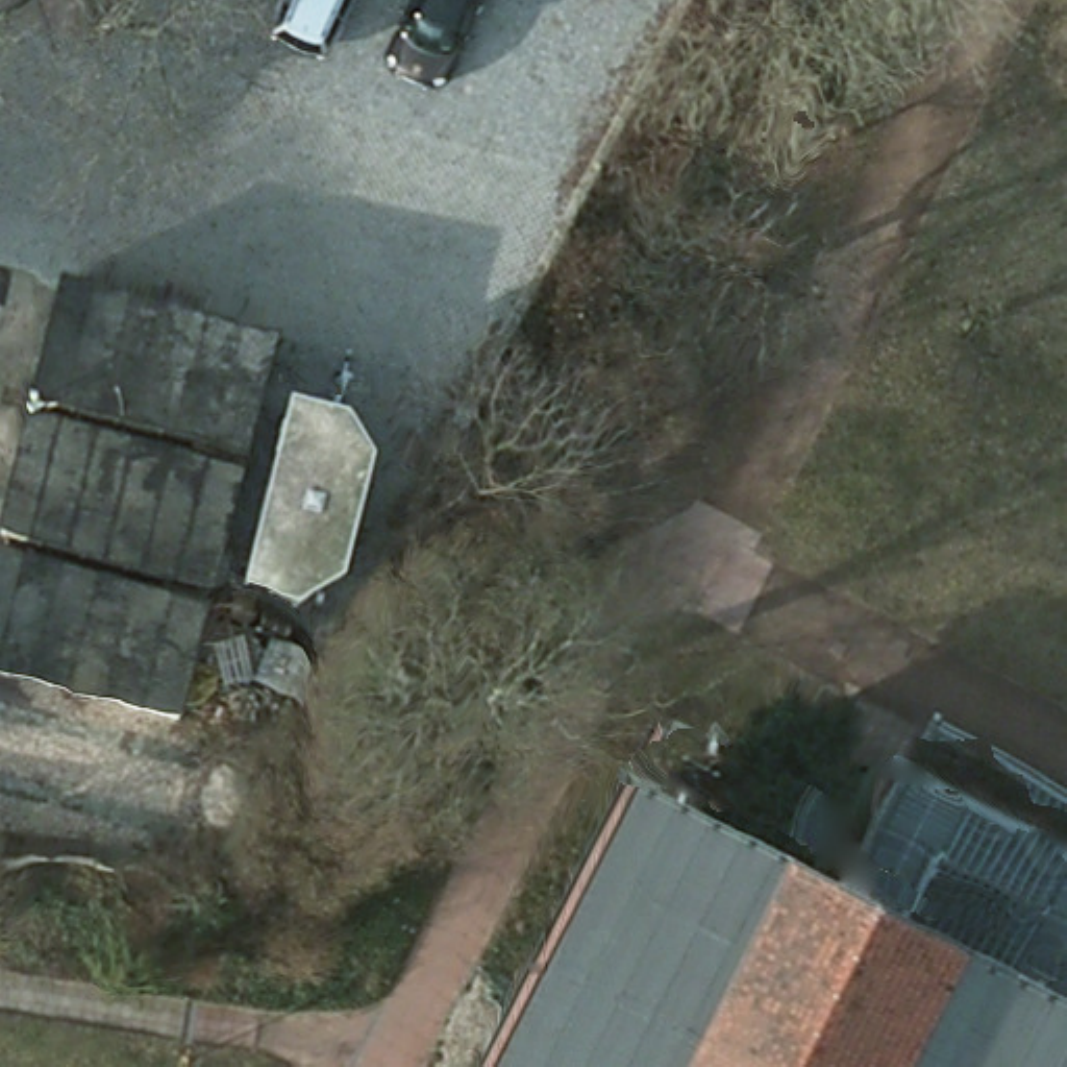}
		
		\vspace{1mm}
		
		\includegraphics[scale=0.094]{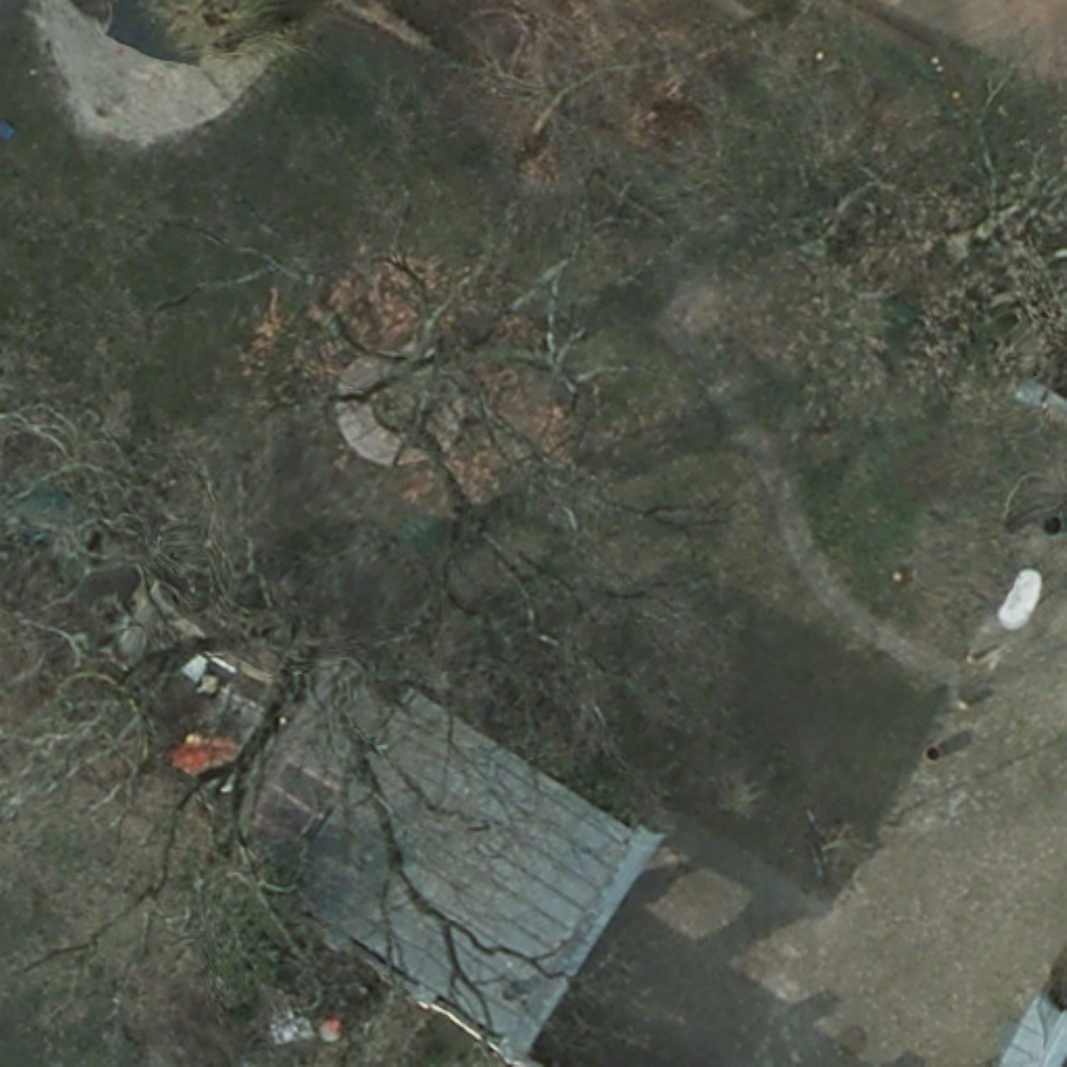}
		\centerline{(a)}
	\end{minipage}
	\begin{minipage}[t]{0.092\linewidth}
		\centering
		\includegraphics[scale=0.094]{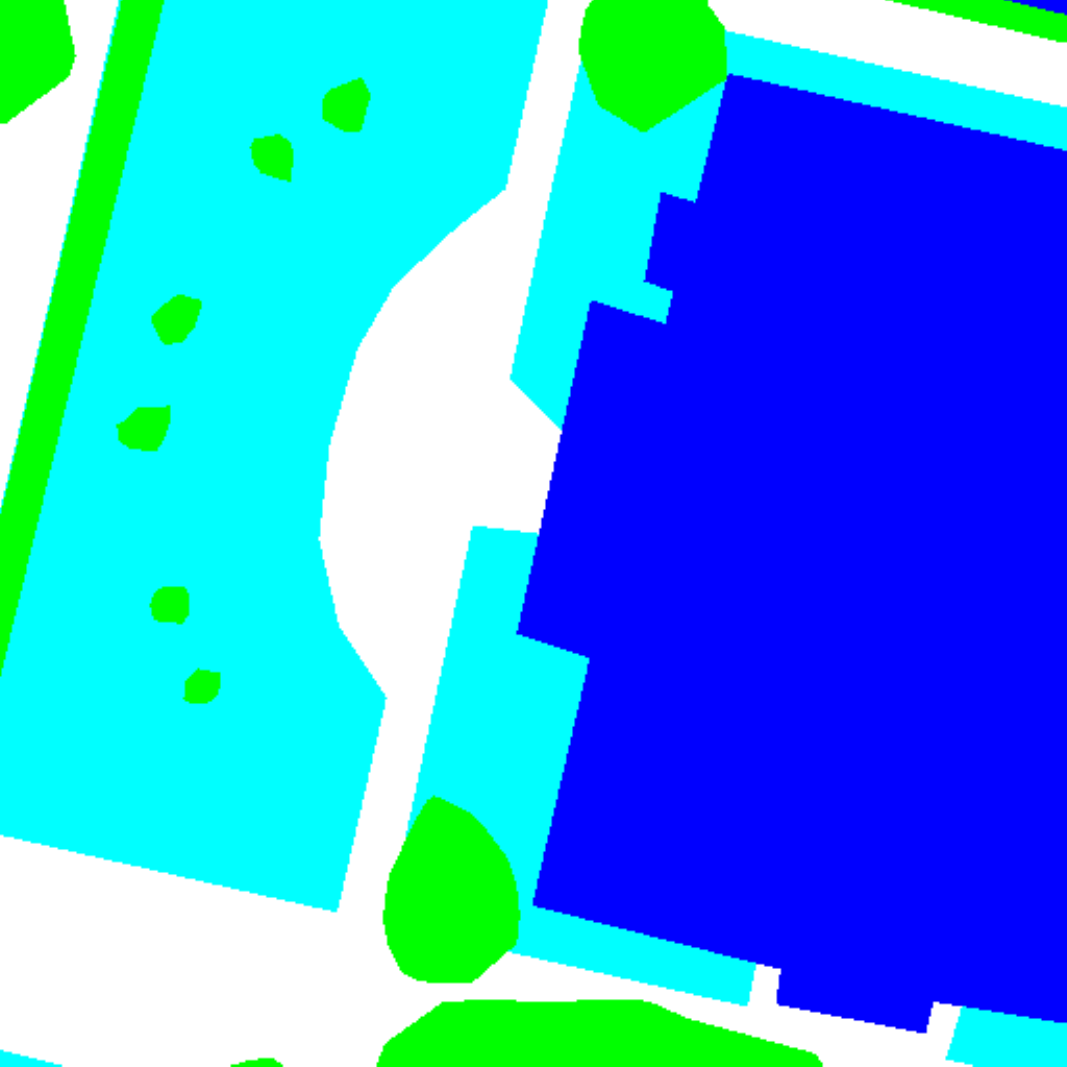}
		
		\vspace{1mm}
		
		\includegraphics[scale=0.094]{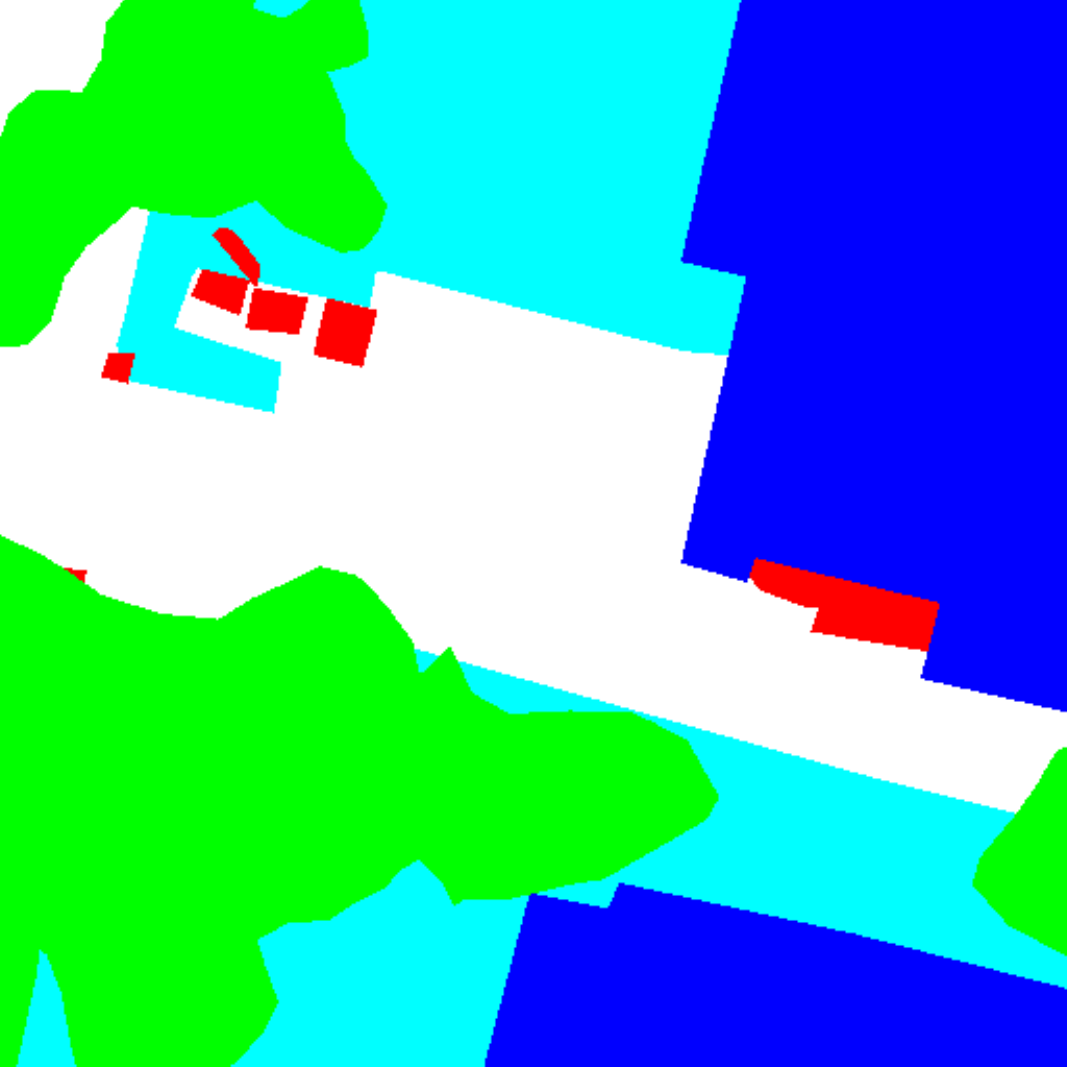}
		
		\vspace{1mm}
		
		\includegraphics[scale=0.094]{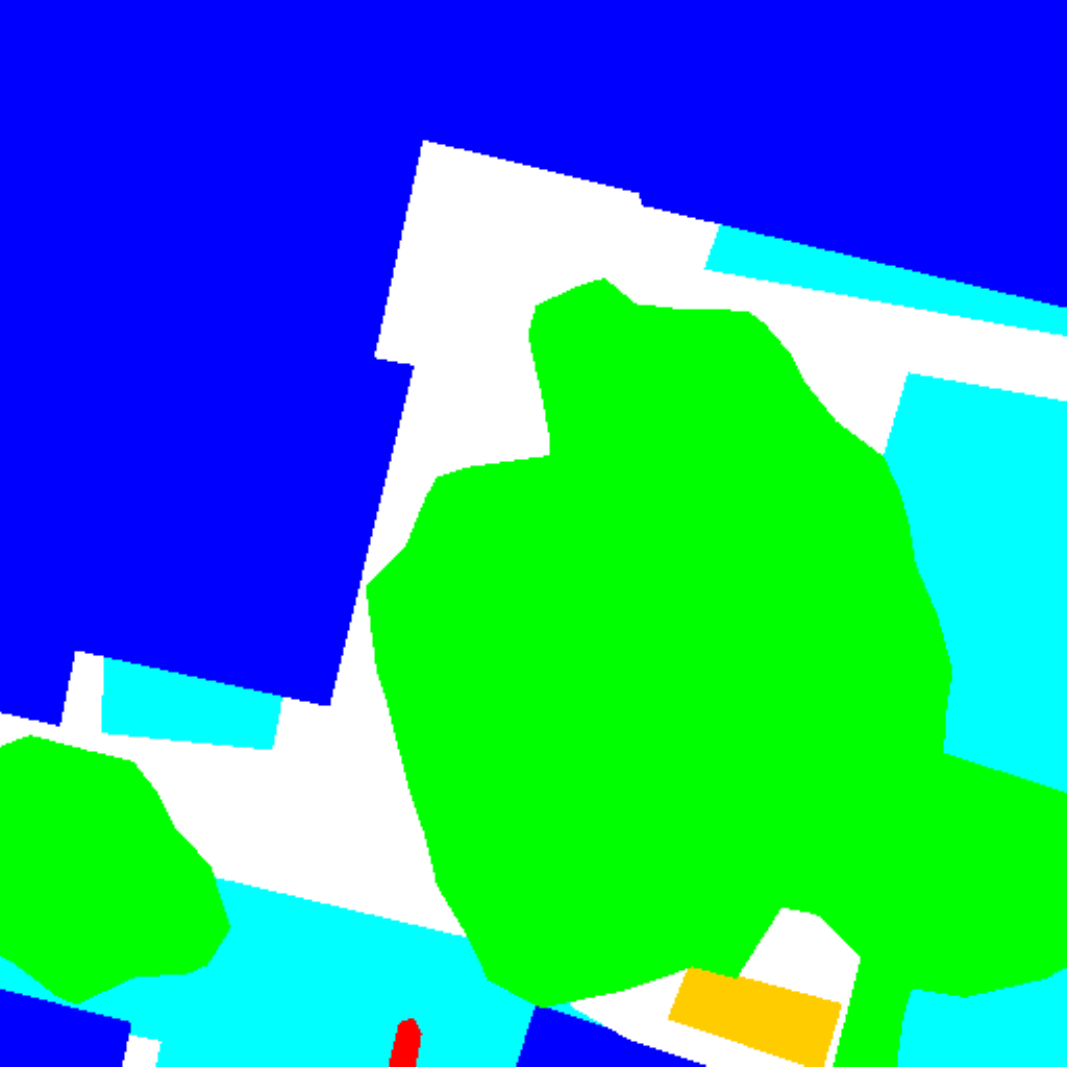}
		
		\vspace{1mm}
		
		\includegraphics[scale=0.094]{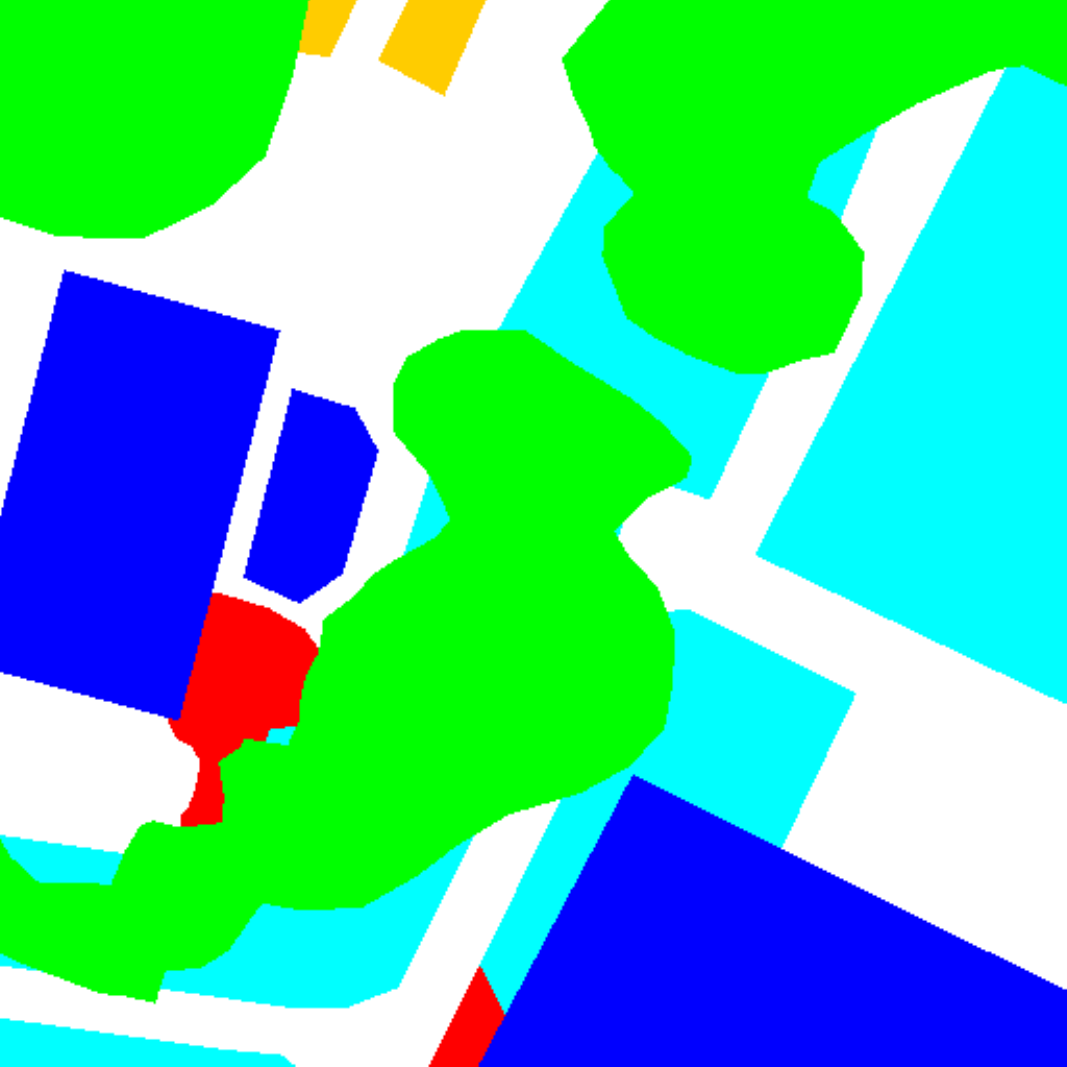}
		
		\vspace{1mm}
		
		\includegraphics[scale=0.094]{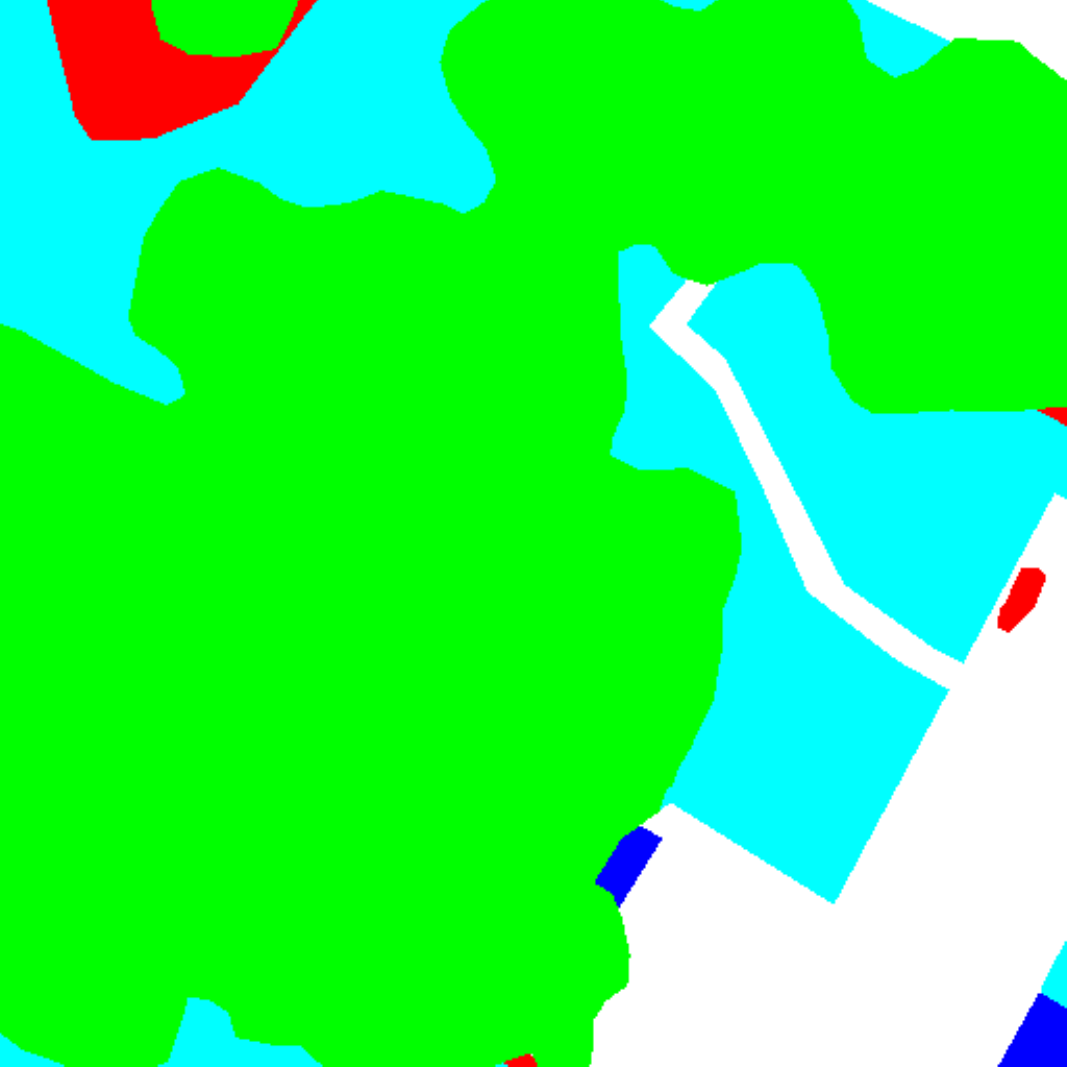}
		\centerline{(b)}
	\end{minipage}
	\begin{minipage}[t]{0.092\linewidth}
		\centering
		\includegraphics[scale=0.094]{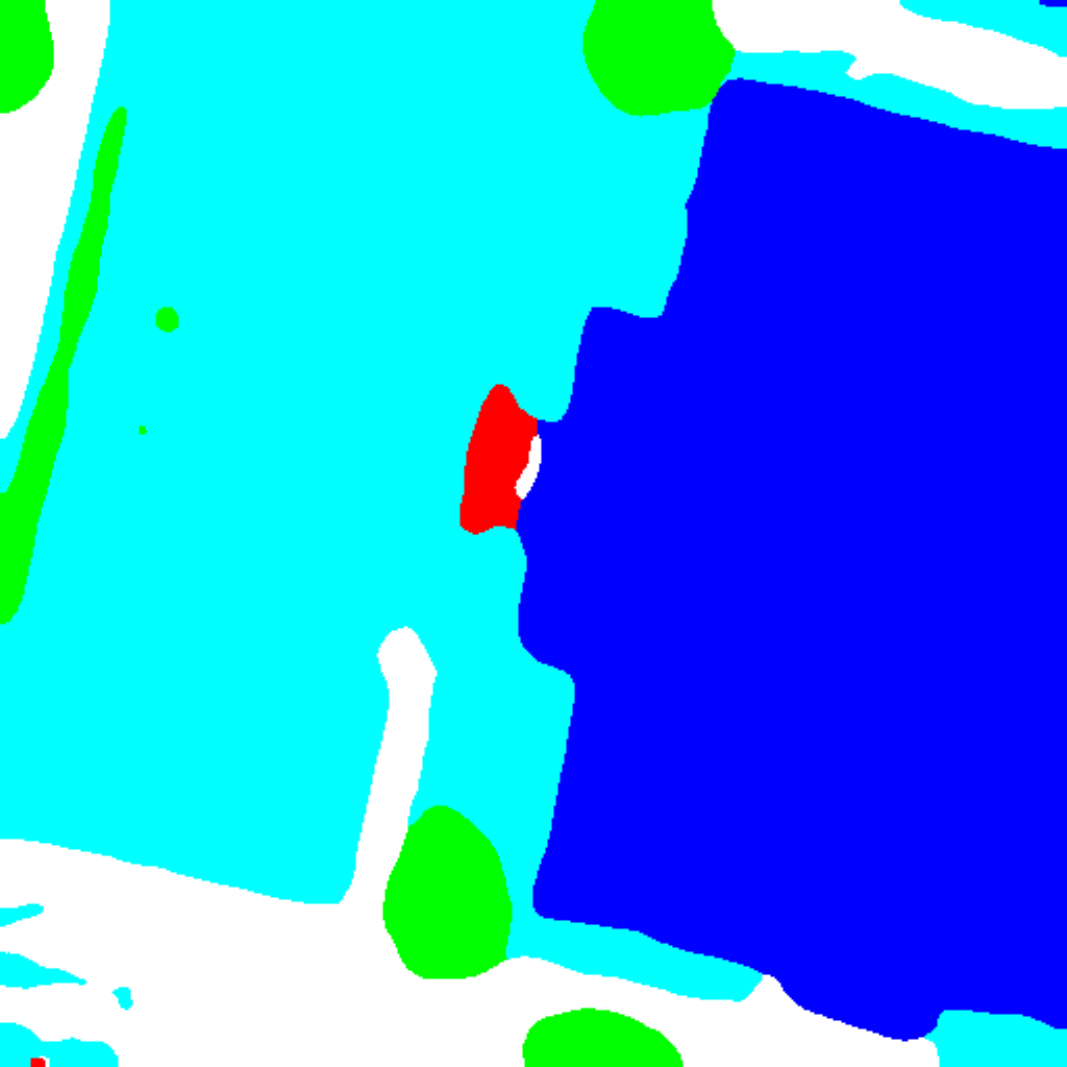}
		
		\vspace{1mm}
		
		\includegraphics[scale=0.094]{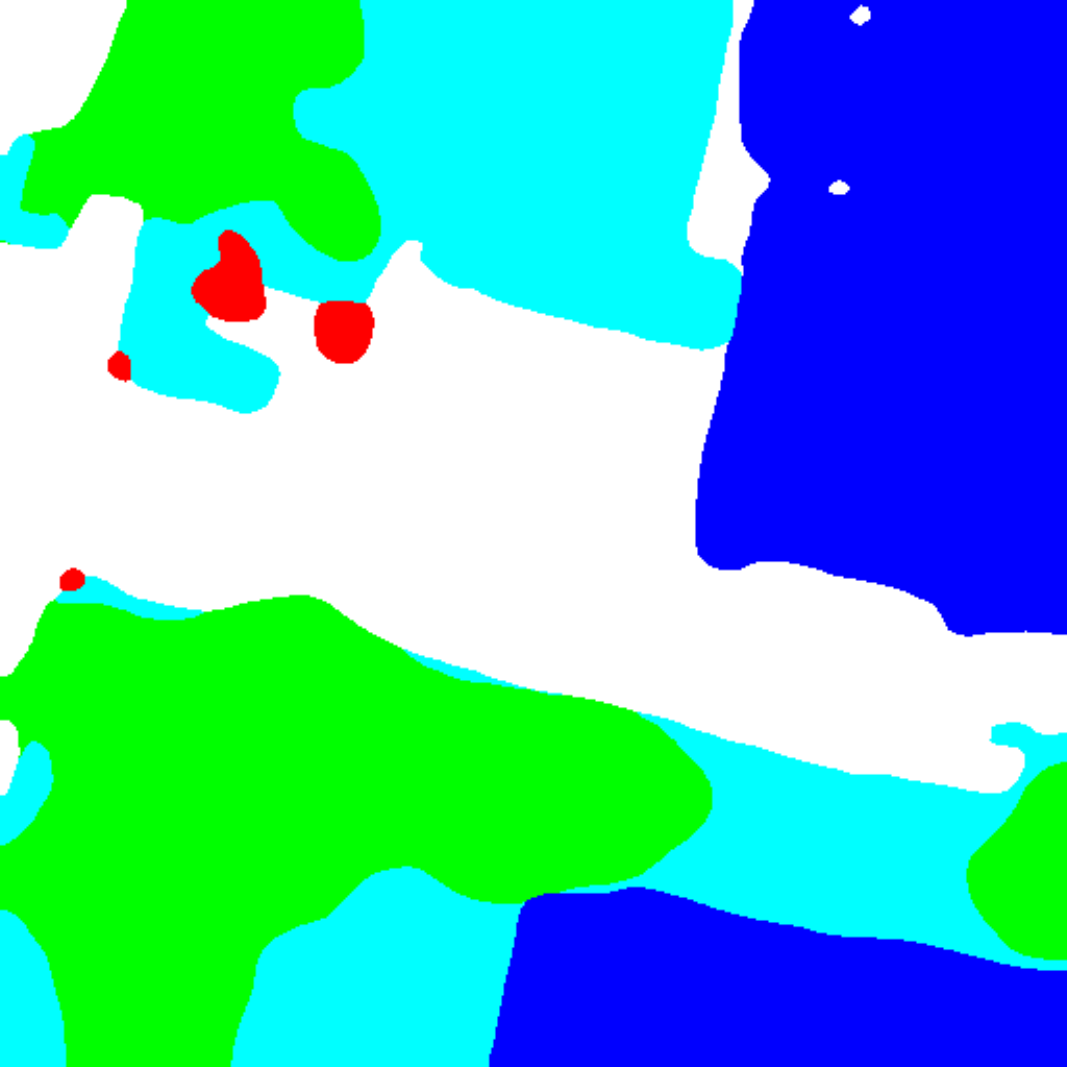}
		
		\vspace{1mm}
		
		\includegraphics[scale=0.094]{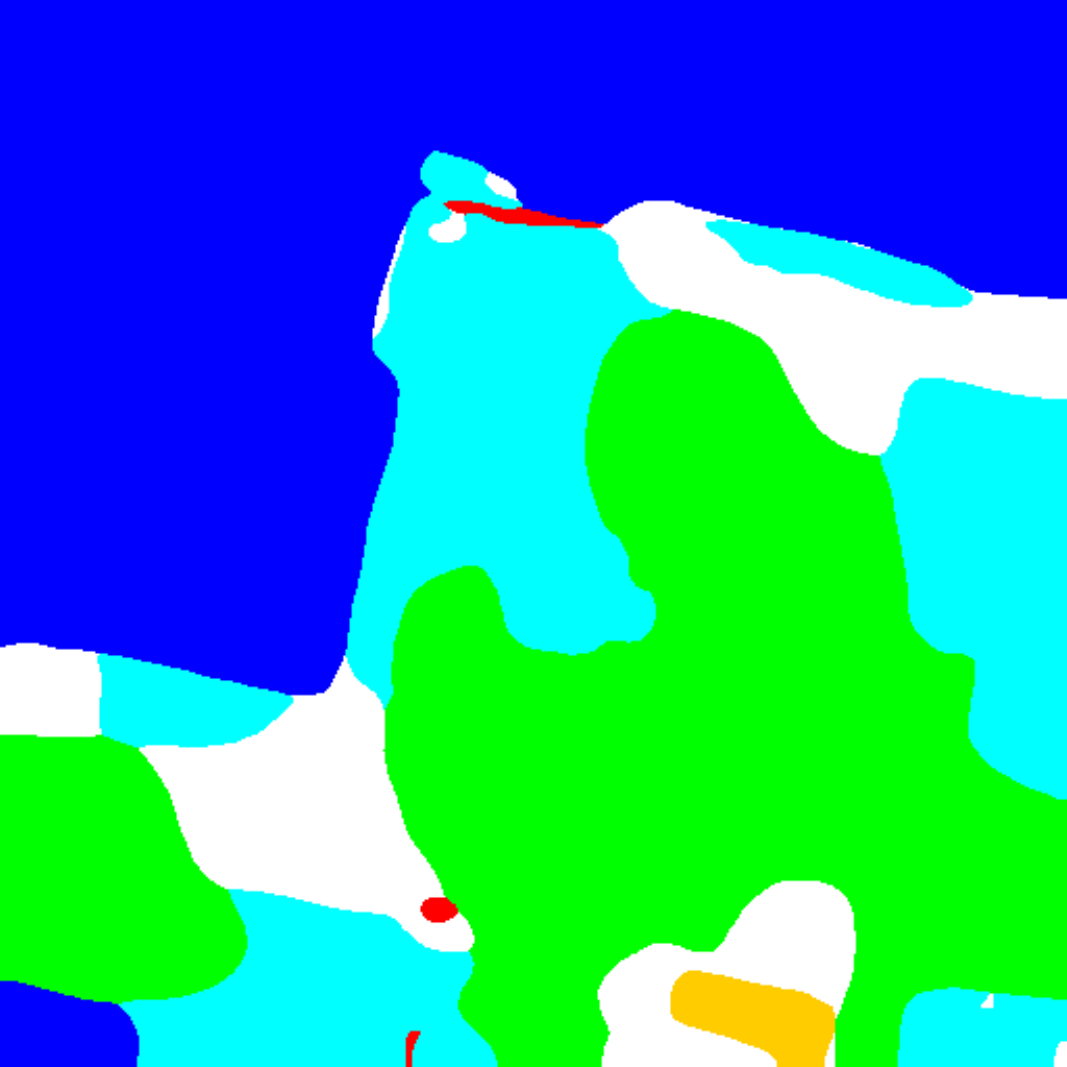}
		
		\vspace{1mm}
		
		\includegraphics[scale=0.094]{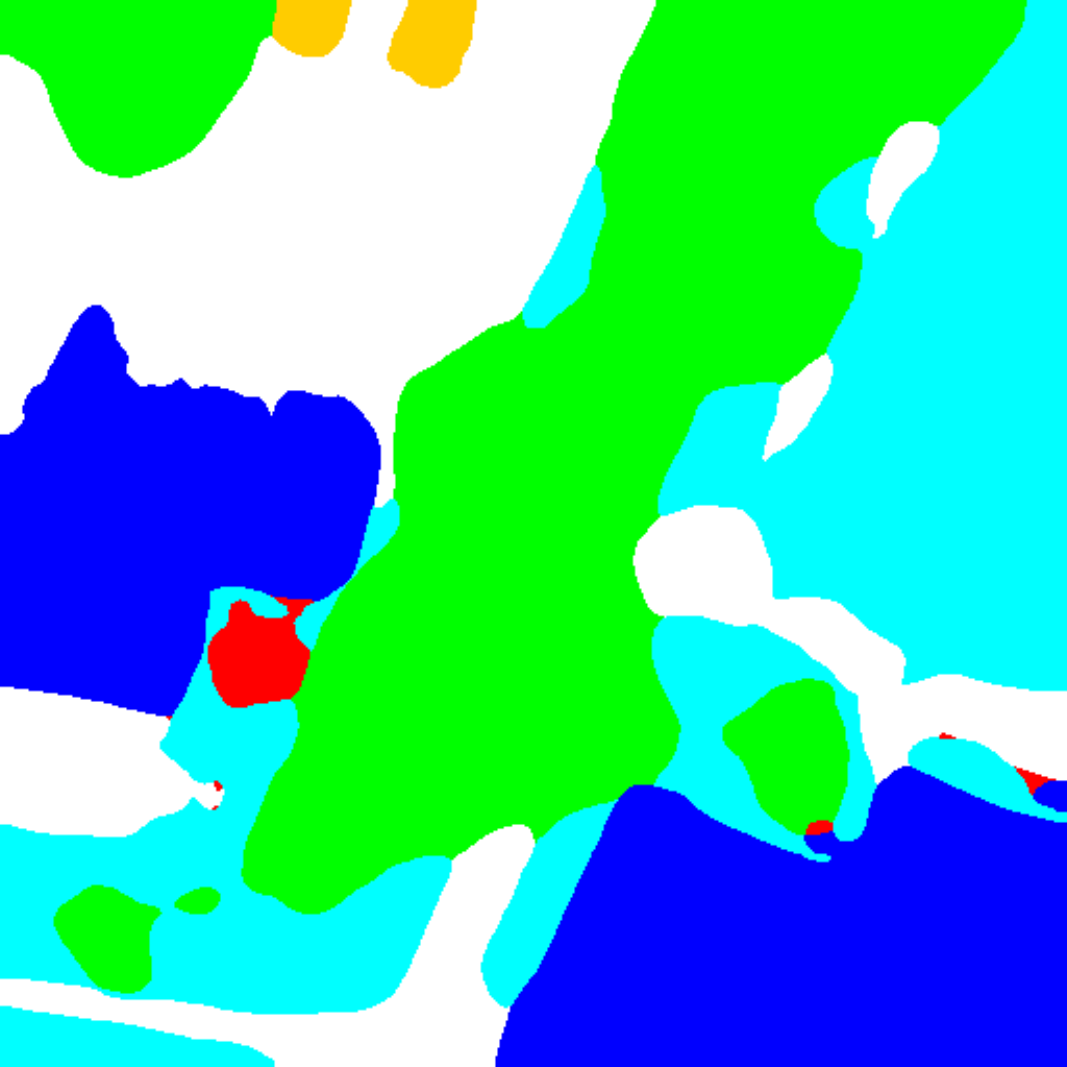}
		
		\vspace{1mm}
		
		\includegraphics[scale=0.094]{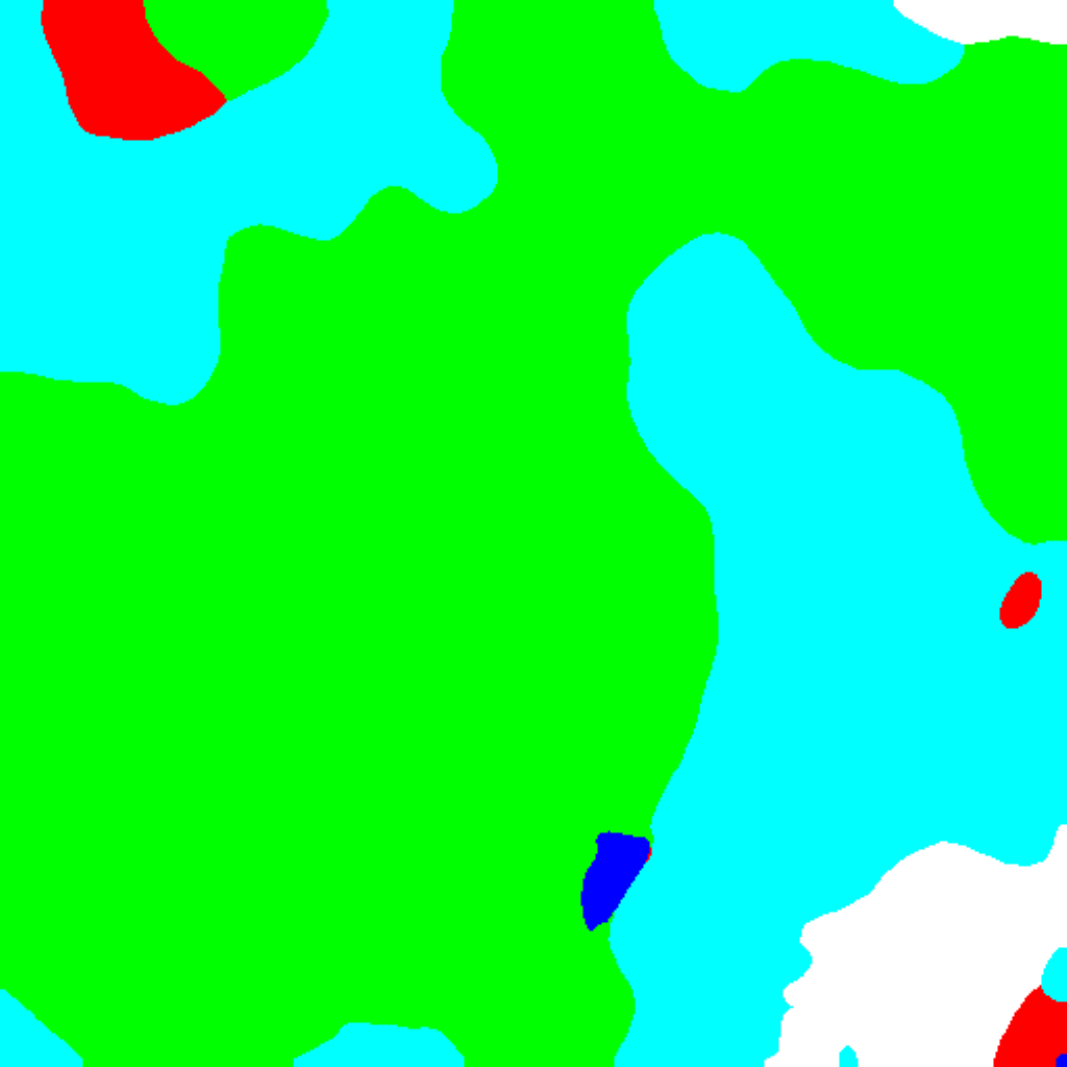}
		\centerline{(c)}
	\end{minipage}
	\begin{minipage}[t]{0.092\linewidth}
		\centering
		\includegraphics[scale=0.094]{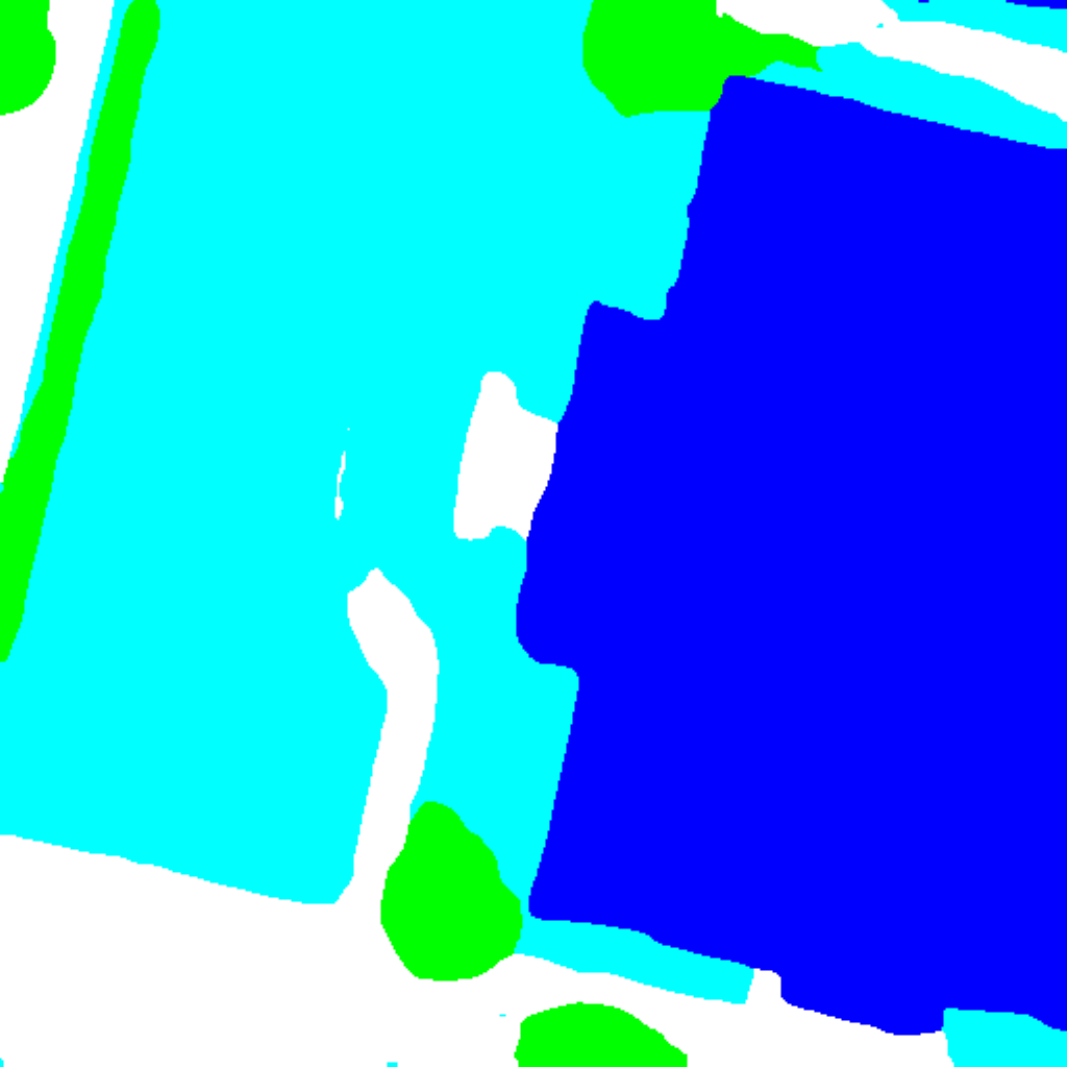}
		
		\vspace{1mm}
		
		\includegraphics[scale=0.094]{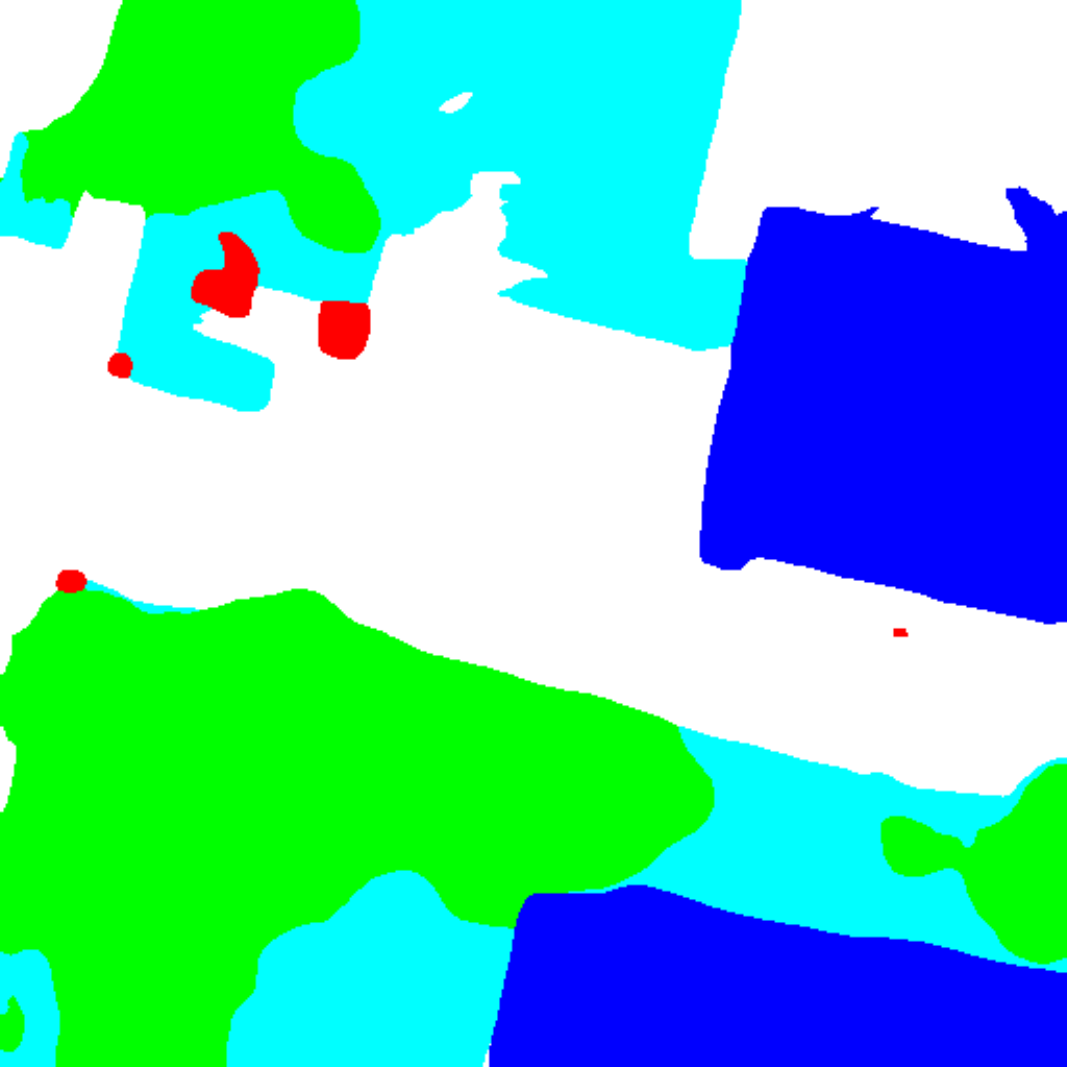}
		
		\vspace{1mm}
		
		\includegraphics[scale=0.094]{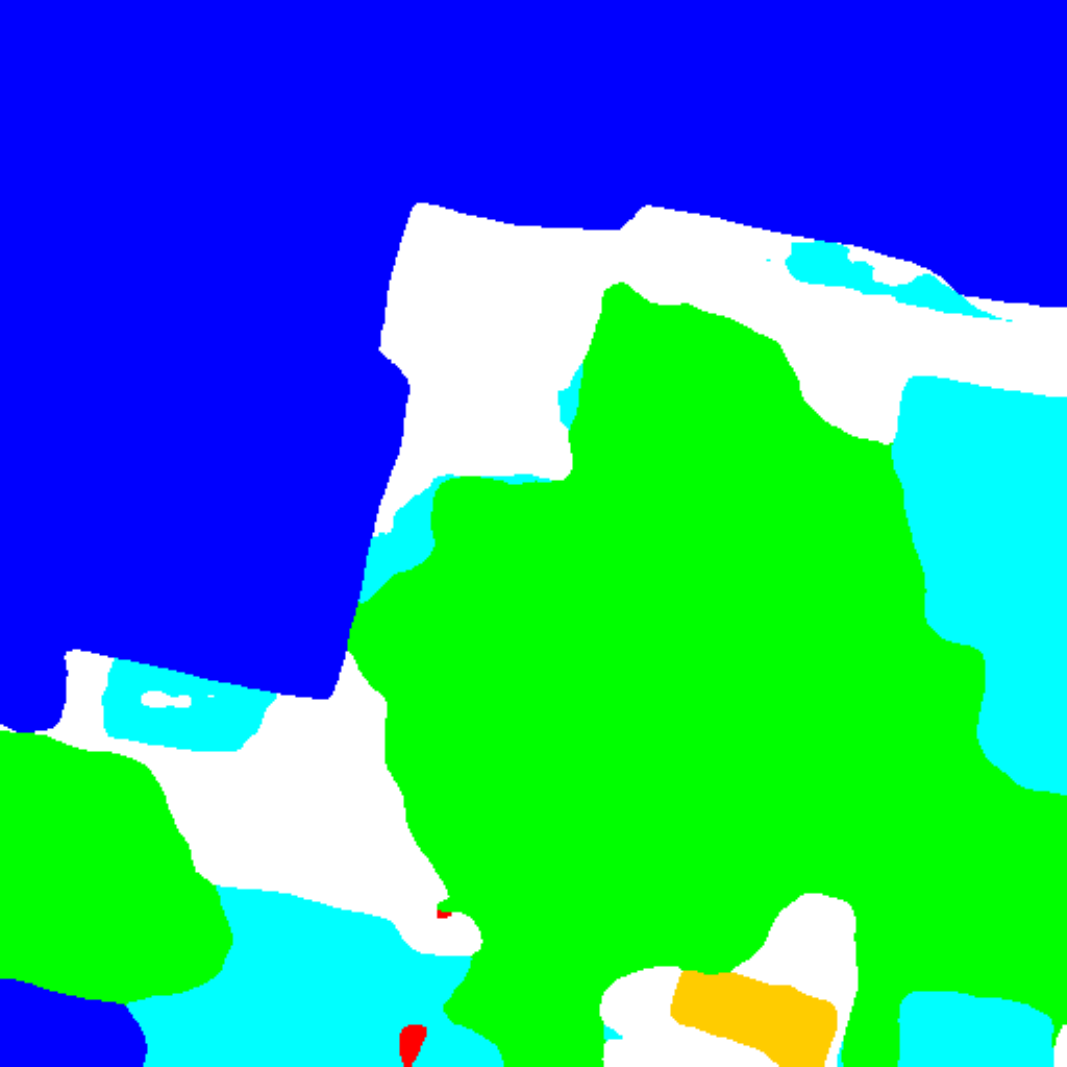}
		
		\vspace{1mm}
		
		\includegraphics[scale=0.094]{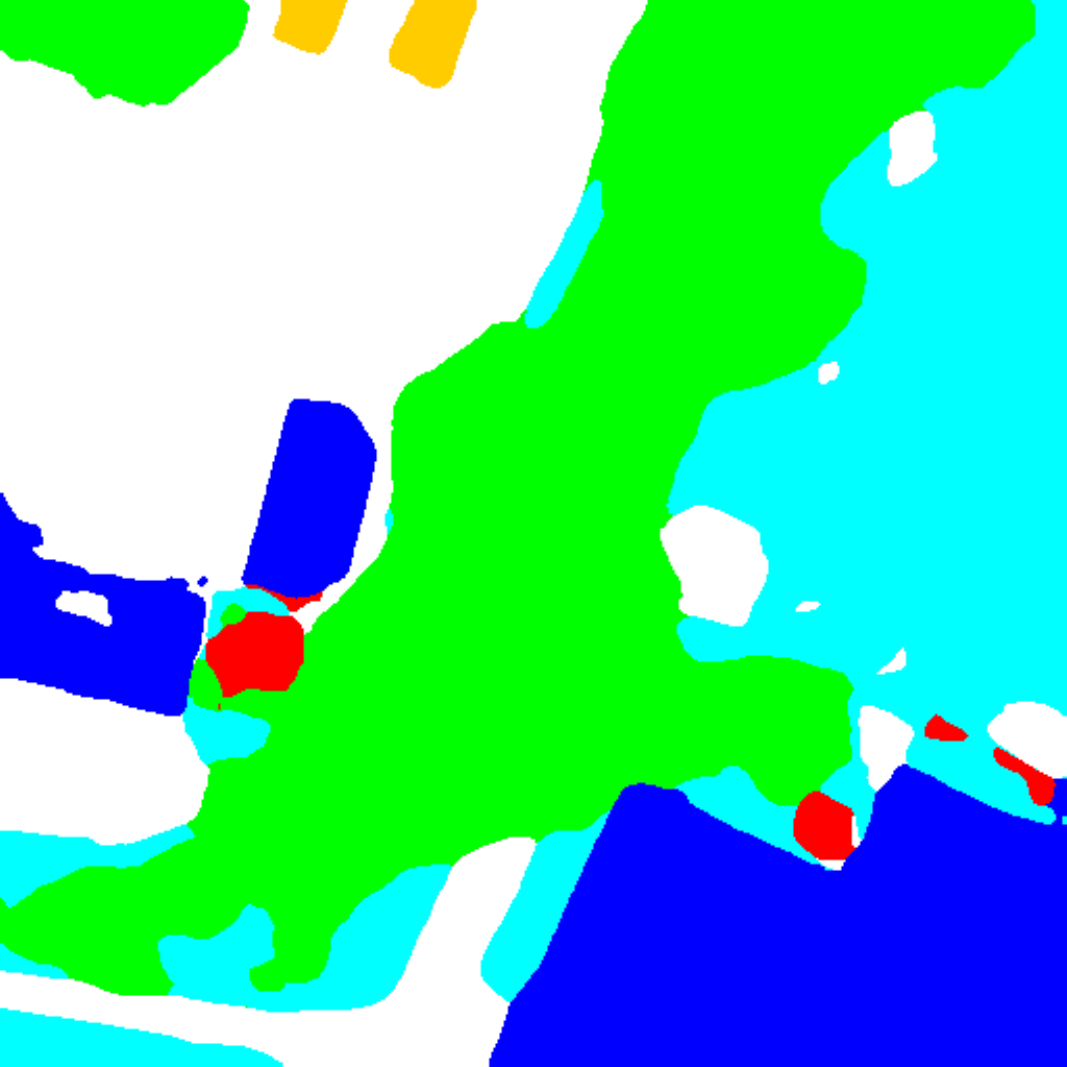}
		
		\vspace{1mm}
		
		\includegraphics[scale=0.094]{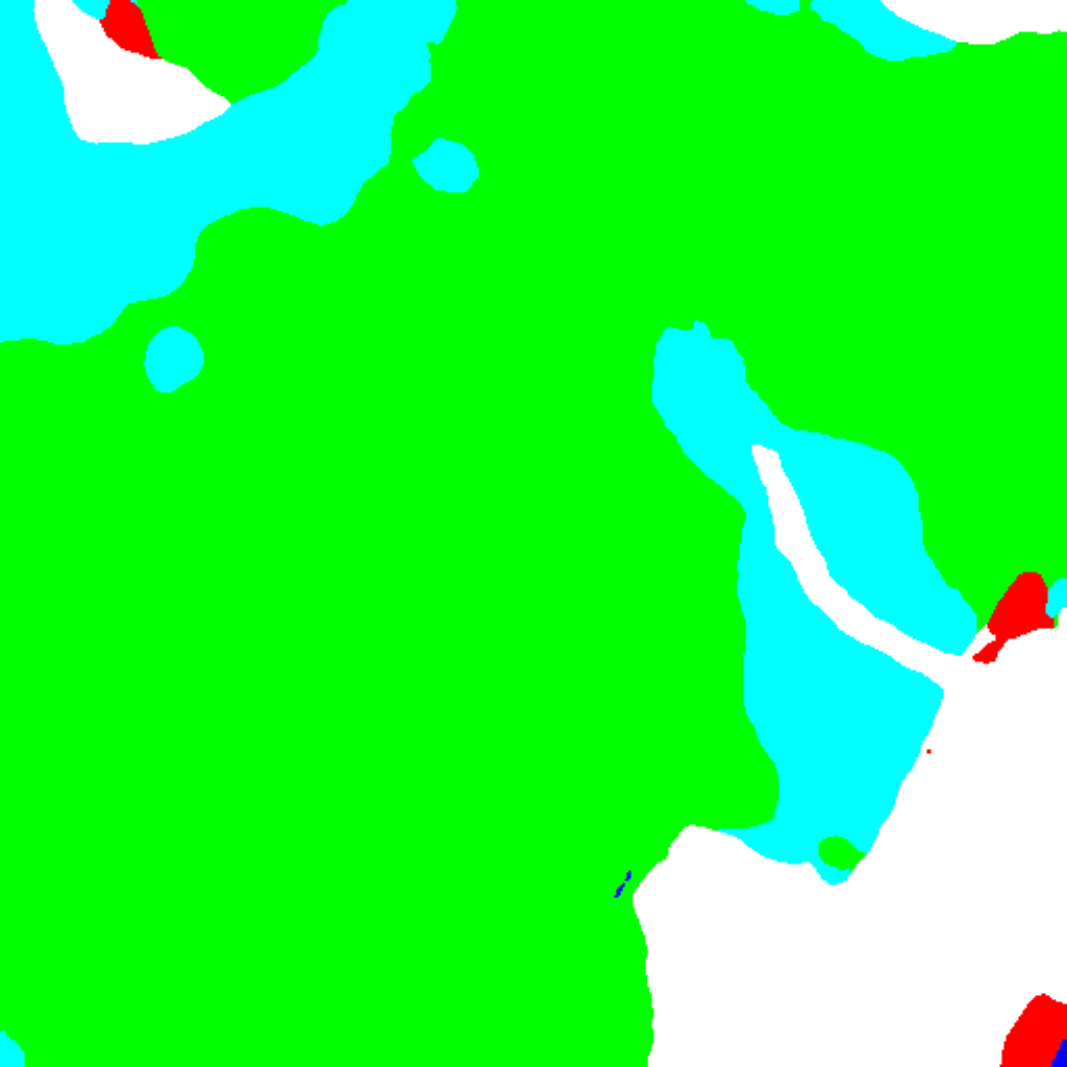}
		\centerline{(d)}
	\end{minipage}
	\begin{minipage}[t]{0.092\linewidth}
		\centering
		\includegraphics[scale=0.094]{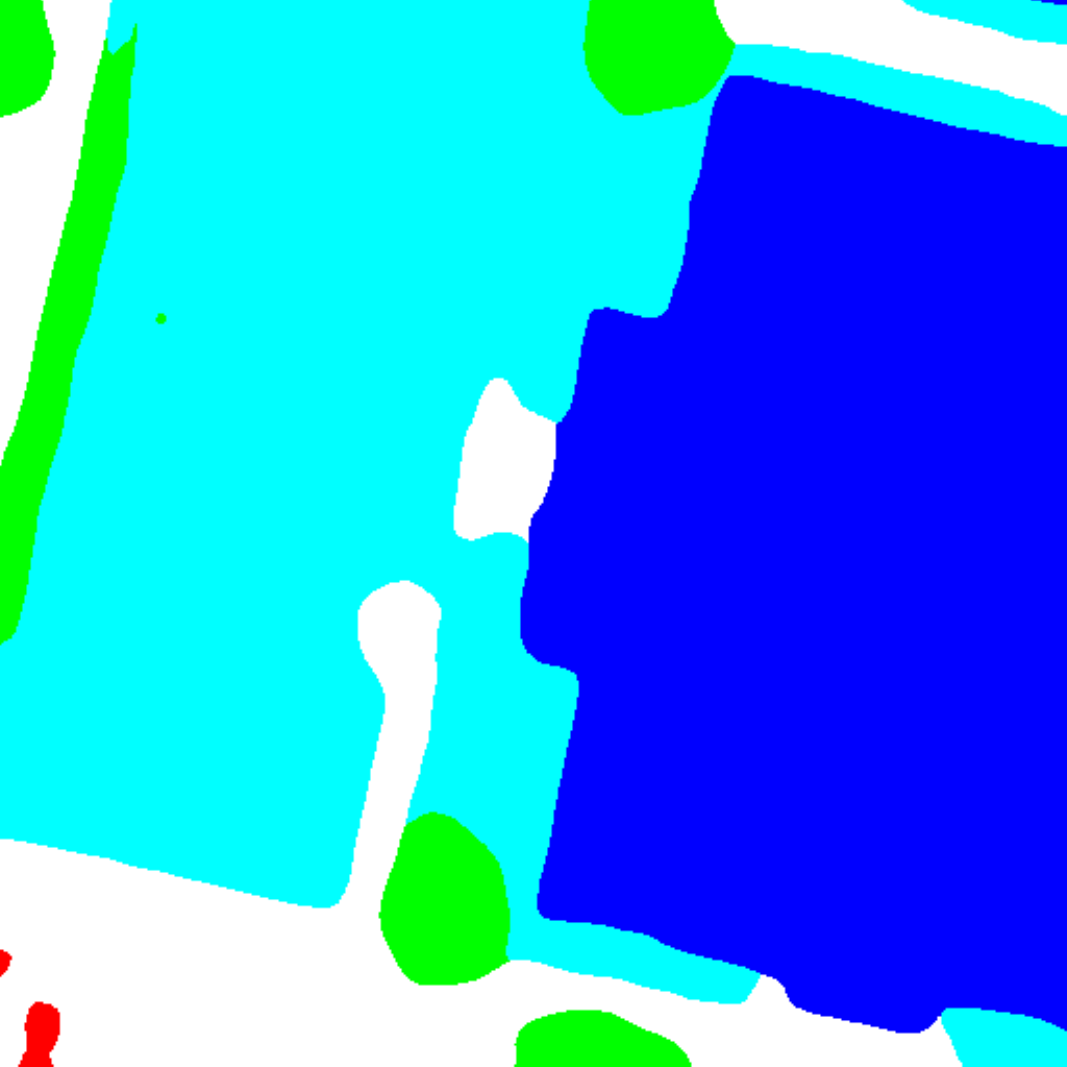}
		
		\vspace{1mm}
		
		\includegraphics[scale=0.094]{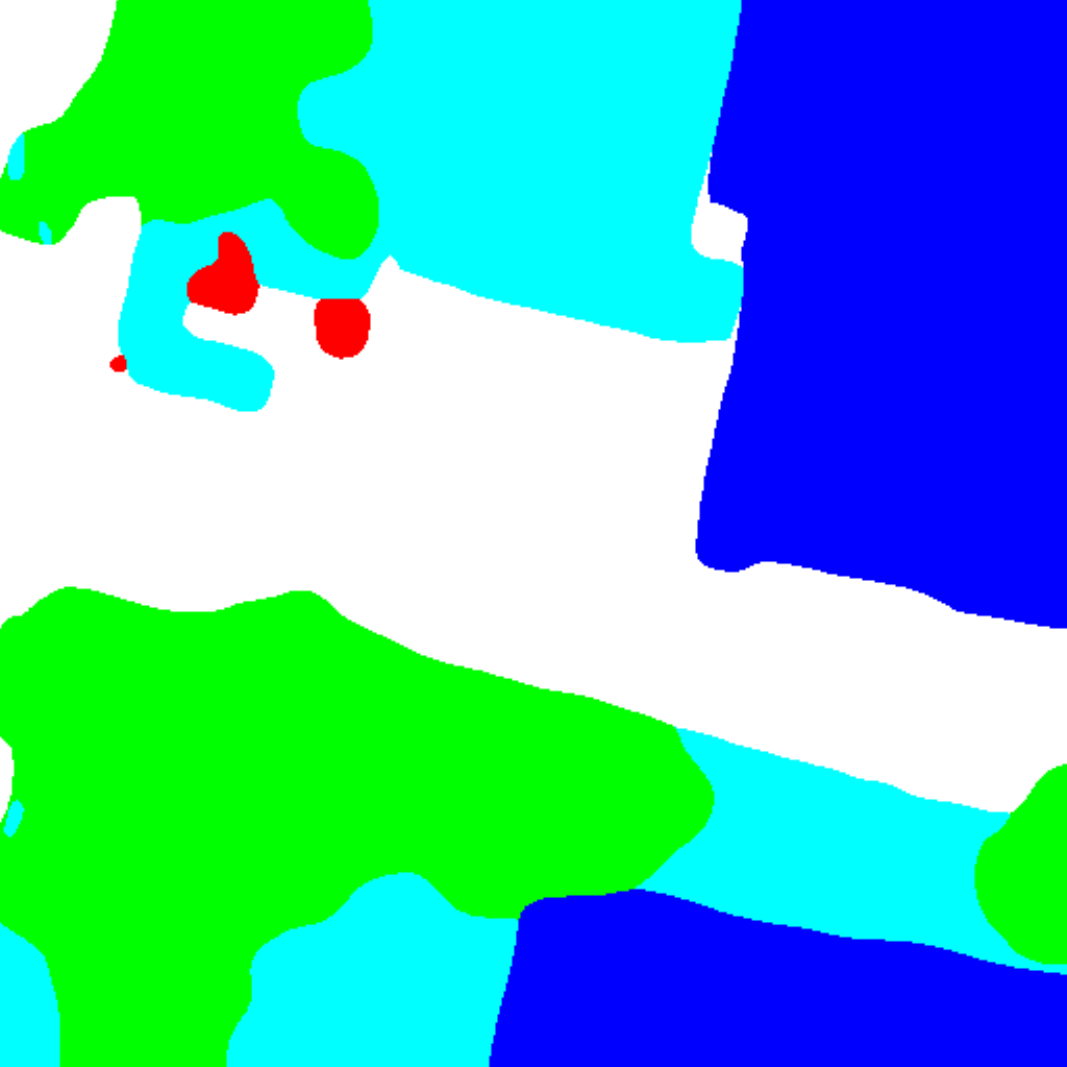}
		
		\vspace{1mm}
		
		\includegraphics[scale=0.094]{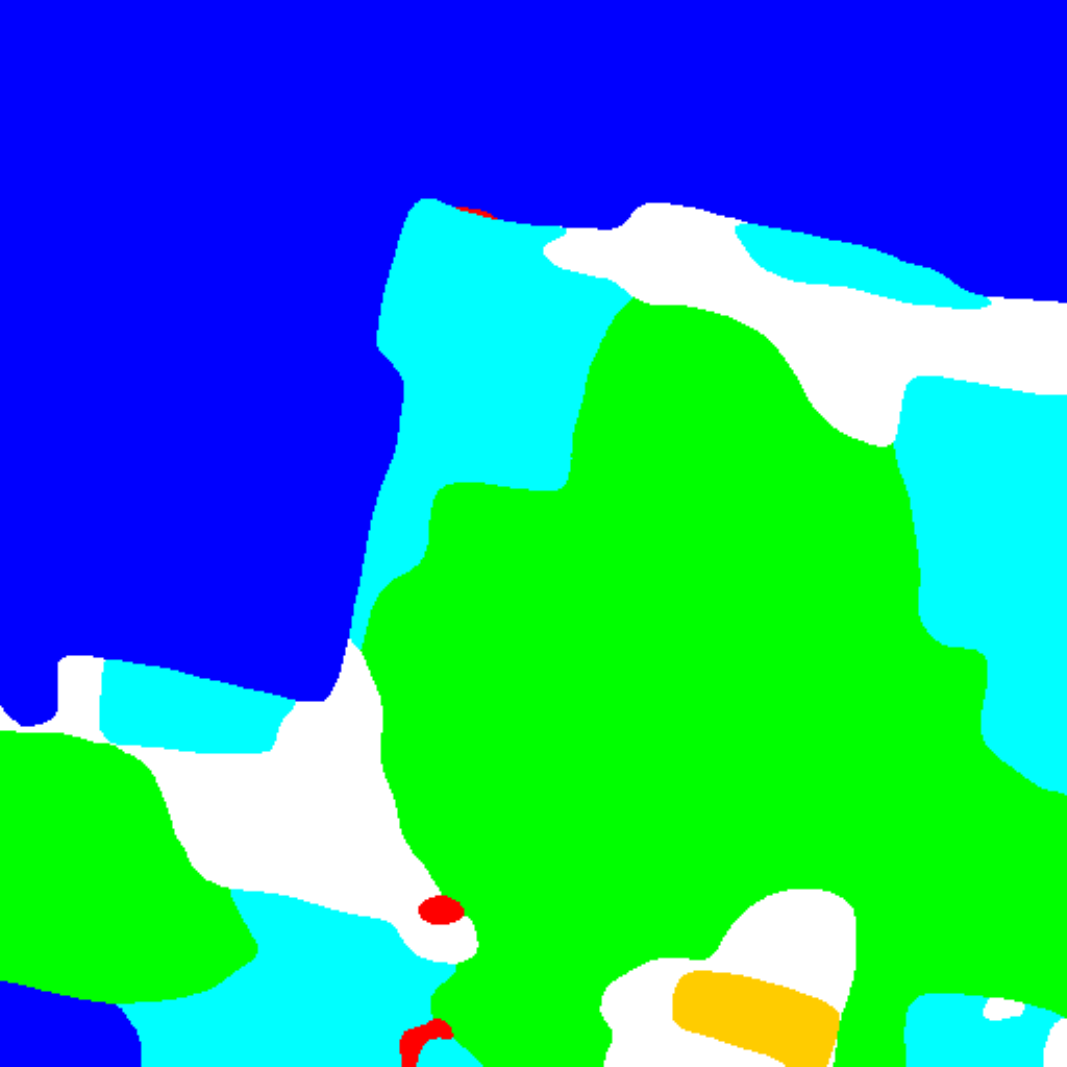}
		
		\vspace{1mm}
		
		\includegraphics[scale=0.094]{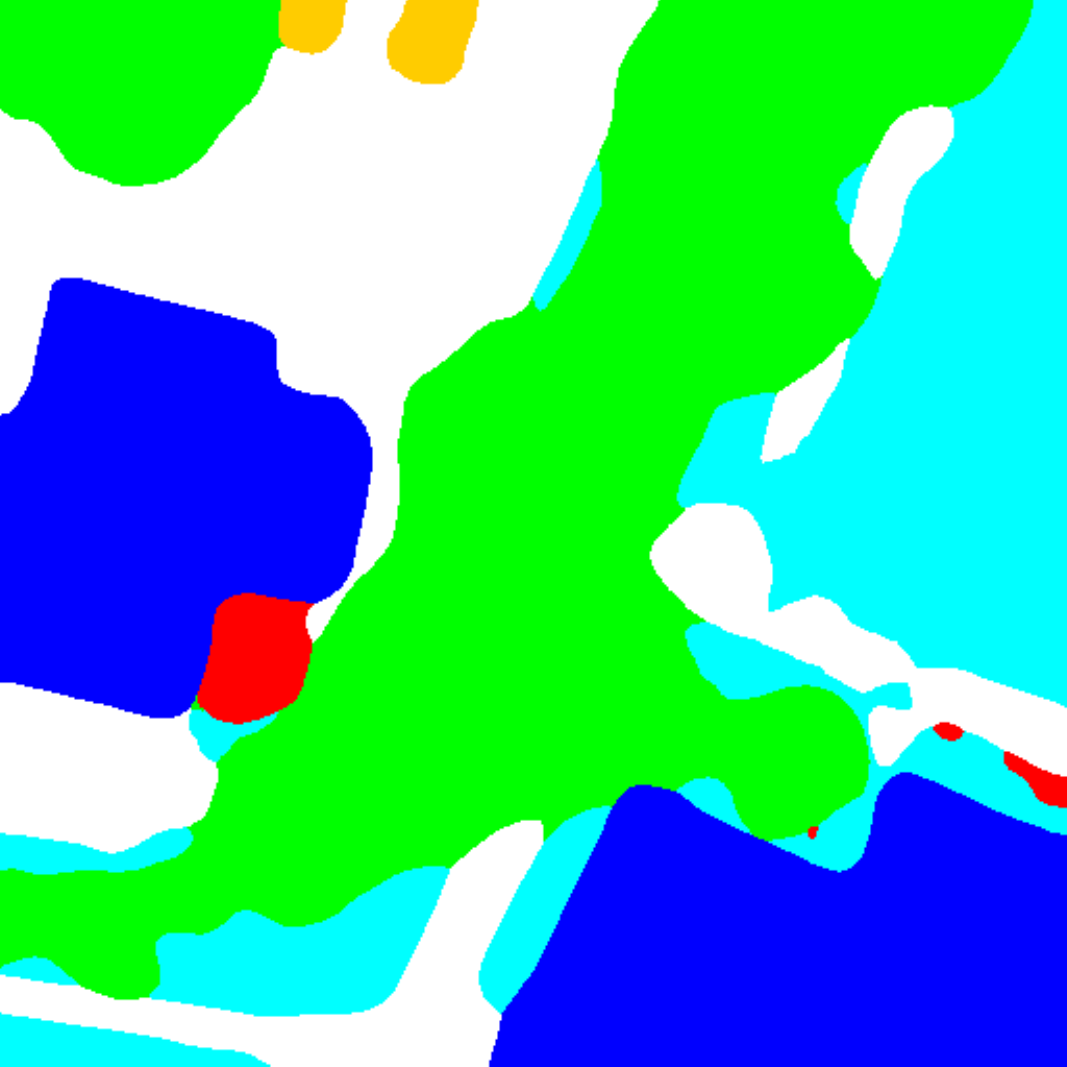}
		
		\vspace{1mm}
		
		\includegraphics[scale=0.094]{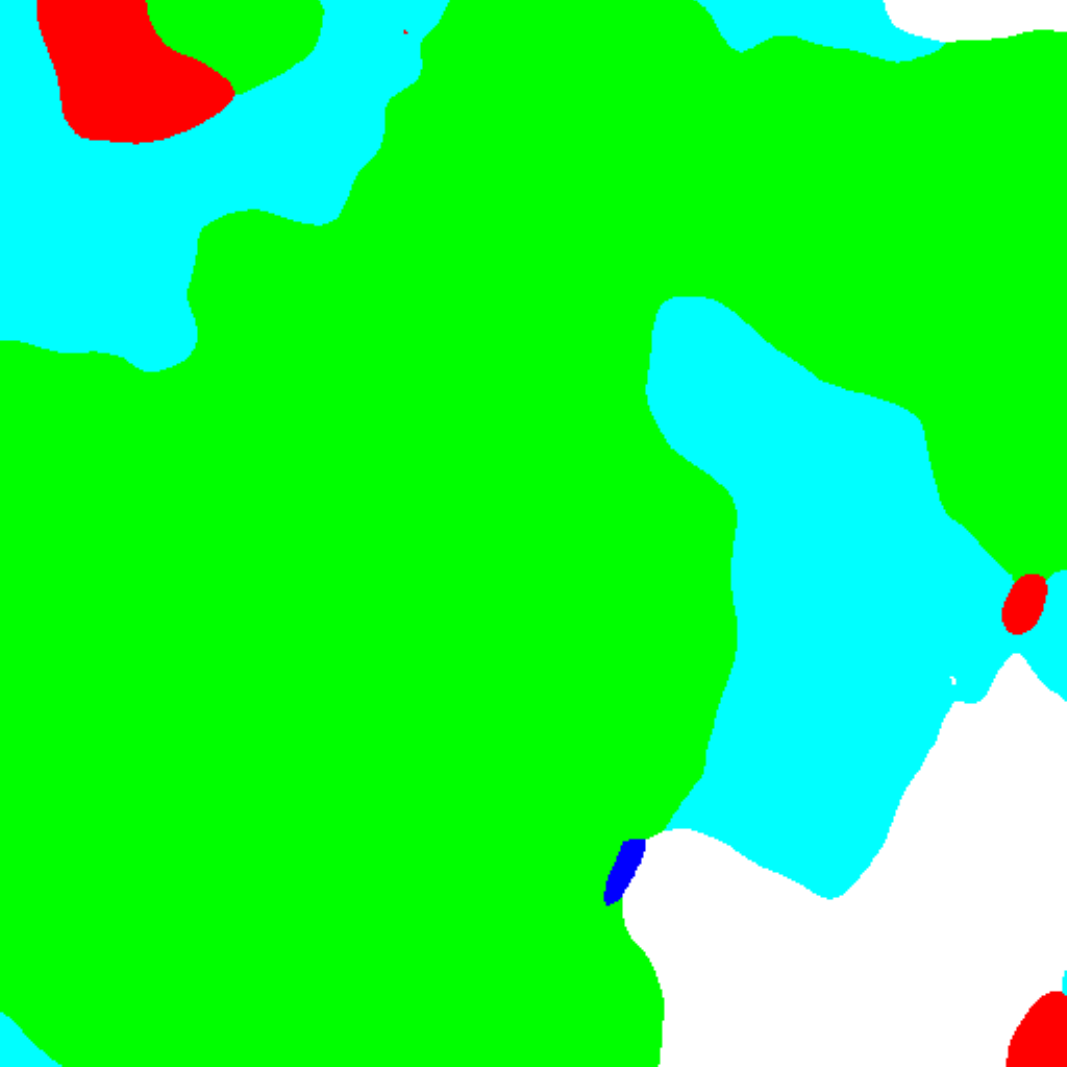}
		\centerline{(e)}
	\end{minipage}
	\begin{minipage}[t]{0.092\linewidth}
		\centering
		\includegraphics[scale=0.094]{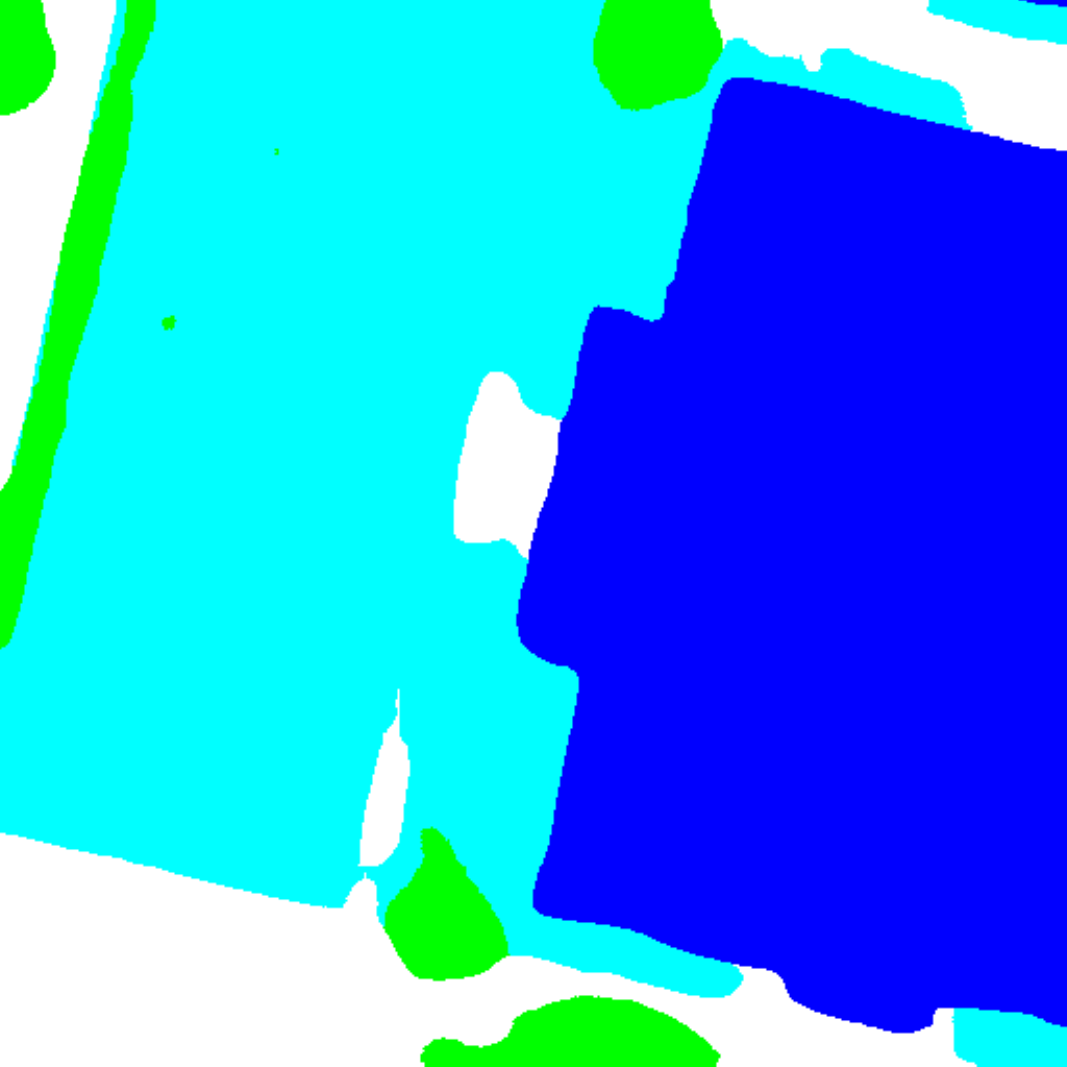}
		
		\vspace{1mm}
		
		\includegraphics[scale=0.094]{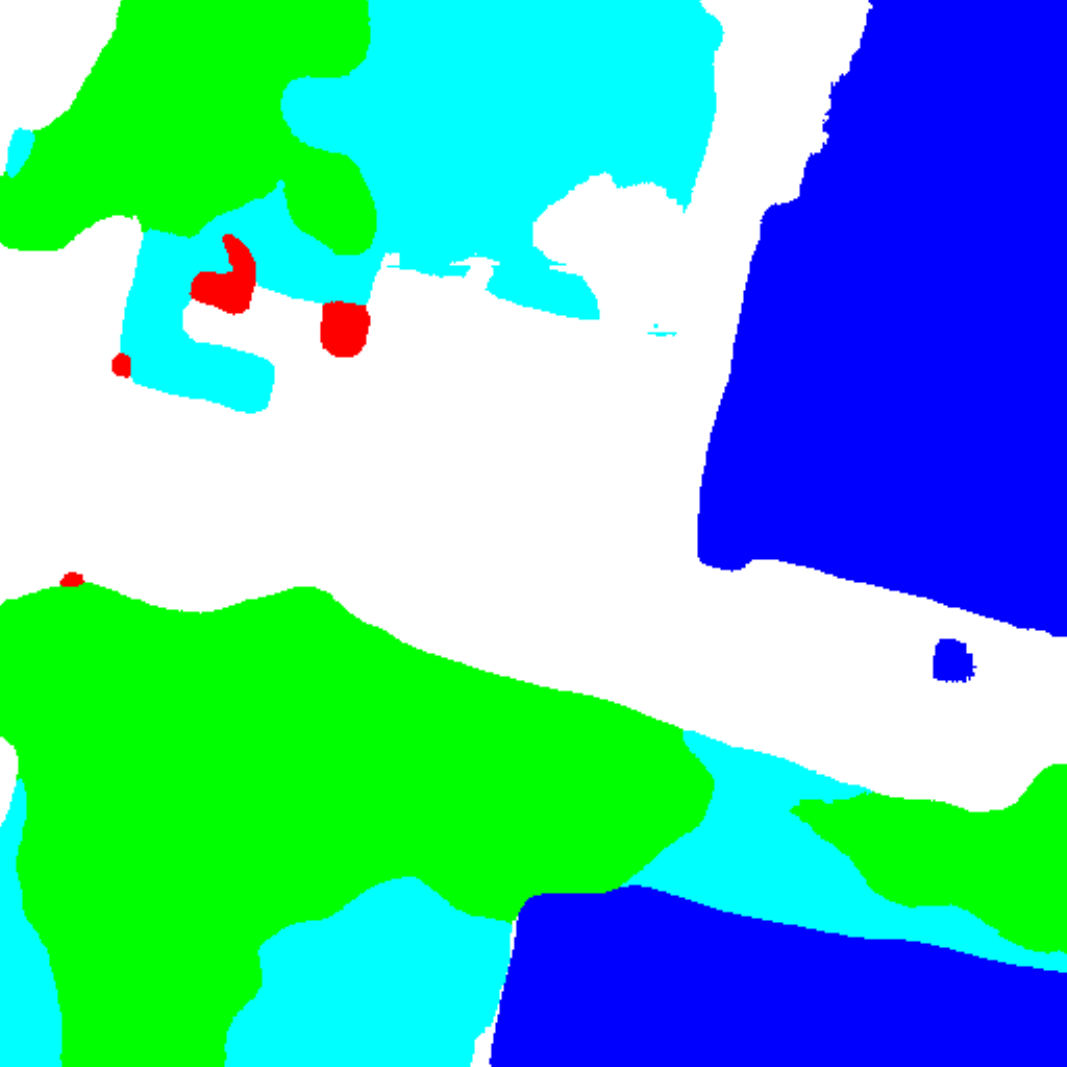}
		
		\vspace{1mm}
		
		\includegraphics[scale=0.094]{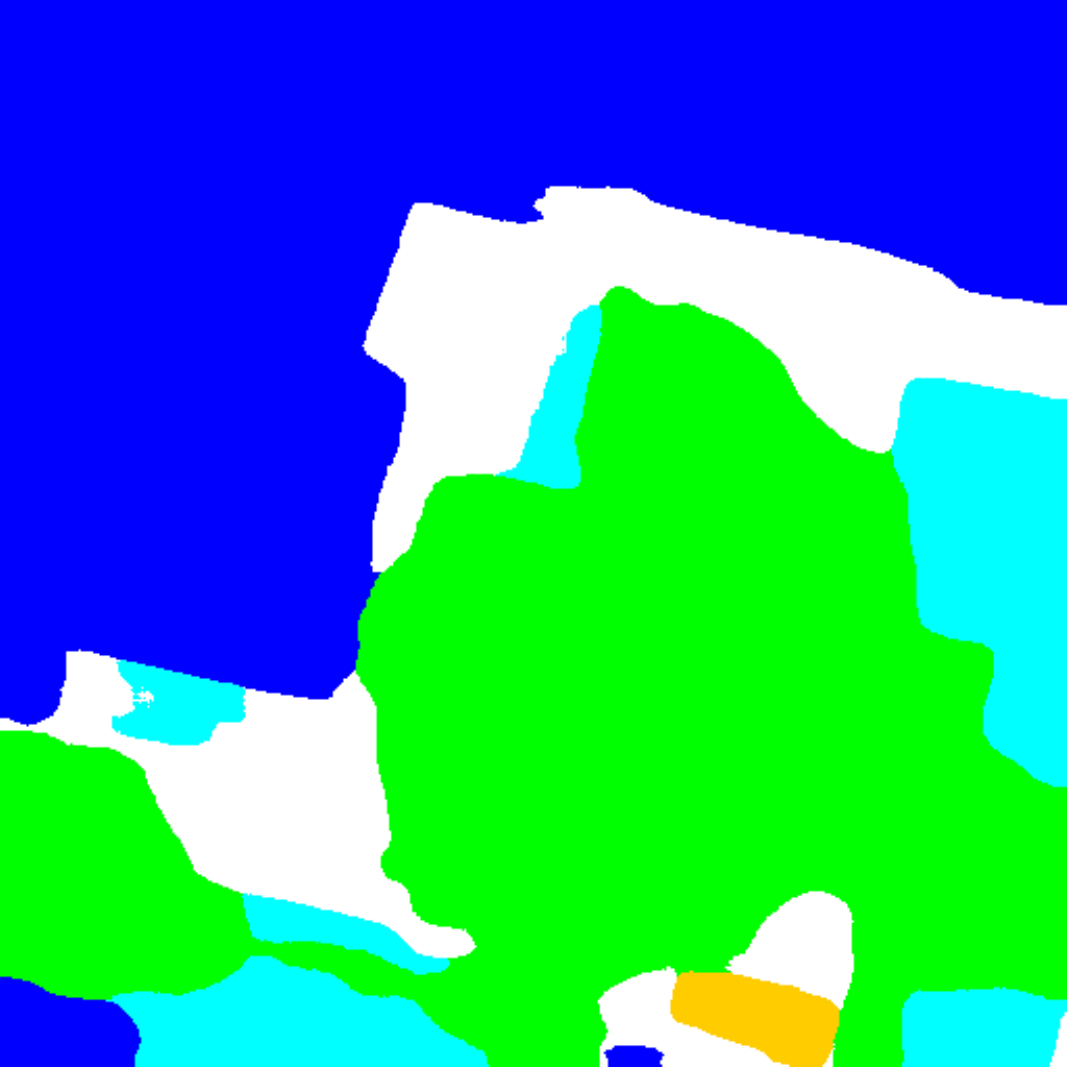}
		
		\vspace{1mm}
		
		\includegraphics[scale=0.094]{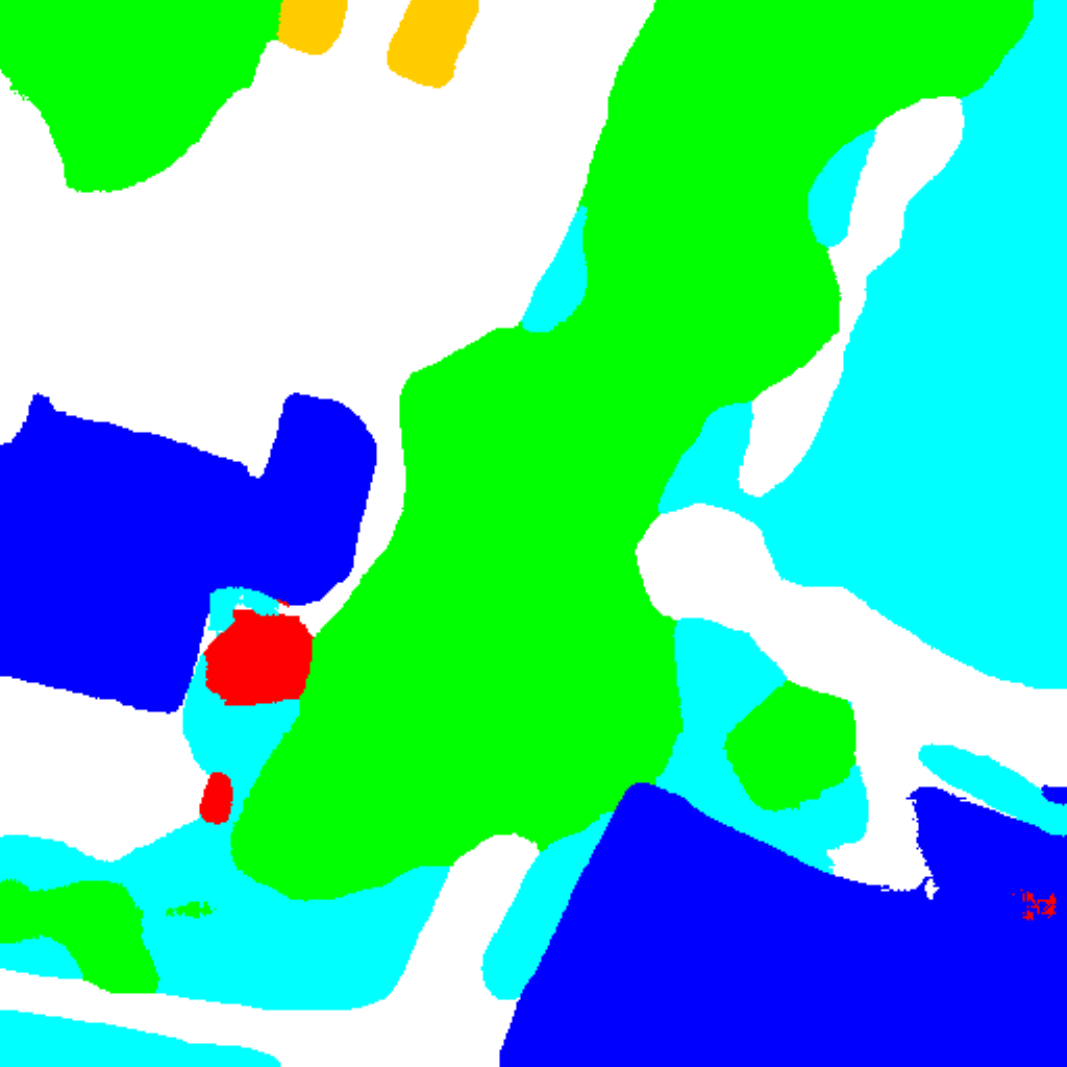}
		
		\vspace{1mm}
		
		\includegraphics[scale=0.094]{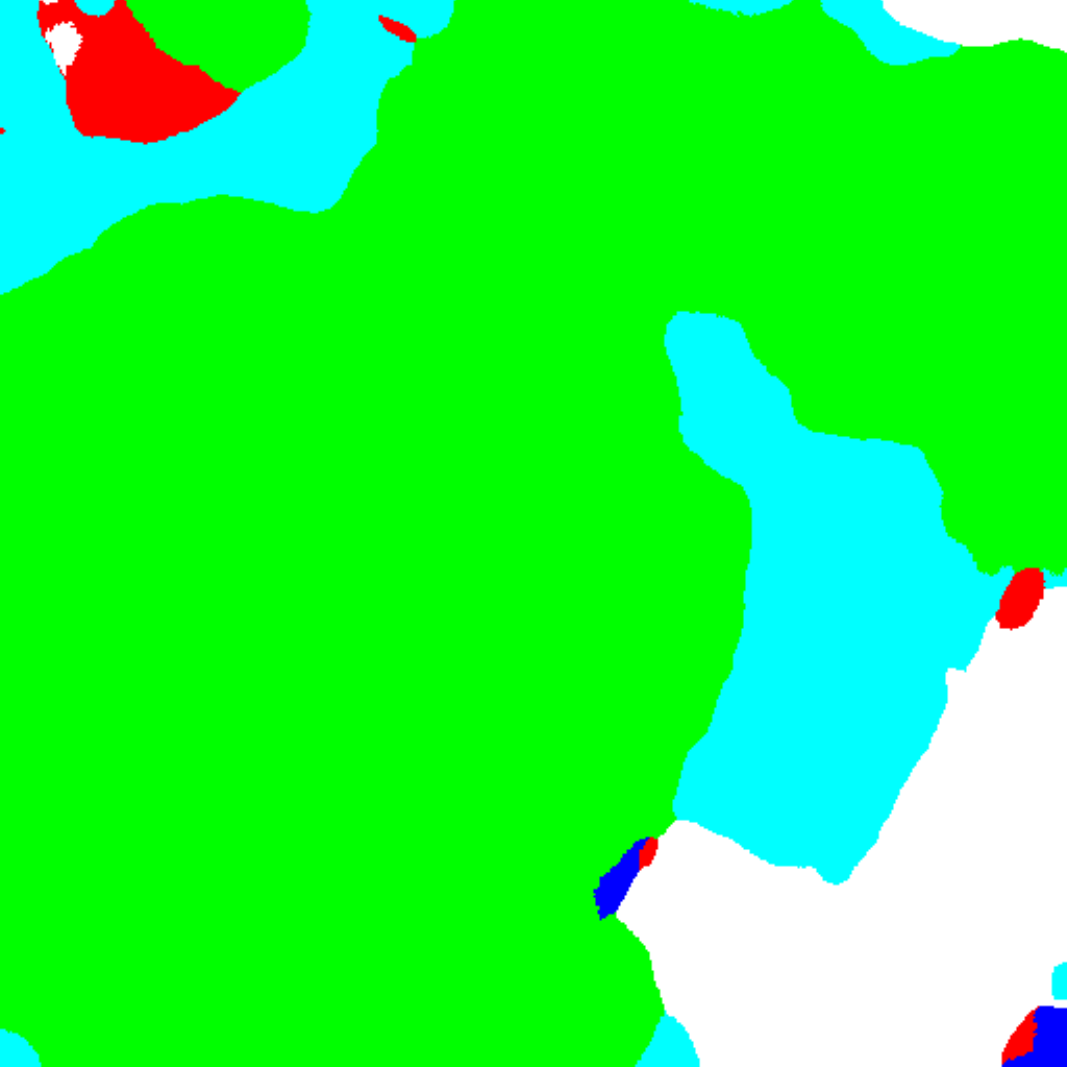}
		\centerline{(f)}
	\end{minipage}
	\begin{minipage}[t]{0.092\linewidth}
		\centering
		\includegraphics[scale=0.094]{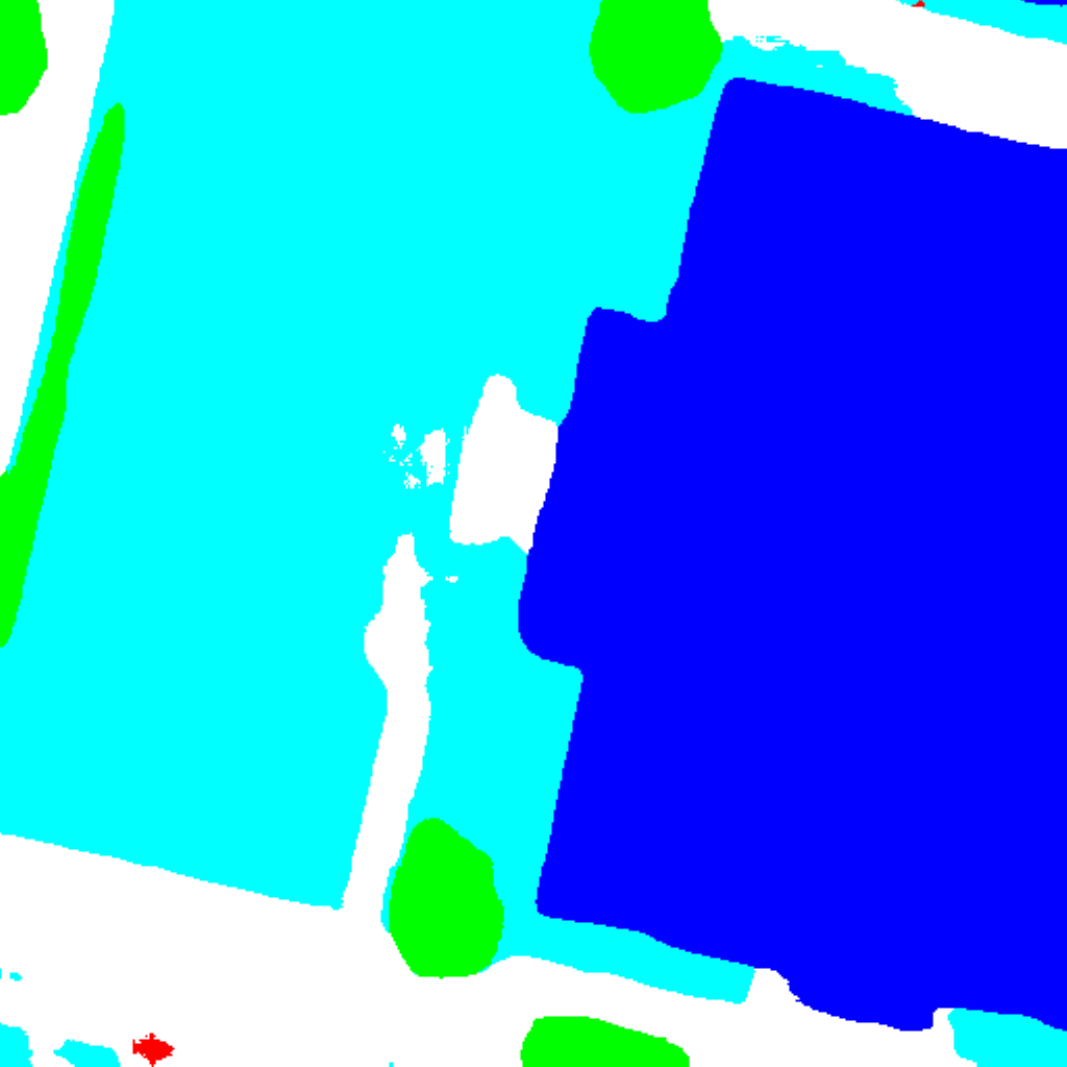}
		
		\vspace{1mm}
		
		\includegraphics[scale=0.094]{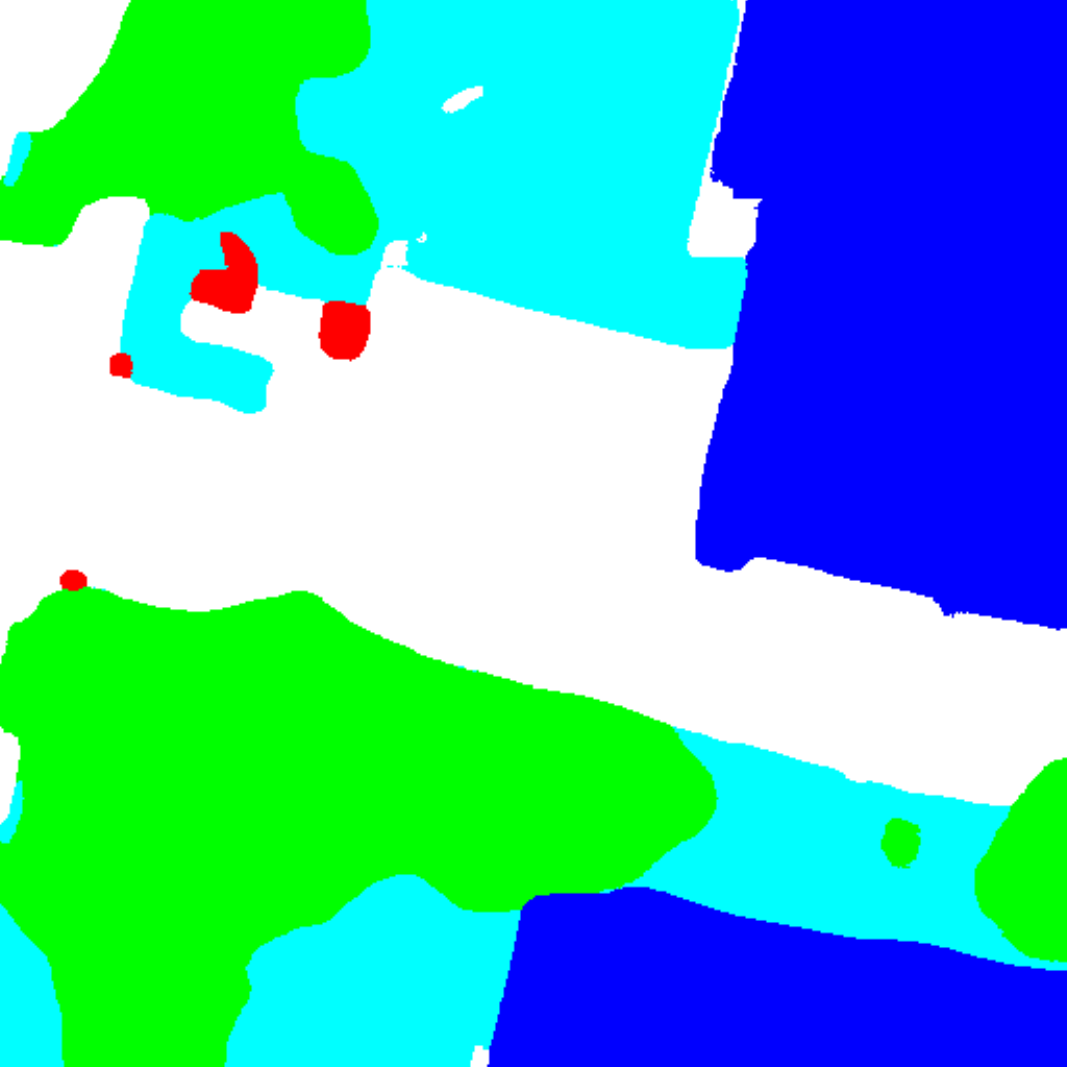}
		
		\vspace{1mm}
		
		\includegraphics[scale=0.094]{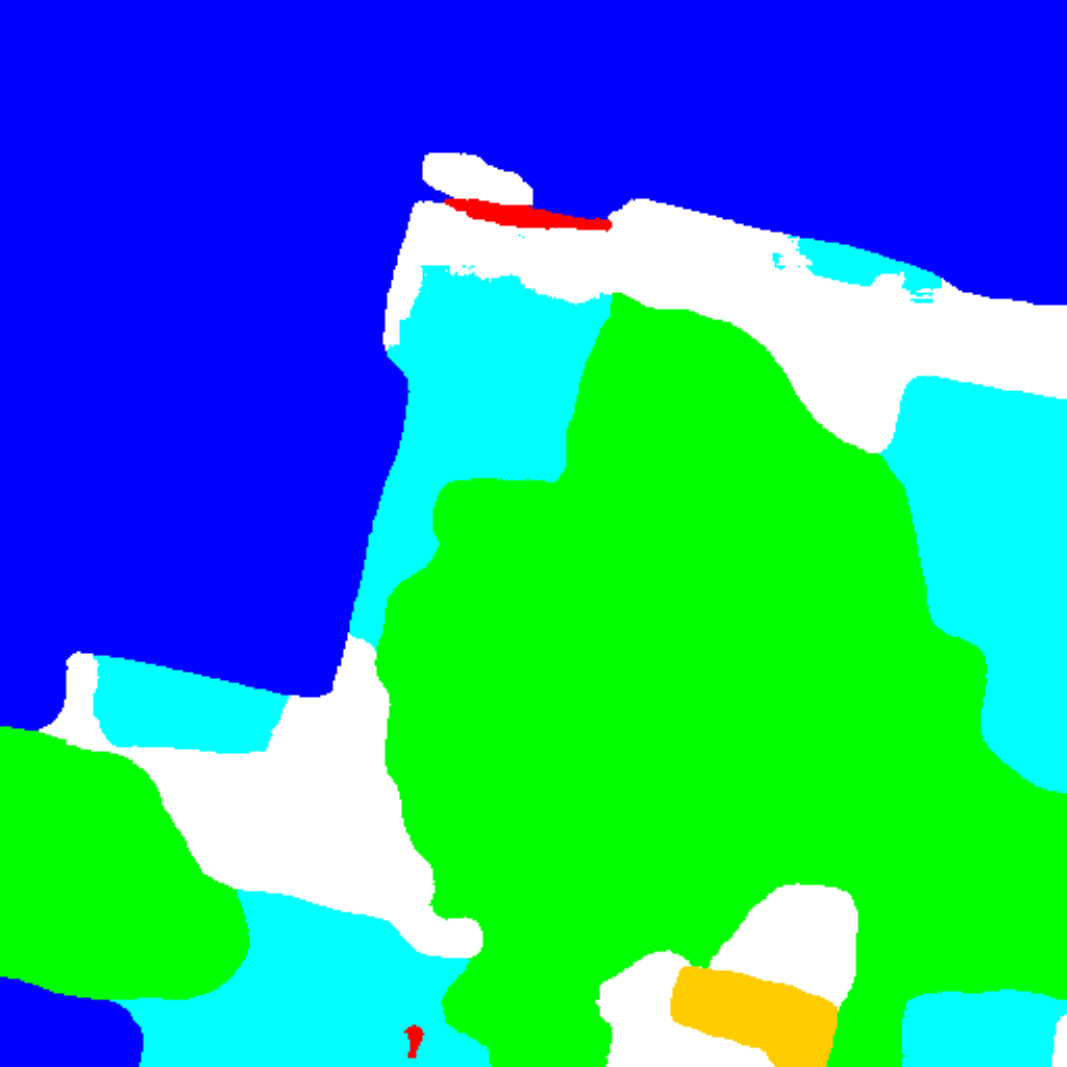}
		
		\vspace{1mm}
		
		\includegraphics[scale=0.094]{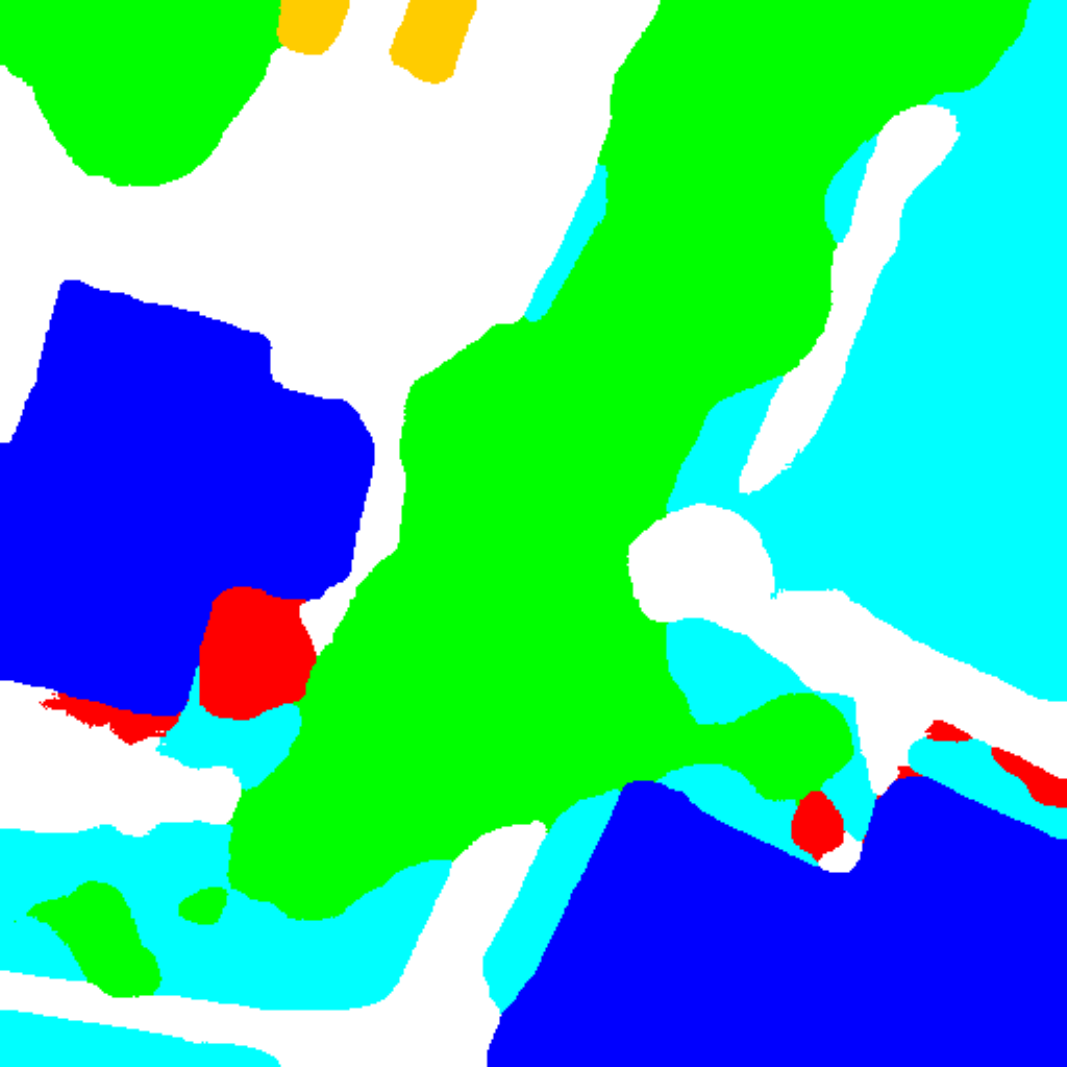}
		
		\vspace{1mm}
		
		\includegraphics[scale=0.094]{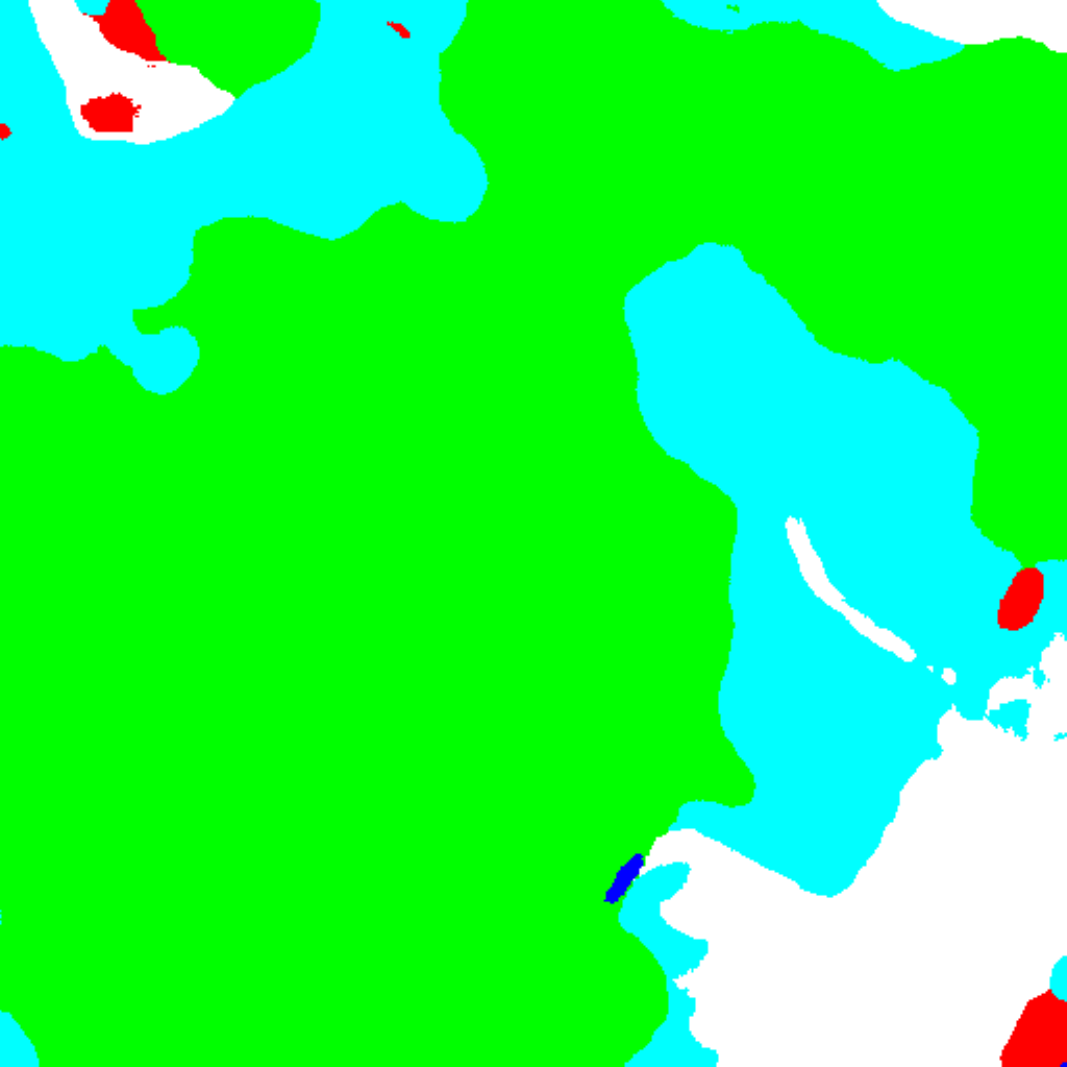}
		\centerline{(g)}
	\end{minipage}
	\begin{minipage}[t]{0.092\linewidth}
		\centering
		\includegraphics[scale=0.094]{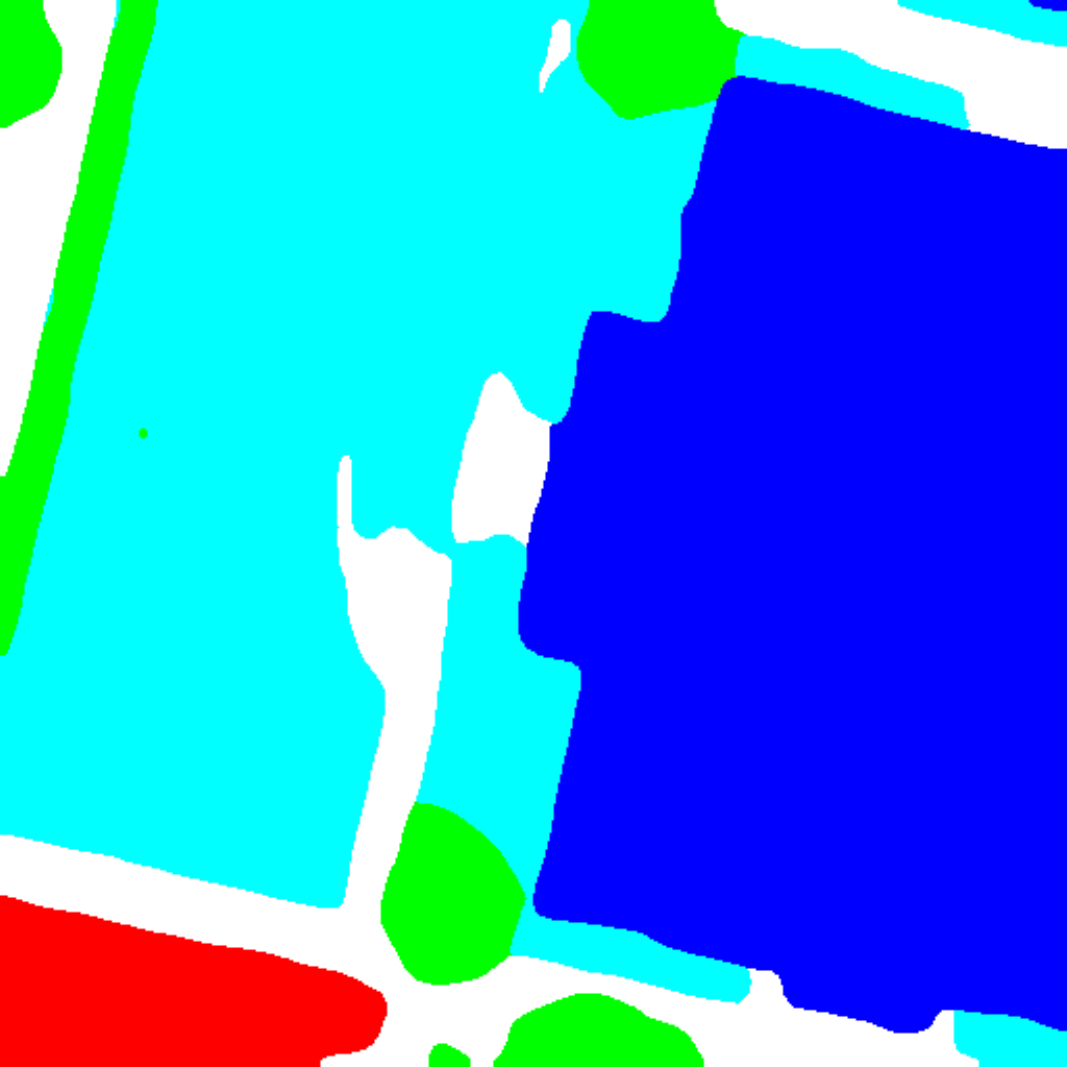}
		
		\vspace{1mm}
		
		\includegraphics[scale=0.094]{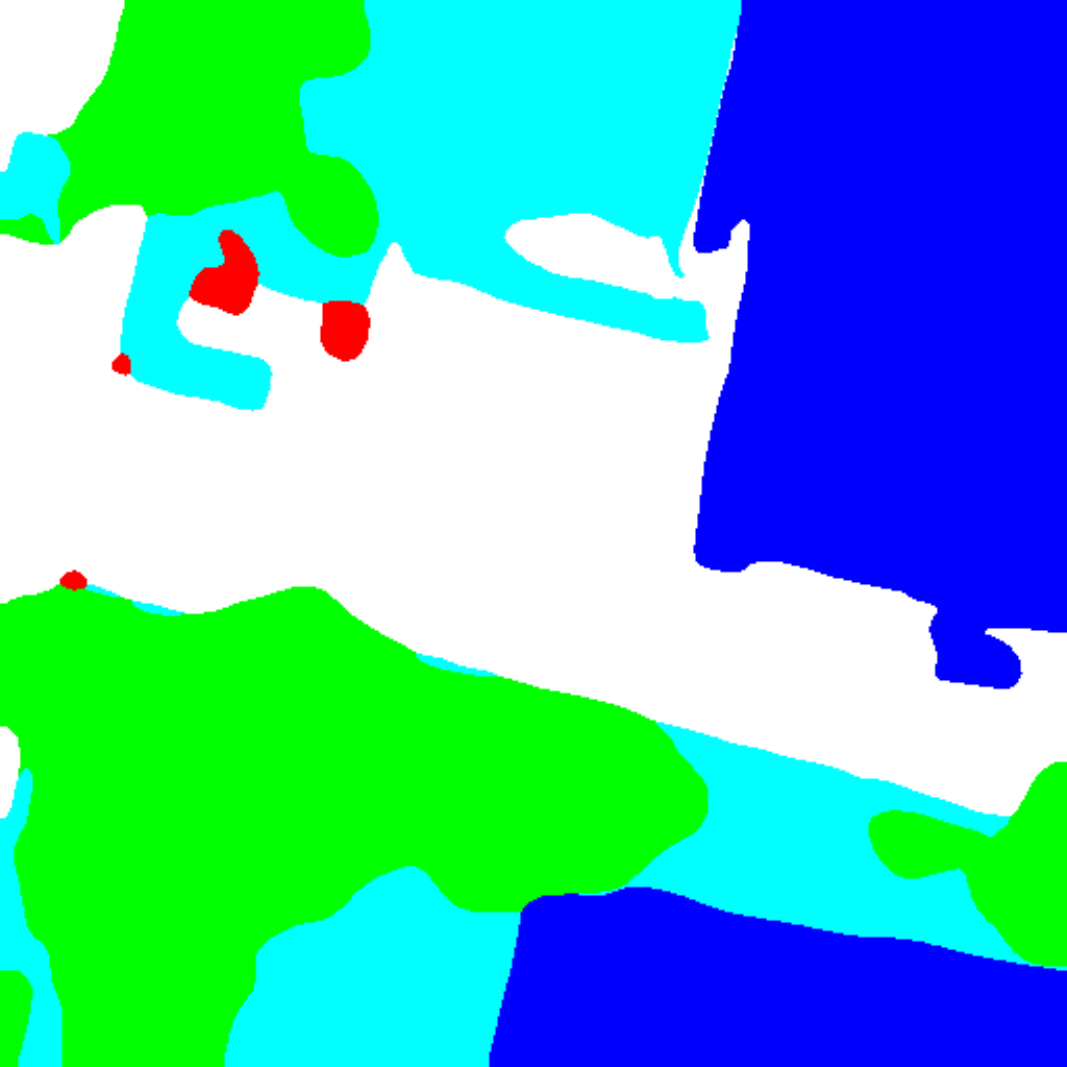}
		
		\vspace{1mm}
		
		\includegraphics[scale=0.094]{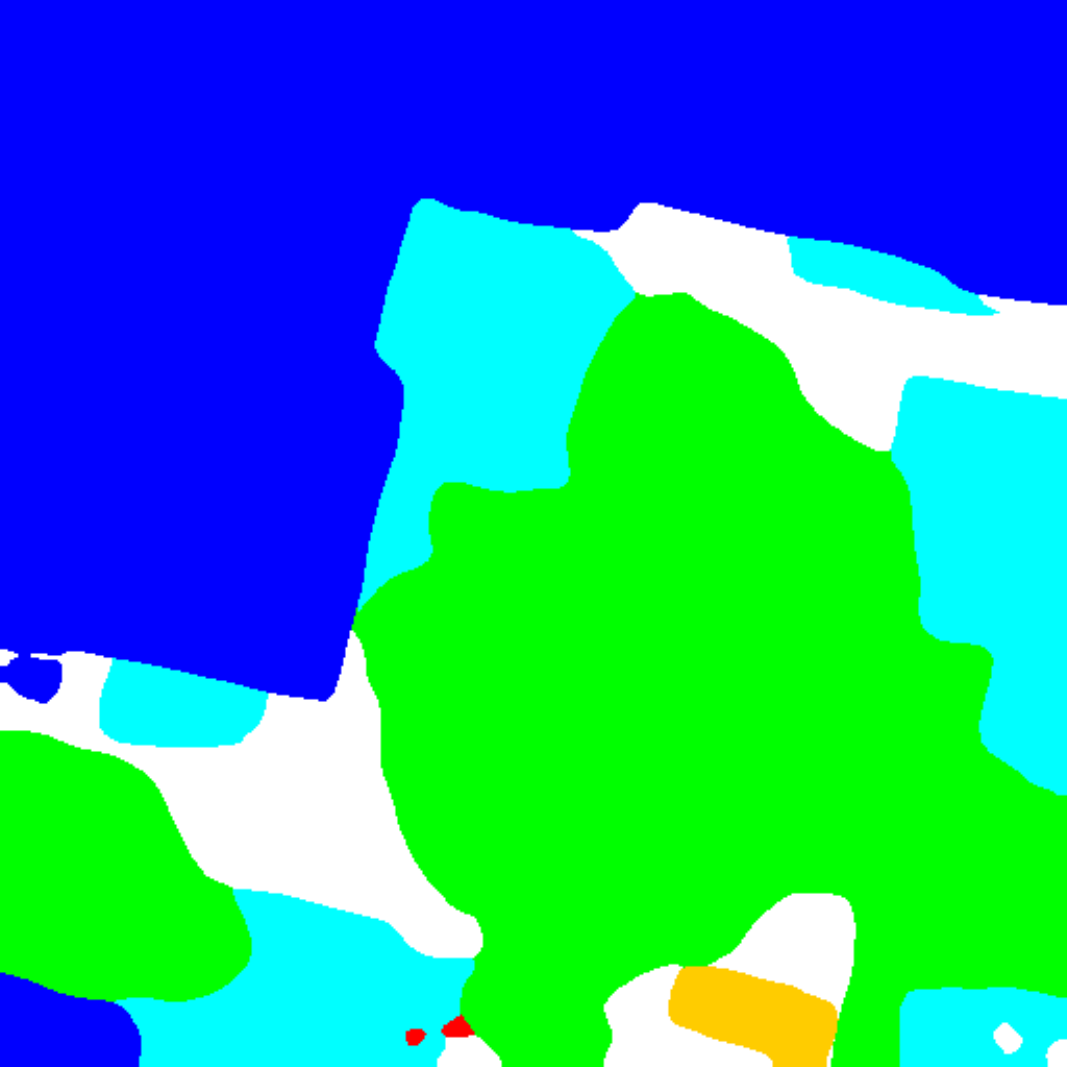}
		
		\vspace{1mm}
		
		\includegraphics[scale=0.094]{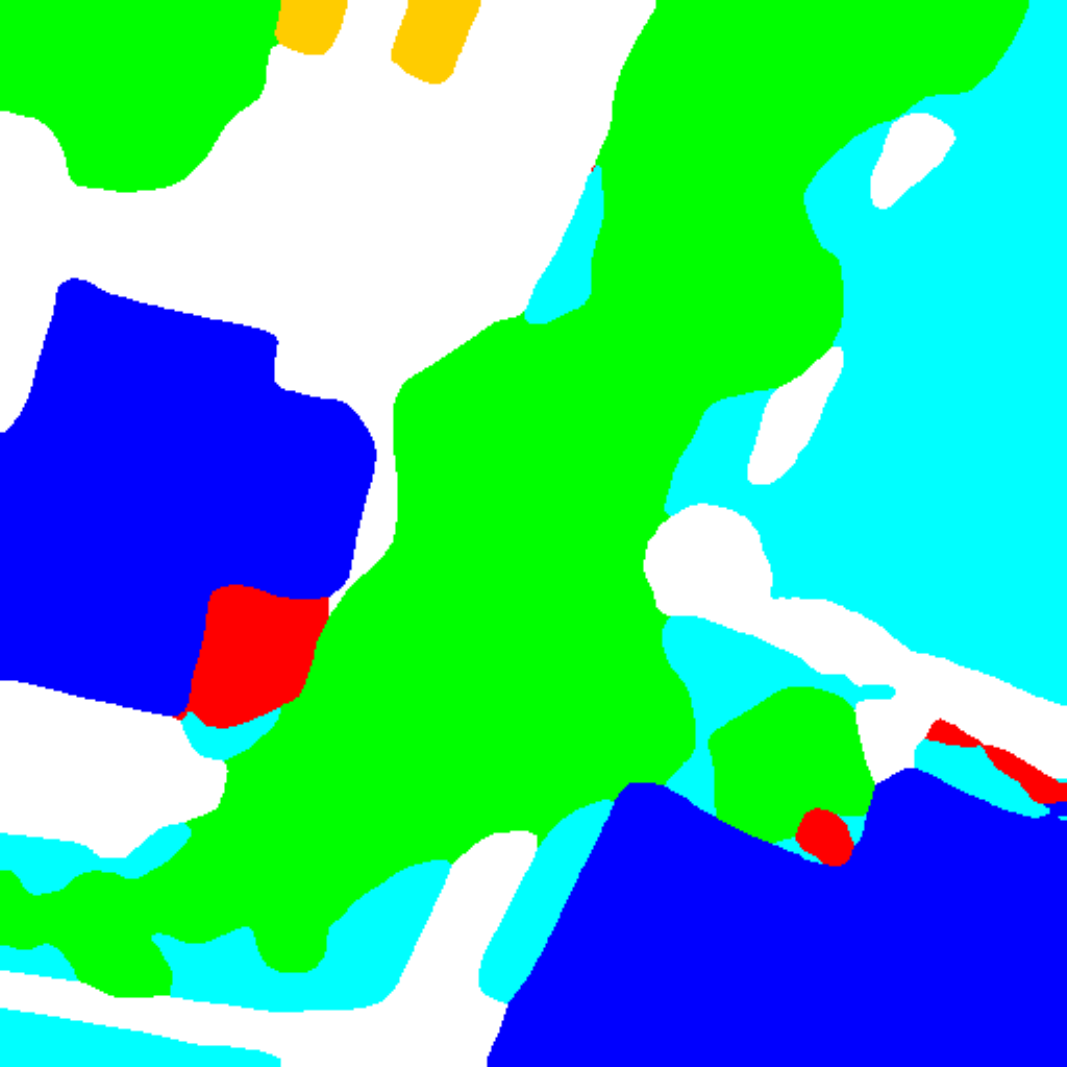}
		
		\vspace{1mm}
		
		\includegraphics[scale=0.094]{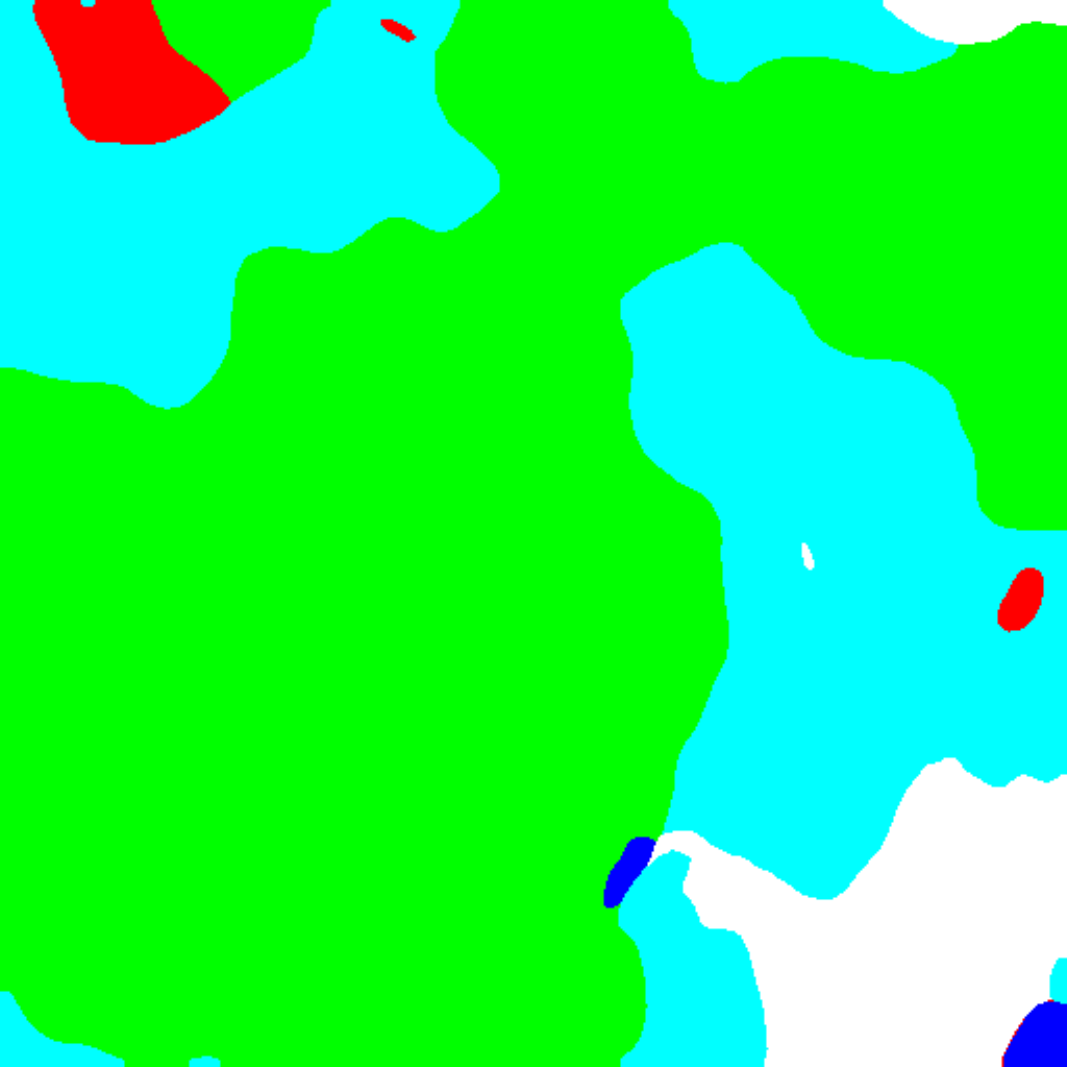}
		\centerline{(h)}
	\end{minipage}
	\begin{minipage}[t]{0.092\linewidth}
		\centering
		\includegraphics[scale=0.094]{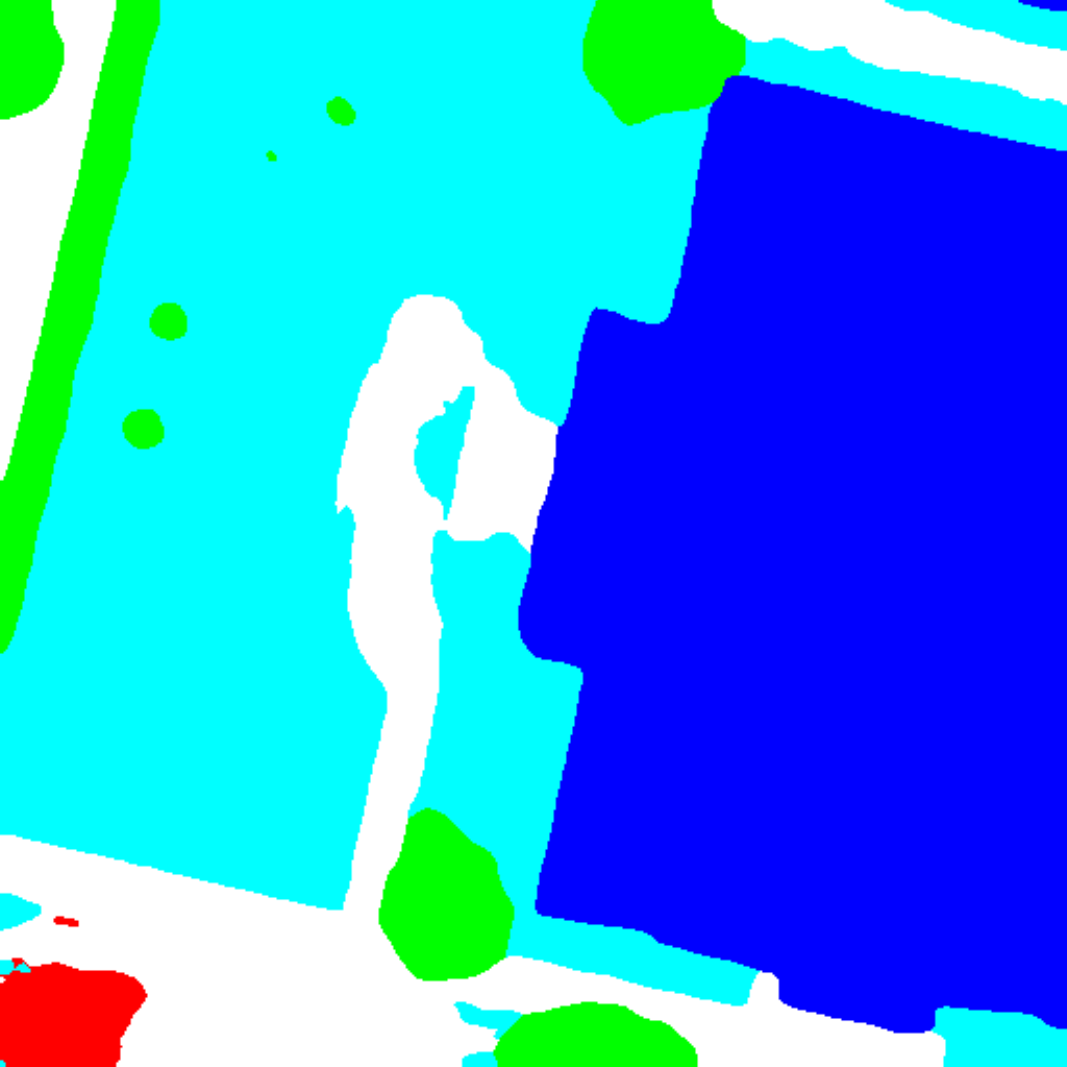}
		
		\vspace{1mm}
		
		\includegraphics[scale=0.094]{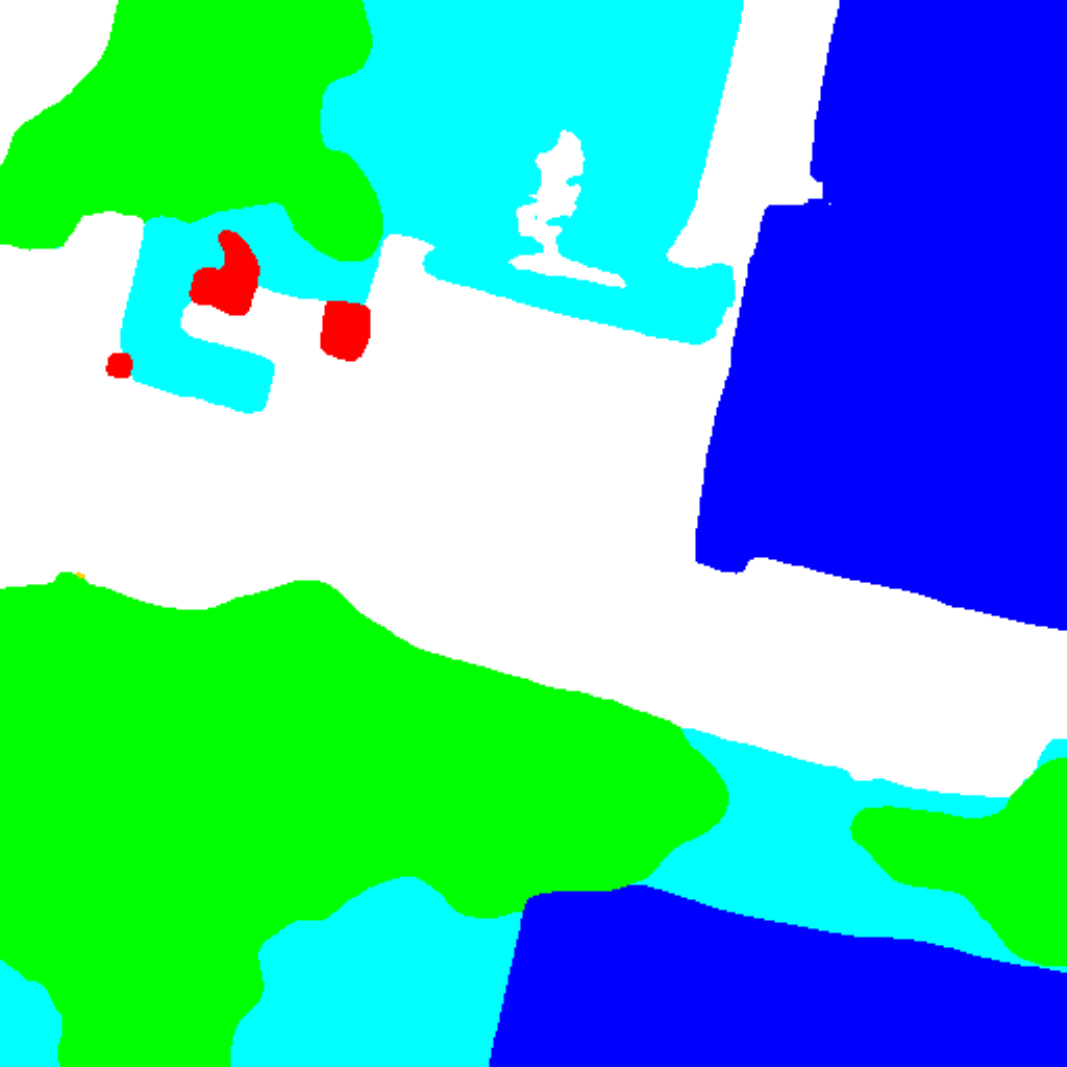}
		
		\vspace{1mm}
		
		\includegraphics[scale=0.094]{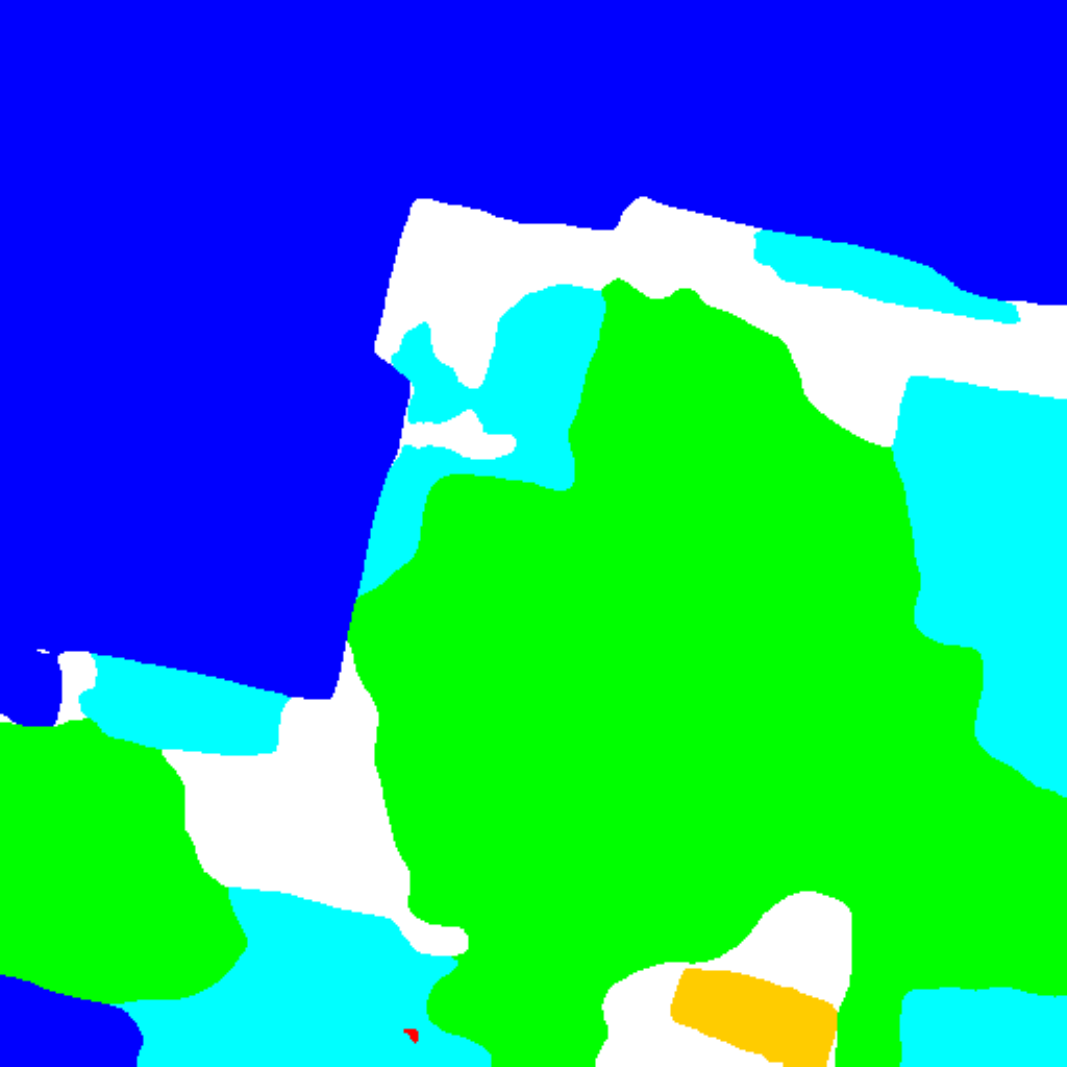}
		
		\vspace{1mm}
		
		\includegraphics[scale=0.094]{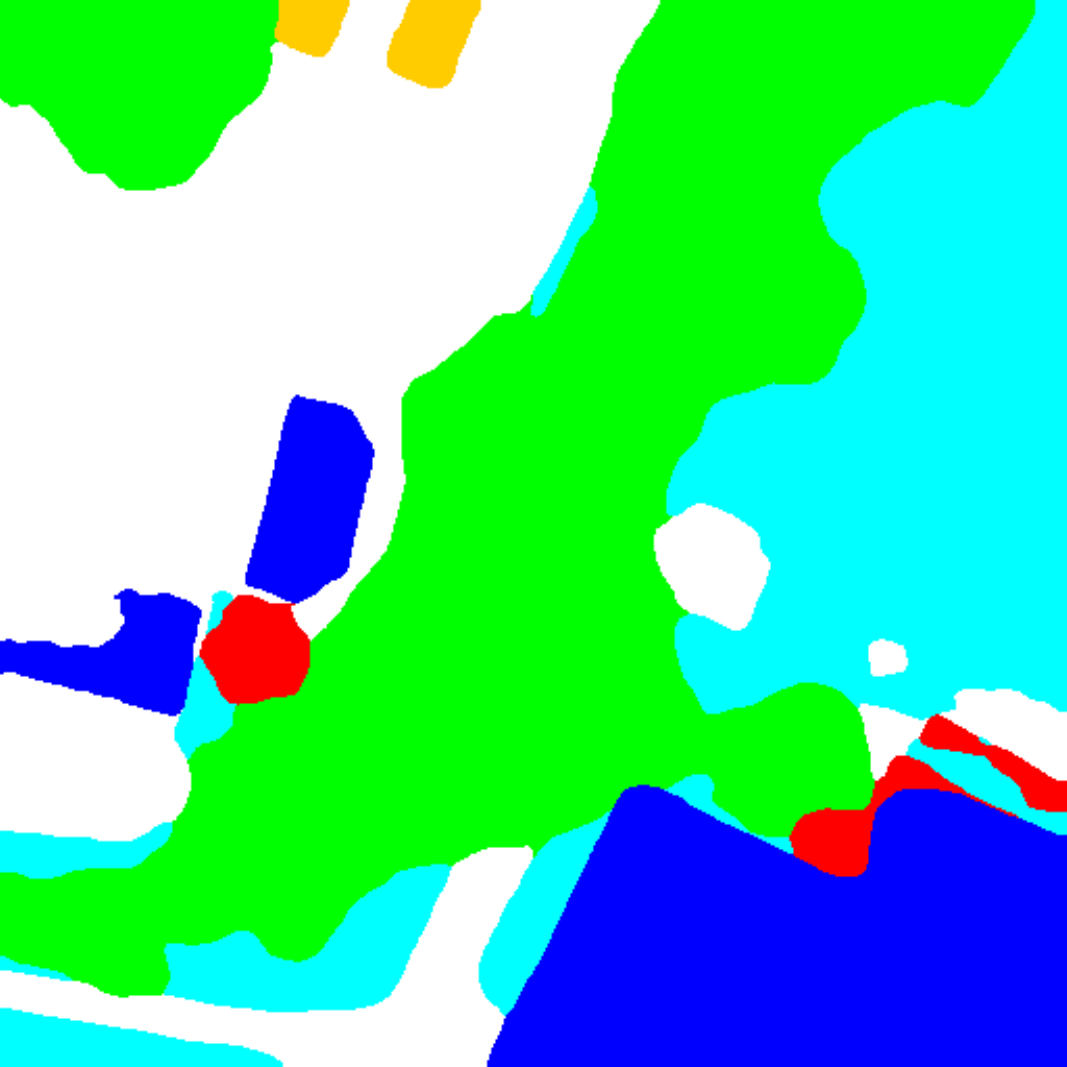}
		
		\vspace{1mm}
		
		\includegraphics[scale=0.094]{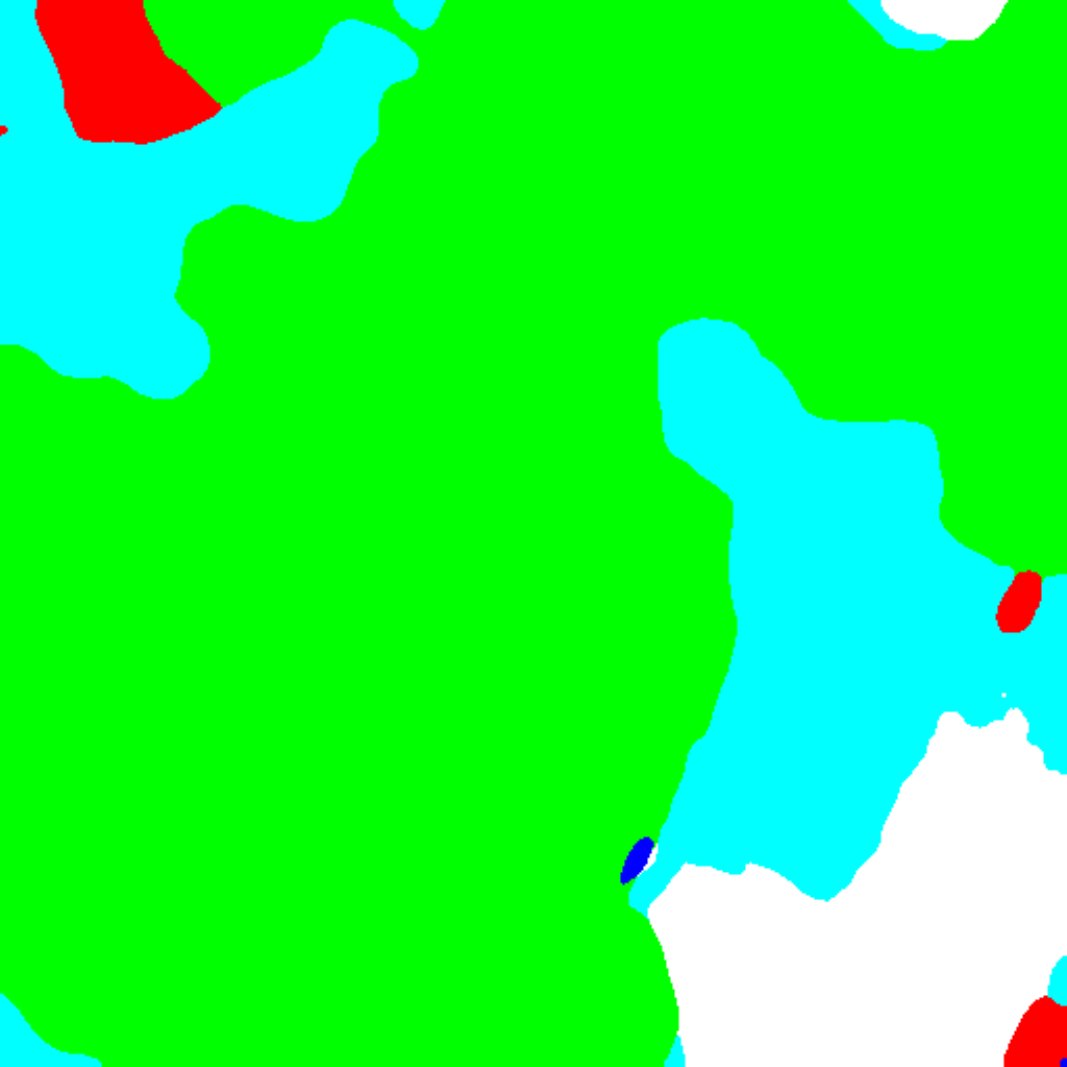}
		\centerline{(i)}
	\end{minipage}
	\begin{minipage}[t]{0.092\linewidth}
		\centering
		\includegraphics[scale=0.094]{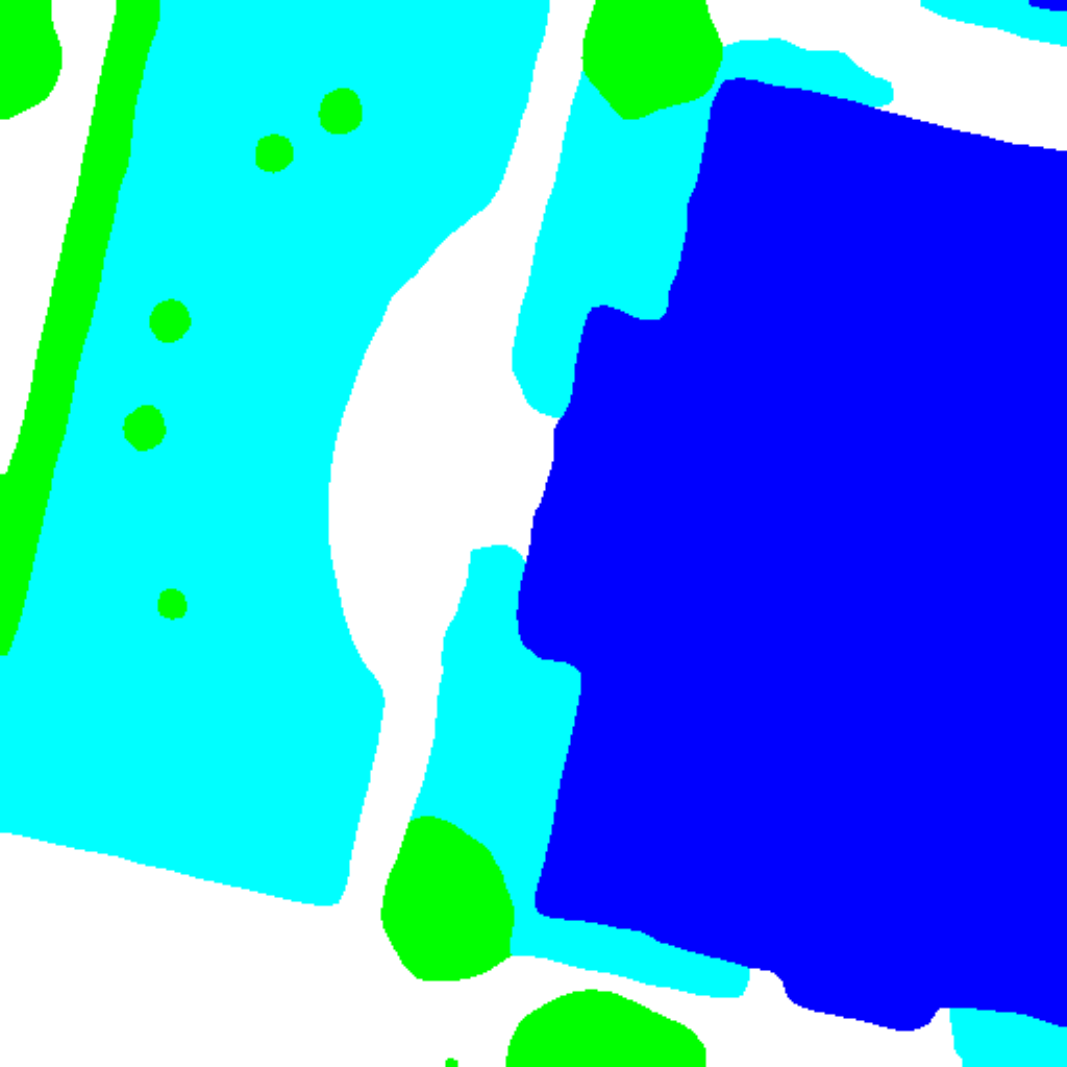}
		
		\vspace{1mm}
		
		\includegraphics[scale=0.094]{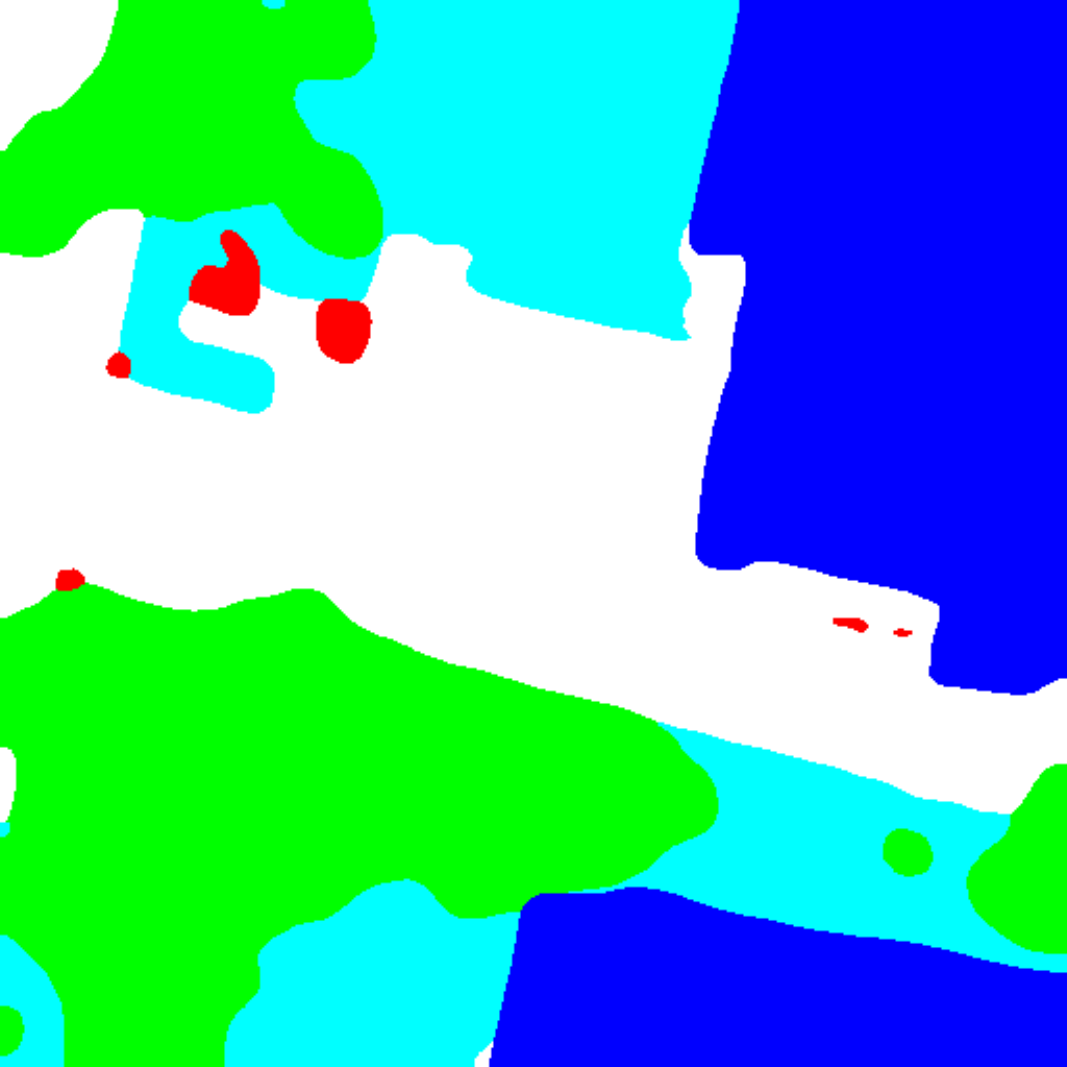}
		
		\vspace{1mm}
		
		\includegraphics[scale=0.094]{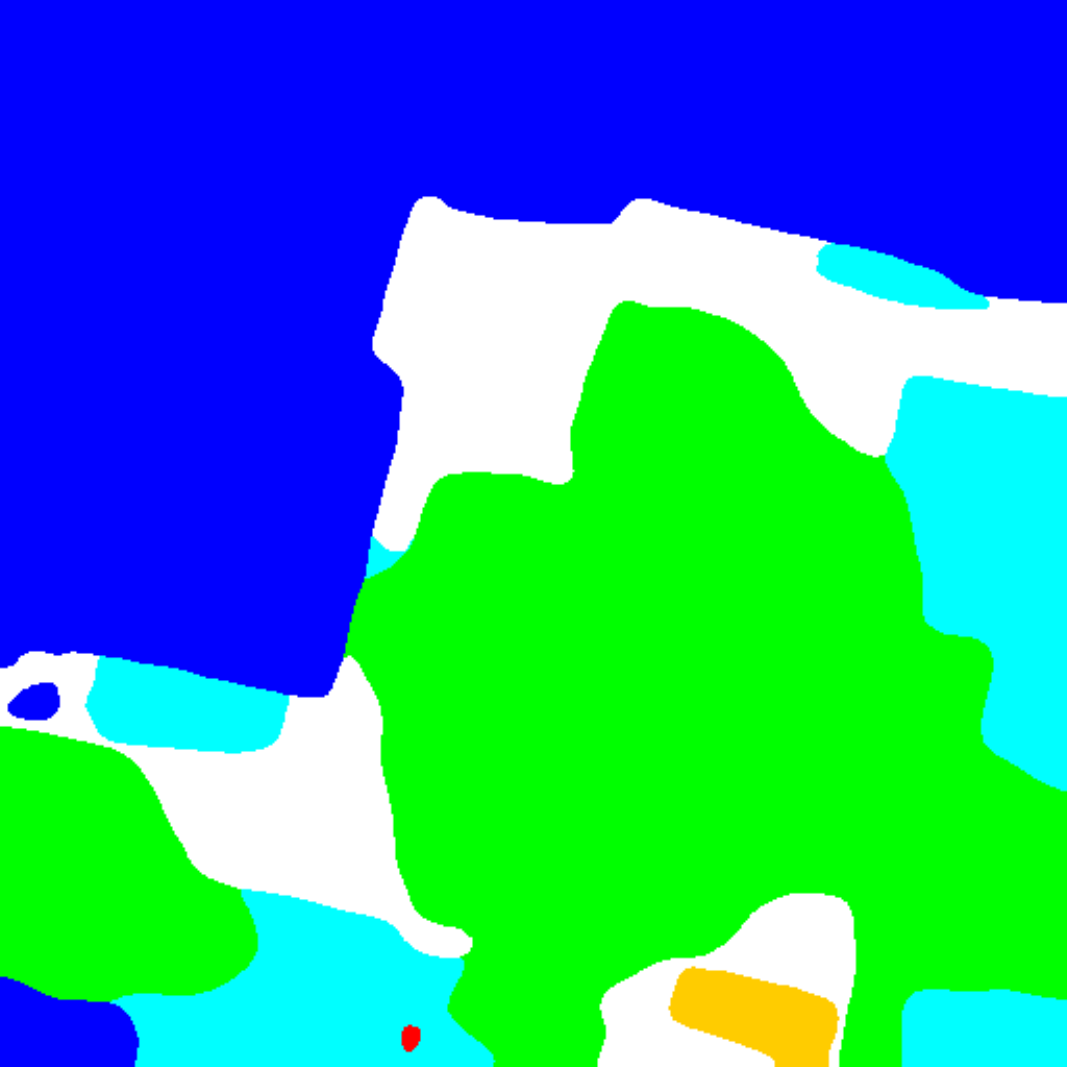}
		
		\vspace{1mm}
		
		\includegraphics[scale=0.094]{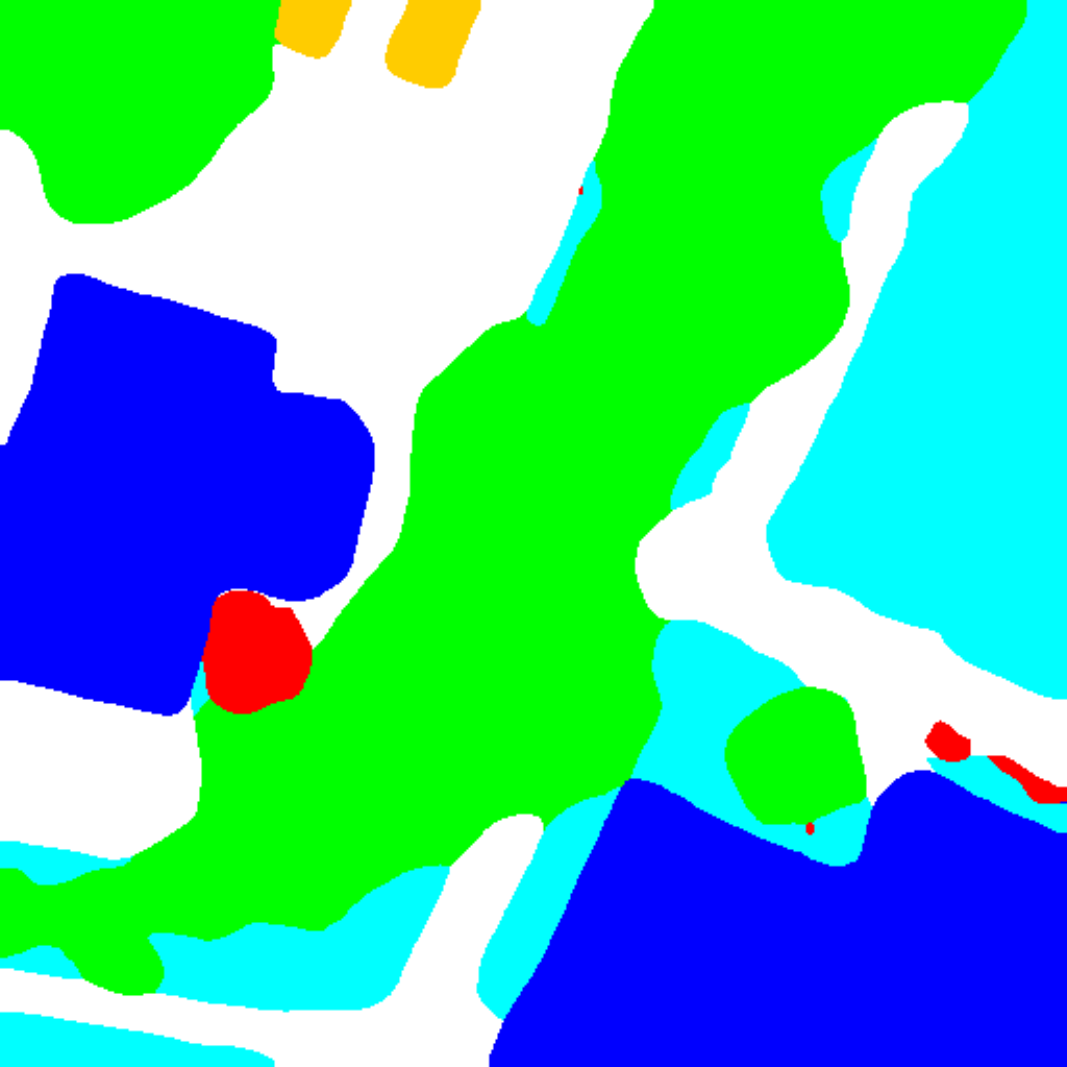}
		
		\vspace{1mm}
		
		\includegraphics[scale=0.094]{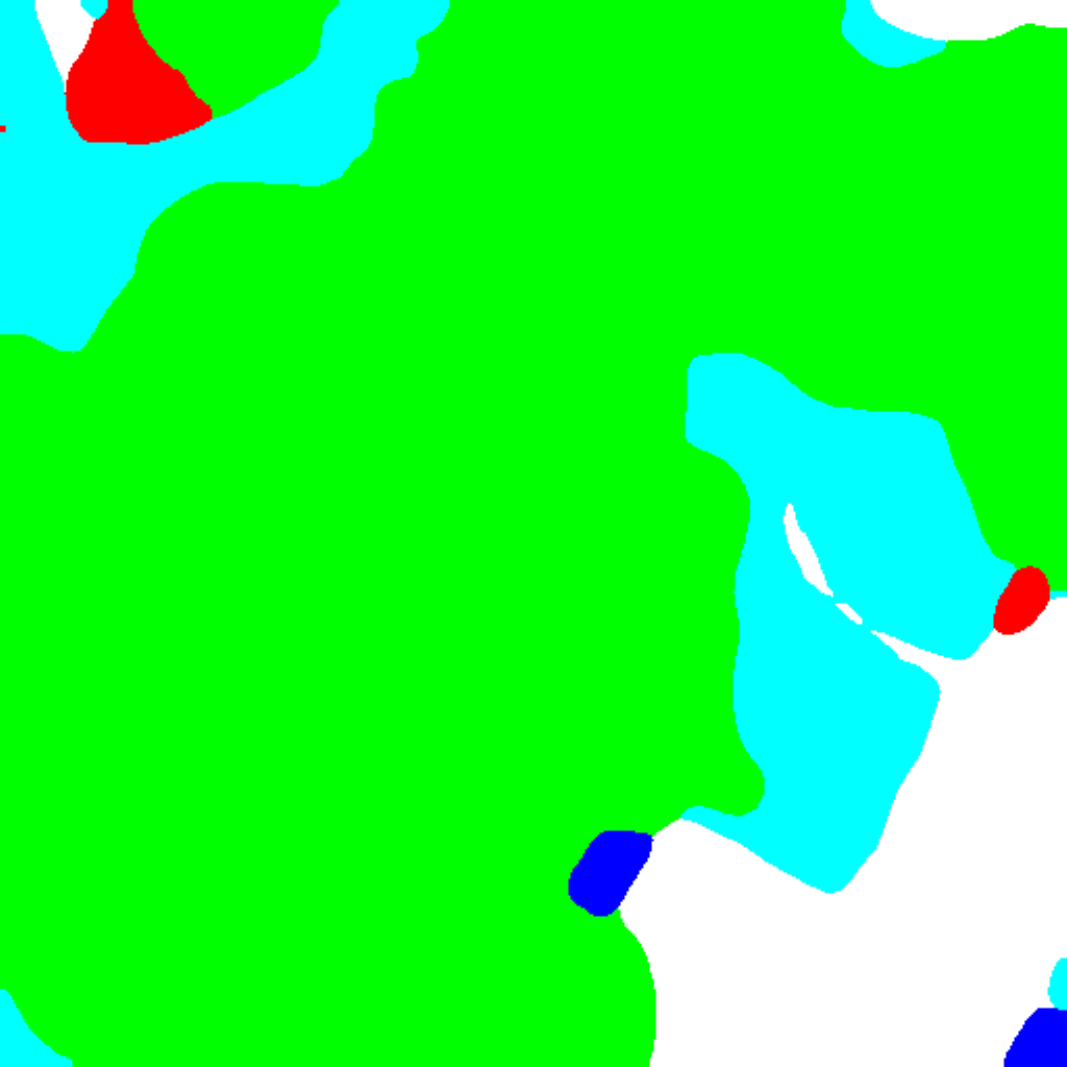}
		\centerline{(j)}
	\end{minipage}
	\begin{minipage}{\linewidth}
		\hfill
		\includegraphics[width=\textwidth]{image/Tuli.pdf}
	\end{minipage}
	\caption{Visualization segmentation results of different methods on Potsdam ((a) Input Image, (b) Ground Truth, (c) BiseNet, (d) SwiftNet, (e) ShelfNet, (f) BANet, (g) ABCNet, (h) DDRNet, (i) UNetFormer, (j) BAFNet)} 
	\label{fig8}
\end{figure*}

\subsection{Ablation Experiment}
Ablation experiments were conducted both on the Vaihingen and Potsdam datasets to evaluate the proposed modules, and the results are shown in Table 3. During the ablation experiments conducted on different parts of the network, the previously mentioned hybrid loss was used for training. Consistent data augmentation was applied during the training phase, and the same augmentation strategy was used during testing. In Table 3, Cp represents the network containing only dependency path, RA represents the remote-local path consisting exclusively of the proposed efficient remote attention module, LA denotes the remote-local path consisting solely of the proposed multi-scale local attention module, and the FAM signifies the proposed feature aggregation module.
\begin{table*}
	\caption{Ablation experiments on Vaihingen and Potsdam datasets\label{tab:table3}}
	\centering
	\begin{tabular}{|c||c||c||c||c||c||c||c||c||c|}
		\hline
		Dataset & Method  & Imp. Surf & Building & Low Veg.& Tree & Car & mean F1 & OA($\%$) &mIoU($\%$) \\
		\hline
		&Cp (ResNet18)&91.43&94.83&83.20&89.34&77.48&87.26&89.69&77.92\\
		\cline{2-10}
		&Cp (VAN-B0)&91.80&94.89&83.38&89.42&77.49&87.40&89.99	&78.15\\	
		\cline{2-10}
		Vaihingen&Cp+LA (sum)&92.97&95.52&84.65&90.11&88.93&90.44	&91.00&82.75\\
		\cline{2-10}
		&Cp+RA (sum)&92.86&95.54&84.87&90.19&88.63&90.42	&91.03&82.71\\
		\cline{2-10}
		&Cp+RA+LA (sum)&92.98&95.68&\pmb{84.97}&\pmb{90.31}&88.68&90.53&91.16&82.89\\	
		\cline{2-10}
		&Cp+RA+LA+FAM&\pmb{93.24}&\pmb{95.90}&84.75&89.98&\pmb{89.66}&\pmb{90.70}&\pmb{91.16}&\pmb{83.20}\\
		\hline
		&Cp (ResNet18)&91.97&95.37&86.72&88.30&93.49	&91.17&89.90&83.93\\
		\cline{2-10}
		&Cp (VAN-B0)&92.00&96.08&86.97&88.63&93.06&91.35	&90.35&84.22\\
		\cline{2-10}
		Potsdam&Cp+LA (sum)&93.02&96.37&87.81&89.19&95.65&92.41&91.01&86.07\\
		\cline{2-10}
		&Cp+RA (sum)&93.07&96.58&87.58&88.95&96.07&92.45&91.09&86.17\\
		\cline{2-10}
		&Cp+RA+LA (sum)&93.08&\pmb{96.79}&87.87&89.15&\pmb{96.22}&92.62&91.16&86.47\\
		\cline{2-10}
		&Cp+RA+LA+FAM&\pmb{93.12}&96.73&\pmb{87.98}&\pmb{89.35}&96.14&\pmb{92.66}&\pmb{91.28}&\pmb{86.53}\\
		\hline
	\end{tabular}
\end{table*}

{\bf{Baseline:}} ResNet18 \cite{he2016deep} is used as the baseline, and the feature map obtained from the last stage is directly upsampled to the size of the original input image.

{\bf{Ablation of dependency path:}} By comparing the segmentation results of VAN-B0 and ResNet18 in Table 3, it can be seen that the effect of VAN-B0 is better than that of ResNet18, with a slight improvement in mean F1, OA, and mIoU on both the Vaihingen and Potsdam datasets. It is worth noting that VAN-B0 has fewer parameters than ResNet18, suggesting that in lightweight segmentation networks, feature extraction networks with large kernel attention are more effective than traditional small-scale convolutions. This comparison verifies the importance of capturing long-range dependencies and the rationality of dependency path design.

{\bf{Ablation of MSLAM:}} 
Compared with CP and CP+LA, the latter incorporates a multi-scale local attention module as the core to establish an additional segmentation path alongside the dependency path, thereby enhancing segmentation performance effectively. On the Vaihingen dataset, the mean F1 score has shown an improvement of 3.04, the OA has increased by 1.01$\%$, and the mIoU has seen a boost of 4.6$\%$. On the Potsdam dataset, the mean F1 score has shown an improvement of 1.06, the OA has increased by 0.66$\%$, and the mIoU has seen an improvement of 1.85$\%$. This suggests that incorporating a path capable of capturing multi-scale local detailed information in addition to dependency path in a segmentation network can significantly enhance its performance.

{\bf{Ablation of ERAM:}} 
Comparing CP and CP+RA methods, the latter incorporates an efficient remote attention module as the core to establish an additional segmentation path alongside the dependency path. This enhancement effectively boosts segmentation performance on both two datasets. The F1 score for each category demonstrates improvement when compared to the segmentation network constructed without the remote attention module. On the Vaihingen and Potsdam datasets, the mIoU is increased by 4.56$\%$ and 1.95$\%$, respectively, which fully demonstrates the significance of incorporating structures capable of capturing long-range dependencies within the remote-local path.

{\bf{Ablation of remote-local path:}} We utilize efficient remote attention and multi-scale local attention to establish the remote-local path.
When comparing CP with CP+RA+LA (sum), the latter incorporates both the proposed ERAM and MSLAM to establish the remote-local path,
and a simple addition operation is employed to fuse features from distinct paths. The results of the two datasets presented in Table 3 demonstrates a significant improvement in segmentation performance by incorporating path consisting of both remote attention and local attention, in contrast to networks that solely rely on dependency path. On the Vaihingen and Potsdam datasets, the mIoU shows an increase of 4.74$\%$ and 2.25$\%$, respectively. Compared to the results of CP+RA and CP+LA, the performance of CP+RA+LA (sum) shows improvement, suggesting the importance of concurrently capturing long-range dependencies and local details along the remote-local path. This demonstrates the rationality and effectiveness of our designed remote-local path.

{\bf{Ablation of FAM:}} 
We design a FAM to effectively integrate the feature maps generated by two different paths. For CP+RA+LA+FAM, which is our proposed BAFNet.
In the results presented in Table 3, compared with CP+RA+LA+FAM and CP+RA+LA (sum), FAM leads to an optimization of the overall segmentation performance on both the Vaihingen and Potsdam datasets. The mean F1 score, OA, and mIoU are better than the direct addition fusion method, indicating the effectiveness of FAM.

\begin{figure*}
	\centering
	\includegraphics[scale=1.25]{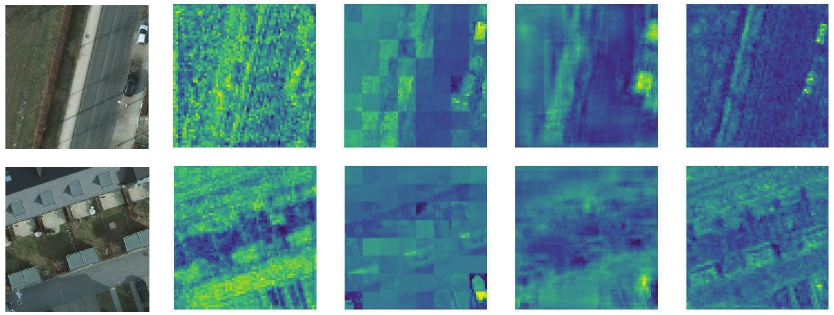}
	\centerline{\hspace {0.9cm}Input Image\hspace {1.6cm}Local Context\hspace {1.3cm}Window Context\hspace {1.2cm}Remote Context\hspace{0.8cm}Remote-Local Context}
	\caption{Feature map visualization for local context, window context, remote context, and remote-local context.}
	\label{fig9}
\end{figure*}

\subsection{Visualization of Feature Maps}
To explore the anticipated impact of our proposed MSLAM and ERAM, we visually represent the feature maps related to RLAM within the remote-local block, as shown in Fig. 9. The MSLAM has been devised for the purpose of extracting intricate details from images, and the resulting feature map is called the Local Context. The native window attention proposed in the Swin Transform is designed to calculate self-attention within a specific window. The resulting feature map is called Window Context. The ERAM improves window self-attention by removing window shift operations and employing deepwise convolution to facilitate interactions among various windows. The resulting feature map is called Remote Context. Finally, the integration of Local Context and Remote Context results in Remote-Local Context, which is the feature map outputted by each RLAM.

Through the analysis of the second column, Local Context in Fig. 9, it is observed that the MSLAM has the capability to extract a significant amount of detailed information. However, the convolution operation's local nature and the lack of guidance from long-range dependencies result in the emergence of many unnecessary lines in the feature map.
For example, the feature map of road in the first row displays varying colors as a result of the shades of trees and lane markings. Likewise, the same low vegetation region in the original image displays various colors, leading to numerous spots in the feature map. The skylight's outline on the building in the second row is overly obvious, while the skylight does not falling into a specific category of its own, it adds to the complexity of details, leading to potential confusion in segmentation.
In the third column of Fig. 9, it is observed that through Window Attention, the features within the same window are relatively smoothed. However, the entire feature map has numerous jagged boundaries, which are attributed to window partitioning operation.
The fourth column in Fig. 9, Remote Context, which is generated by convolution following the window attention. This approach facilitate information exchange between neighboring windows, effectively reducing the jagged edges. However, feature maps with a wide range of contexts exhibit a deficiency in detail. The final Remote-Local Context incorporates detailed information acquired by the local module and long-range dependencies acquired by the remote module. 
The features of the same category derived from the fusion of two feature maps exhibit consistency, and features of different categories display noticeable distinctions, resulting in clear boundaries that facilitate segmentation. This demonstrates the rationality of our constructed RLAM.

\section{Summary}

In this paper, we propose a novel lightweight bilateral network, BAFNet, designed for RS image segmentation tasks with limited computational resources. It employs a lightweight backbone with large kernel attention to extract semantic information, effectively capturing long-range dependencies within the images under the constraints of computation.
Meanwhile, we propose an efficient remote attention module and a multi-scale local attention module as the fundamental components to establish a remote-local path. Additionally, we introduce a feature aggregation module to effectively leverage the features extracted from both paths. The experiments carried out on the Vaihingen and Potsdam datasets demonstrated the effectiveness of the proposed module, and BAFNet further improved the performance of the lightweight segmentation network. Compared to multiple advanced lightweight segmentation models, BAFNet demonstrates advantages in both segmentation accuracy and network complexity.
Compared to the non-lightweight state-of-the-art models on the dataset, BAFNet has achieved comparable segmentation performance while significantly reducing network parameters and floating-point operations by over tenfold. 
We hope that future researches will prioritize the exploration of lightweight backbones to fully exploit their potential and develop more rational network architectures to improve accuracy and practicality.
\bibliographystyle{IEEEtran}
\bibliography{reference.bib}

\end{document}